\documentclass[acmsmall]{acmart}
\AtBeginDocument{%
  \providecommand\BibTeX{{%
    \normalfont B\kern-0.5em{\scshape i\kern-0.25em b}\kern-0.8em\TeX}}}

\usepackage{soul}
\usepackage{url}
\usepackage{caption}
\usepackage{graphicx}
\usepackage{amsmath}
\usepackage{amsthm}
\usepackage{booktabs}
\usepackage{algorithm}
\usepackage{algpseudocode}
\usepackage{caption}
\usepackage{subcaption}
\usepackage{wrapfig}
\usepackage{xcolor}
\usepackage{tabularx}
\usepackage{enumitem}

\newtheorem{definition}{Definition}
\newtheorem{theorem}{Theorem}
\newtheorem{assumption}{Assumption}
\newtheorem{lemma}{Lemma}

\newcommand{\sysname}{FairSAOML}
\newcommand{\sysnameregret}{FairSAR}

\makeatletter
\newcommand{\multiline}[1]{%
  \begin{tabularx}{\dimexpr\linewidth-\ALG@thistlm}[t]{@{}X@{}}
    #1
  \end{tabularx}
}
\makeatother

\makeatletter
\def\endthebibliography{%
  \def\@noitemerr{\@latex@warning{Empty `thebibliography' environment}}%
  \endlist
}
\makeatother

\setcopyright{acmlicensed}
\acmYear{2024}
\acmDOI{10.1145/3648684}

\acmJournal{TKDD}
\acmVolume{1}
\acmNumber{1}
\acmArticle{1}
\acmMonth{1}

\begin{document}

\title{Dynamic Environment Responsive Online Meta-Learning with Fairness Awareness}

\author{Chen Zhao}
\email{chen\_zhao@baylor.edu}
\orcid{0000-0002-6400-0048}
\affiliation{%
  \institution{Baylor University}
  \streetaddress{One Bear Place \#97356}
  \city{Waco}
  \state{Texas}
  \country{USA}
  \postcode{76798-7356}
}

\author{Feng Mi}
\email{feng.mi@utdallas.edu}
\affiliation{%
  \institution{The University of Texas at Dallas}
  \streetaddress{800 W. Campbell Road}
  \city{Richardson}
  \state{Texas}
  \country{USA}
  \postcode{75080-3021}
}

\author{Xintao Wu}
\email{xintaowu@uark.edu}
\affiliation{%
  \institution{University of Arkansas}
  \streetaddress{4183 Bell Engineering Center}
  \city{Fayetteville}
  \state{Arkansas}
  \country{USA}
  \postcode{72701}
}

\author{Kai Jiang}
\email{kai.jiang@utdallas.edu}
\affiliation{%
  \institution{The University of Texas at Dallas}
  \streetaddress{800 W. Campbell Road}
  \city{Richardson}
  \state{Texas}
  \country{USA}
  \postcode{75080-3021}
}

\author{Latifur Khan}
\email{lkhan@utdallas.edu}
\affiliation{%
  \institution{The University of Texas at Dallas}
  \streetaddress{800 W. Campbell Road}
  \city{Richardson}
  \state{Texas}
  \country{USA}
  \postcode{75080-3021}
}

\author{Feng Chen}
\email{feng.chen@utdallas.edu}
\affiliation{%
  \institution{The University of Texas at Dallas}
  \streetaddress{800 W. Campbell Road}
  \city{Richardson}
  \state{Texas}
  \country{USA}
  \postcode{75080-3021}
}

\renewcommand{\shortauthors}{Zhao, et al.}

\begin{abstract}
  The fairness-aware online learning framework has emerged as a potent tool within the context of continuous lifelong learning. In this scenario, the learner's objective is to progressively acquire new tasks as they arrive over time, while also guaranteeing statistical parity among various protected sub-populations, such as race and gender, when it comes to the newly introduced tasks.
A significant limitation of current approaches lies in their heavy reliance on the \textit{i.i.d} (independent and identically distributed) assumption concerning data, leading to a static regret analysis of the framework. Nevertheless, it's crucial to note that achieving low static regret does not necessarily translate to strong performance in dynamic environments characterized by tasks sampled from diverse distributions.
In this paper, to tackle the fairness-aware online learning challenge in evolving settings, we introduce a unique regret measure, FairSAR, by incorporating long-term fairness constraints into a strongly adapted loss regret framework.
Moreover, to determine an optimal model parameter at each time step, we introduce an innovative adaptive fairness-aware online meta-learning algorithm, referred to as \sysname{}. 
This algorithm possesses the ability to adjust to dynamic environments by effectively managing bias control and model accuracy. The problem is framed as a bi-level convex-concave optimization, considering both the model's primal and dual parameters, which pertain to its accuracy and fairness attributes, respectively.
Theoretical analysis yields sub-linear upper bounds for both loss regret and the cumulative violation of fairness constraints. Our experimental evaluation on various real-world datasets in dynamic environments demonstrates that our proposed \sysname{} algorithm consistently outperforms alternative approaches rooted in the most advanced prior online learning methods.


\end{abstract}

\ccsdesc[500]{Computing methodologies~Artificial intelligence}
\ccsdesc[500]{Computing methodologies~Machine learning}
\ccsdesc[300]{Applied computing~Law, social and behavioral sciences}
\ccsdesc[100]{Social and professional topics~User characteristics}

\keywords{
Fairness, Online meta-learning, Changing environments, Adaption
}


\maketitle

\section{Introduction}
\label{sec:introduction}
    In the real world, data that includes biases are often collected incrementally over time, and the underlying distribution assumptions can undergo significant changes at critical junctures. 
A case in point is a recent report by the \textit{New York Times} \cite{Miller-2020-NYTimes}, which highlights that systematic algorithms exhibited increased discriminatory tendencies towards African Americans in the context of bank loans during the COVID-19 pandemic compared to the pre-pandemic era. These algorithms are constructed from a series of sequentially gathered data streams, where decision-making exhibits bias towards the protected racial population at each step. This situation underscores two key issues: (1) Online algorithms typically neglect the crucial aspect of fairness in learning, where fairness is defined as the equality of predictive performance across different sub-populations, ensuring that a model's predictions remain statistically independent of protected characteristics (\textit{e.g.,} race). (2) Machine learning models heavily rely on the \textit{i.i.d} assumption, which becomes untenable when the environment undergoes changes, as exemplified by shifts occurring before and after the pandemic.

To effectively manage bias over time, particularly in the context of ensuring fairness across various protected sub-populations, fairness-aware online algorithms are designed to address supervised learning problems where fairness is a prominent concern. These algorithms aim to sequentially train predictive models that remain unbiased. In particular, the objective of these algorithms is twofold: first, to ensure that the static loss regret, which measures the cumulative loss of the learner against the best-fixed action in hindsight, and second, to limit the violation of various fairness principles, both exhibit sub-linear growth in the total number of time steps \cite{zhao-KDD-2021}. It's worth noting that while these approaches achieve cutting-edge theoretical guarantees, it's important to recognize that the metric of static regret holds significance primarily in stable or stationary environments. Low static regret, however, doesn't necessarily translate to excellent performance in changing environments because time-invariant benchmarks may perform poorly under such circumstances \cite{zhang-2020-AISTATS}.

To overcome the challenge posed by changing environments in online learning, two distinct notions of regret have garnered attention: strongly adaptive regret \cite{daniely15-2015-ICML} and dynamic regret \cite{Zinkevich-ICML-2003}. These concepts offer differing perspectives on handling changes over time. Dynamic regret takes a global approach, addressing changes in environments by comparing the cumulative loss of the learner against a sequence of comparators. Importantly, it allows these comparators to evolve over time, reflecting the dynamic nature of the learning process. Conversely, strongly adaptive regret adopts a more localized viewpoint, giving greater consideration to short time intervals. This type of regret can be seen as the maximum regret statistic across all intervals \cite{daniely15-2015-ICML}. While some recent works \cite{zhang-2020-AISTATS, zhang-nips-2018, Jun-2017-AISTATS} have made strides in achieving sub-linear loss regret in online learning within changing environments, they often overlook the crucial aspect of learning with fairness. This neglect of fairness, which is a fundamental characteristic of human intelligence, remains a significant limitation in these approaches.

In this paper, we present a new challenge, namely fairness-aware online meta-learning in changing environments. In this scenario, a series of data batches or tasks are collected sequentially over time, with the environments associated with these tasks potentially undergoing variations. Our primary objectives in this research are twofold: Firstly, we aim to extend the applicability of predictive learning accuracy and model fairness to novel and evolving environments. Secondly, we endeavor to minimize both loss regret and the cumulative violation of fairness constraints, ensuring that they exhibit sublinear growth over time.

To achieve these goals, we introduce a novel online learning algorithm named fair strongly adaptive online meta-learner (\sysname{}). This algorithm updates model parameters through a two-level approach: online fair interval-level learning and meta-level learning. These two levels of problems interact with two sets of parameters: primal parameters $\boldsymbol{\theta}$, which pertain to model accuracy, and dual parameters $\boldsymbol{\lambda}$, which govern fairness considerations. To provide more details, we draw inspiration from the concept of \textit{learning with expert advice} \cite{Jun-2017-AISTATS}, and we carefully design three alternative sets of intervals. At each time step $t\in[T]$, a subset of intervals is chosen to activate several experts, with each active expert running an interval-specific algorithm. An expert takes a meta-solution pair $(\boldsymbol{\theta}_{t-1},\boldsymbol{\lambda}_{t-1})$ from the previous time as input and generates an interval-level solution $(\boldsymbol{\theta}_{t,I},\boldsymbol{\lambda}_{t,I})$ for the specific interval $I$. A meta-algorithm combines the weighted contributions of all experts to form a solution pair $(\boldsymbol{\theta}_{t},\boldsymbol{\lambda}_{t})$ at time $t$, which is then utilized to make predictions for the subsequent time step. This approach allows us to address the challenges of fairness-aware online meta-learning in changing environments effectively.

The main contributions of this paper are summarized:
\begin{itemize}
    \item In this paper, we propose a novel framework addressing the problem of fairness-aware online meta-learning in changing environments. We start with the introduction of a novel adaptive fairness-aware regret FairSAR. A novel algorithm \sysname{} is further proposed to find a good decision sequentially. At each time, the problem is formulated as a constrained bi-level convex-concave optimization with respect to a primal-dual parameter pair.

    \item Based on varying assumptions and motivations, we introduce three distinct sets of intervals, leading to the creation of three different versions of our proposed \sysname{} algorithm.
    
    \item Theoretically grounded analysis justifies the efficiency and effectiveness of all variants of \sysname{} by demonstrating tighter bounds $\mathcal{O}\Big((\tau\log T)^{1/2}\Big)$ for the loss regret and $\mathcal{O}\Big((\tau T\log T)^{1/4}\Big)$ for violation of fairness constraints.
    
    \item We validate the performance of our approach with state-of-the-art techniques on real-world datasets. Our results demonstrate that \sysname{} can effectively adapt both accuracy and fairness in changing environments, and it shows substantial improvements over the best prior works.
\end{itemize}

This paper is organized as follows. In Section \ref{sec:relatedwork}, some related works are introduced. Section \ref{sec:preliminaries} provides notations and some backgrounds of this paper. In Section \ref{sec:methodology}, we detail the proposed methodology. In Section \ref{sec:analysis}, we discuss the theoretically grounded analysis for the learning approach. Empirical settings and results on real-world benchmarks compared with cutting-edge techniques are given in Section \ref{sec:experiments} and Section \ref{sec:results}. Finally, this paper is concluded in Section \ref{sec:conclusion}.

\section{Related Work}
\label{sec:relatedwork}
    \textbf{Changing environments in online learning.} Since the pioneering work \cite{Zinkevich-ICML-2003} in online learning, numerous subsequent researches \cite{Hazan-2020-LT,xie-2020-aaai} have been developed under the assumption of a stationary environment with static regret. Low static regret, however, cannot imply a good performance in a changing environment due to time-invariant comparators. To address this limitation, two regret metrics, dynamic regret \cite{Zinkevich-ICML-2003} and adaptive regret \cite{Hazan-2007-ARegret}, is devised to measure the learner's performance in changing environments. To bound the general dynamic regret, the path-length of comparators \cite{Zinkevich-ICML-2003,zhang-nips-2018} is introduced and further developed.
Unlike dynamic regret, adaptive regret handles changing environments from a local perspective by focusing on comparators in short intervals. 
To reduce the time complexity of adaptive regret-based online algorithms, geometric covering intervals \cite{daniely15-2015-ICML,Jun-2017-AISTATS,zhang-2020-AISTATS} and data streaming techniques \cite{gyorgy-2012-efficient} are developed. 
Although existing methods achieve state-of-the-art performance, a major drawback is that they immerse in minimizing objective functions but ignore the model fairness of prediction. 

\textbf{Fairness-aware online learning} problems assume individuals arrive one at a time and the goal of such algorithms is to train predictive models free from biases. 
From the perspective of optimization, group fairness notions are normally considered as constraints added to learning objectives.
However, when the constraints are complex, the computational burden of the projection onto constraints may be too high. 
Several closely related works, including FairFML \cite{zhao-KDD-2021}, FairGLC \cite{GenOLC-2018-NeurIPS}, FairAOGD \cite{AdpOLC-2016-ICML}, aims to improve the theoretic guarantees by relaxing the output through a simpler closed-form projection. 
However, these methods are not ideal for continual lifelong learning with changing task distributions, as they assume that all samples come from the same data distribution.

\textbf{Online meta-learning} addresses the issue of learning with fast adaptation, where a meta-learner learns knowledge transfer from history tasks onto new coming ones. FTML \cite{Finn-ICML-2019} can be considered as an application of MAML \cite{Finn-ICML-2017-(MAML)} in the setting of online learning. 
FairFML \cite{zhao-KDD-2021} extends FTML by controlling bias in an online working paradigm with task-specific adaptation. 
Unfortunately, none of such techniques are devised to adapt to changing environments. 

Although a recent work \cite{zhao2022adaptive} tackles the problem of fairness-aware online learning for changing environments, it heavily depends on the assumption that the number of times is known in advance and unchanged. The number of learning processes is hence fixed. Besides, due to the setting of intervals in this work, the learning efficiency at the beginning times is low.

In this paper, to bridge the above-mentioned areas, we study the problem of fairness-aware online meta-learning to deal with changing task environments. In particular, at each time, model parameters are determined by the proposed novel algorithm \sysname{}. This algorithm refers to ideas of dynamic programming and expert tracking techniques. Inspired by fairness-aware online learning and meta-learning, a bi-level adaptation strategy is used to accommodate changing environments and learn models with accuracy and fairness.

\section{Preliminaries}
\label{sec:preliminaries}

\subsection{Notations}
An index set of a sequence of tasks is defined as $[T]=\{1,2,...,T\}$ and $[t,T]=\{t,t+1,\cdots,T\}$. 
Vectors are denoted by lowercase boldface letters. 
Scalars are denoted by lowercase italic letters. 
Some important notations are listed in Table \ref{tab:notation}.

\label{sec:notations}
\begin{table}[!t]
    \centering
    \caption{Important notations and corresponding descriptions.}
    \begin{tabular}{l|l}
        \toprule
        \textbf{Notations} & \textbf{Descriptions}  \\
        \midrule
        $T$ & Total number of learning tasks\\
        $t$ & Indices of tasks\\
        $\tau$ & Length of time intervals in general\\
        $\mathcal{D}_t^S, \mathcal{D}_t^V, \mathcal{D}_t^Q$ & Support/Validation/Query set of data $\mathcal{D}_t$\\
        $\boldsymbol{\theta}_t, \boldsymbol{\lambda}_t$ & Meta-level primal/dual parameters at round $t$\\
        $\boldsymbol{\theta}_{t,I}, \boldsymbol{\lambda}_{t,I}$ & Interval-level primal/dual parameters for an expert $E_I$ at round $t$\\
        $f_t(\cdot)$ & Loss function at round $t$  \\
        $g_i(\cdot)$ & Fairness function \\
        $m$ & Total number of fairness notions\\
        $i$ & Indices of fairness notions\\
        $\mathcal{G}(\cdot)$ & Base learner\\
        $\mathcal{U}$ & Expert set\\
        $\mathcal{A}_t,\mathcal{S}_t$ & Active/Sleeping expert set at round $t$\\
        $\mathcal{I}$ & AGC interval set\\
        $\mathcal{C}_t$ & Target set of intervals at round $t$\\
        $\mathcal{B}$ & Relaxed primal domain \\
        $\prod_{\mathcal{B}}$ & Projection operation onto domain $\mathcal{B}$\\
        $\eta_1, \eta_2$ & Learning rates\\
        $p_{t,I}$ & Expert weight of $E_I$ at round $t$\\
        $\delta$ & Augmented constant\\
        \bottomrule
    \end{tabular}
    \label{tab:notation}
\end{table}

\subsection{Constraints for Group Fairness}
In general, group fairness criteria used for evaluating and designing machine learning models focus on the relationships between the protected attribute and the system output \cite{Zhao-ICDM-2019,Zhao-ICKG-1-2020,wang-2021-clear}. The problem of group unfairness prevention can be seen as a constrained optimization problem. For simplicity, we consider one binary protected attribute (\textit{e.g.} gender) in this work. However, our ideas can be easily extended to many protected attributes with multiple levels.

Let $\mathcal{Z=X\times Y}$ be the data space, where $\mathcal{X} = \mathcal{E} \cup\mathcal{S}$. Here $\mathcal{E} \subset \mathbb{R}^d$ is an input space, $\mathcal{S} = \{-1,1\}$ is a protected space, and $\mathcal{Y} = \{-1,1\}$ is an output space for binary classification. 
Given a task (batch) of samples
$\{\mathbf{e}_{i}, y_{i}, s_{i}\}_{i=1}^n \in(\mathcal{E\times Y\times S})$ 
where $n$ is the number of datapoints, 
a fine-grained measurement to ensure fairness in class label prediction is to design fair classifiers by controlling the notions of fairness between protected subgroups, such as demographic parity and equality of opportunity \cite{Wu-2019-WWW, Lohaus-2020-ICML}.
\begin{definition}[Notions of Fairness \cite{Wu-2019-WWW,Lohaus-2020-ICML}]
A classifier $h:\Theta\times\mathbb{R}^d\rightarrow\mathbb{R}$ is fair when its predictions are independent of the protected attribute $\mathbf{s}=\{s_i\}_{i=1}^n$.
To get rid of the indicator function and relax the exact values, a linear approximated form of the difference between protected subgroups is defined \cite{Lohaus-2020-ICML},
\begin{equation}
\begin{aligned}
\label{dbc definition}
    g(\boldsymbol{\theta})=\Big|\mathbb{E}_{(\mathbf{e},y,s)\in\mathcal{Z}}\Big[\frac{1}{\hat{p}_1(1-\hat{p}_1)}\Big(\frac{s+1}{2}-\hat{p}_1\Big)h(\boldsymbol{\theta},\mathbf{e})\Big] \Big|-\epsilon
\end{aligned}
\end{equation}
where $|\cdot|$ is the absolute function and $\epsilon>0$ is the fairness relaxation determined by empirical analysis. $\hat{p}_1$ is an empirical estimate of $pr_1$. $pr_1$ is the proportion of samples in group $s=1$ and correspondingly $1-pr_1$ is the proportion of samples in group $s=-1$. 
\end{definition}
Notice that, in Definition \ref{dbc definition}, when $\hat{p}_1=\mathbb{P}_{(\mathbf{e},y,s)\in\mathcal{Z}}(s=1)$, the fairness notion $g(\boldsymbol{\theta})$ is defined as the difference of demographic parity (DDP). Similarly, when $\hat{p}_1=\mathbb{P}_{(\mathbf{e},y,s)\in\mathcal{Z}}(y=1, s=1)$, $g(\boldsymbol{\theta})$ is defined as the difference of equality of opportunity (DEO) \cite{Lohaus-2020-ICML}.
Therefore, parameters $\boldsymbol{\theta}$ in the domain of a task is feasible if it satisfies the fairness constraint $g(\boldsymbol{\theta})\leq 0$. 

\subsection{Fairness-Aware Online Learning}
\label{sec:fairness-aware online learning}
The protocol of fairness-aware online convex optimization can be viewed as a repeated game between a learner and an adversary, where the learner is faced with tasks $\{\mathcal{D}_t\}_{t=1}^T$ one after another. At each round $t\in[T]$, 
\begin{itemize}
    \item \textbf{Step 1}: The learner selects a model parameter $\boldsymbol{\theta}_t$ in the fair domain $\Theta$.
    \item \textbf{Step 2}: The adversary reveals a loss function $f_t:\Theta\times\mathbb{R}^d\rightarrow\mathbb{R}$ and $m$ fairness functions $g_i:\Theta\times\mathbb{R}^d\rightarrow\mathbb{R},\forall i\in[m]$.
    \item \textbf{Step 3}: The learner incurs an instantaneous loss $f_t(\boldsymbol{\theta}_t,\mathcal{D}_t)$ and $m$ fairness notions $g_i(\boldsymbol{\theta}_t,\mathcal{D}_t),\forall i\in[m]$. 
    \item \textbf{Step 4}: Advance to $t+1$.
\end{itemize}


The goal of fairness-aware online learning \cite{GenOLC-2018-NeurIPS,zhao-KDD-2021} is to (1) minimize the loss regret over the rounds, which is to compare to the cumulative loss of the best-fixed model in hindsight and (2) ensure the total violation of fair constraints sublinearly increase in $T$. The loss regret is typically referred to as \textit{static regret} since the comparator is time-invariant. To control bias and ensure group fairness across different protected sub-populations, fairness notions are considered as constraints on optimization problems.
\begin{equation}
\begin{aligned}
\label{eq:static-regret}
    \min_{\boldsymbol{\theta}_1,...,\boldsymbol{\theta}_T\in\Theta} \quad & \text{Regret}(T) = \sum_{t=1}^T f_t(\boldsymbol{\theta}_t,\mathcal{D}_t)
    - \min_{\boldsymbol{\theta}\in\Theta}\sum_{t=1}^T f_t(\boldsymbol{\theta},\mathcal{D}_t) \\
    \text{subject to} \quad &\sum_{t=1}^T g_i(\boldsymbol{\theta}_t,\mathcal{D}_t) \leq \mathcal{O}(T^\gamma), \quad \forall i\in[m], \quad \gamma\in(0,1)
\end{aligned}
\end{equation}
where the summation of fair constraints is defined as \textit{long-term constraints} in \cite{OGDLC-2012-JMLR}. 
The big $\mathcal{O}$ notation in the constraint is to bound the total violation of fairness sublinear in $T$.
The main drawback of using the metric of static regret is that it is only meaningful
for stationary environments, and low static regret cannot imply a good performance in changing environments since the time-invariant comparator in Eq.(\ref{eq:static-regret}) may behave badly \cite{zhang-2020-AISTATS}.

\section{Methodology}
\label{sec:methodology}
    \subsection{Settings and Problem Formulation}
\label{sec:settings and problem formulation}
To address the limitation of changing environments in online learning, \textit{adaptive regret} (AR) based on \cite{Hazan-2007-ARegret} is defined as the maximum static regret over any contiguous intervals. However, AR does not respect short intervals well. To this end, \textit{strongly adaptive regret} (SAR) \cite{daniely15-2015-ICML} is proposed to improve AR, which emphasizes the dependence on lengths of intervals, and it takes the form that
\begin{equation}
\begin{aligned}
\label{eq:SAR}
    \text{SAR}(T,\tau) = \max_{[s,s+\tau-1]\subseteq[T]} \Big ( \sum_{t=s}^{s+\tau-1} f_t(\boldsymbol{\theta}_t,\mathcal{D}_t) - \min_{\boldsymbol{\theta}\in\Theta}\sum_{t=s}^{s+\tau-1} f_t(\boldsymbol{\theta},\mathcal{D}_t) \Big )
\end{aligned}
\end{equation}
where $\tau$ indicates the length of time interval. In SAR, the learner is competing with changing comparators, as $\boldsymbol{\theta}$ varies with $s$ over $[s, s+\tau-1]$. 

In this paper, we consider the online meta-learning setting similar in \cite{Finn-ICML-2019,zhao-KDD-2021,zhao2021fairness}, but tasks are sampled from heterogeneous distributions. Instead of static regret, we define a novel regret FairSAR in Eq.(\ref{eq:ourRegret}). Let $\{\boldsymbol{\theta}_t\}_{t=1}^T$ be the sequence of model parameters generated in the \textit{Step 1} of the learning protocol (see Section \ref{sec:fairness-aware online learning}).
The goal of our problem is to minimize FairSAR under the long-term fair constraints:
\begin{equation}
\begin{aligned}
\label{eq:ourRegret}
    \text{FairSAR}(T,\tau)=&\max_{[s,s+\tau-1]\subseteq[T]} \bigg( \sum_{t=s}^{s+\tau-1} f_t\Big(\mathcal{G}_t(\boldsymbol{\theta}_t,\mathcal{D}_t^S),\mathcal{D}_t^V\Big) - \min_{\boldsymbol{\theta}\in\Theta}\sum_{t=s}^{s+\tau-1} f_t\Big(\mathcal{G}_t(\boldsymbol{\theta},\mathcal{D}_t^S),\mathcal{D}_t^V\Big) \bigg) \\
    \text{subject to} \quad &\max_{[s,s+\tau-1]\subseteq[T]} \bigg (\sum_{t=s}^{s+\tau-1} g_i\Big(\mathcal{G}_t(\boldsymbol{\theta}_t,\mathcal{D}_t^S),\mathcal{D}_t^V\Big) \bigg)\leq \mathcal{O}(T^\gamma), \quad\forall i \in[m] 
\end{aligned}
\end{equation}
where $\gamma\in(0,1)$. 
$\mathcal{D}_t^S, \mathcal{D}^{V}_t\subset\mathcal{D}_t$ are the support and validation set.
$\mathcal{G}_t(\cdot)$ is the base learner which corresponds to one or multiple gradient steps \cite{Finn-ICML-2017-(MAML)}.
Different from traditional online learning settings, the long-term constraint violation $g(\cdot): \mathcal{B}\times\mathbb{R}^d\rightarrow\mathbb{R}$ is satisfied. To facilitate our analysis, $\boldsymbol{\theta}_t$ is originally chosen from its domain
$\Theta=\{\boldsymbol{\theta}\in\mathbb{R}^d:g_i(\boldsymbol{\theta},\mathcal{D}_t)\leq 0, \forall i\in[m]\}$. A projection operator is hence typically applied to the updated variables to make them feasible \cite{OGDLC-2012-JMLR,GenOLC-2018-NeurIPS,AdpOLC-2016-ICML}.
To lower the computational complexity and accelerate the online processing speed, we relax the domain $\Theta$ to $\mathcal{B}$, where $\Theta\subseteq\mathcal{B}=S\mathbb{K}$ with $\mathbb{K}$ being the unit $\ell_2$ ball centered at the origin, and $S=\max\{r>0: r=||\boldsymbol{\theta}_1-\boldsymbol{\theta}_2||_2, \forall \boldsymbol{\theta}_1,\boldsymbol{\theta}_2\in\Theta \}$. 

In the protocol stated in Section \ref{sec:fairness-aware online learning}, the key step (\textit{Step 1}) is to find a good parameter $\boldsymbol{\theta}_t$ at each time $t$. In the following subsections, we first introduce three types of intervals where each interval combines a list of tasks (Section \ref{sec:intervals}); then, for each interval, a learning process (an expert) is proposed to output an interval-level model parameter (Section \ref{sec:experts}); finally, $\boldsymbol{\theta}_t$ is estimated using a meta-algorithm in which it combines weighted interval-level model parameters (Section \ref{sec:bi-level}). 

\begin{figure}[t!]
    \centering
    \includegraphics[width=0.5\linewidth]{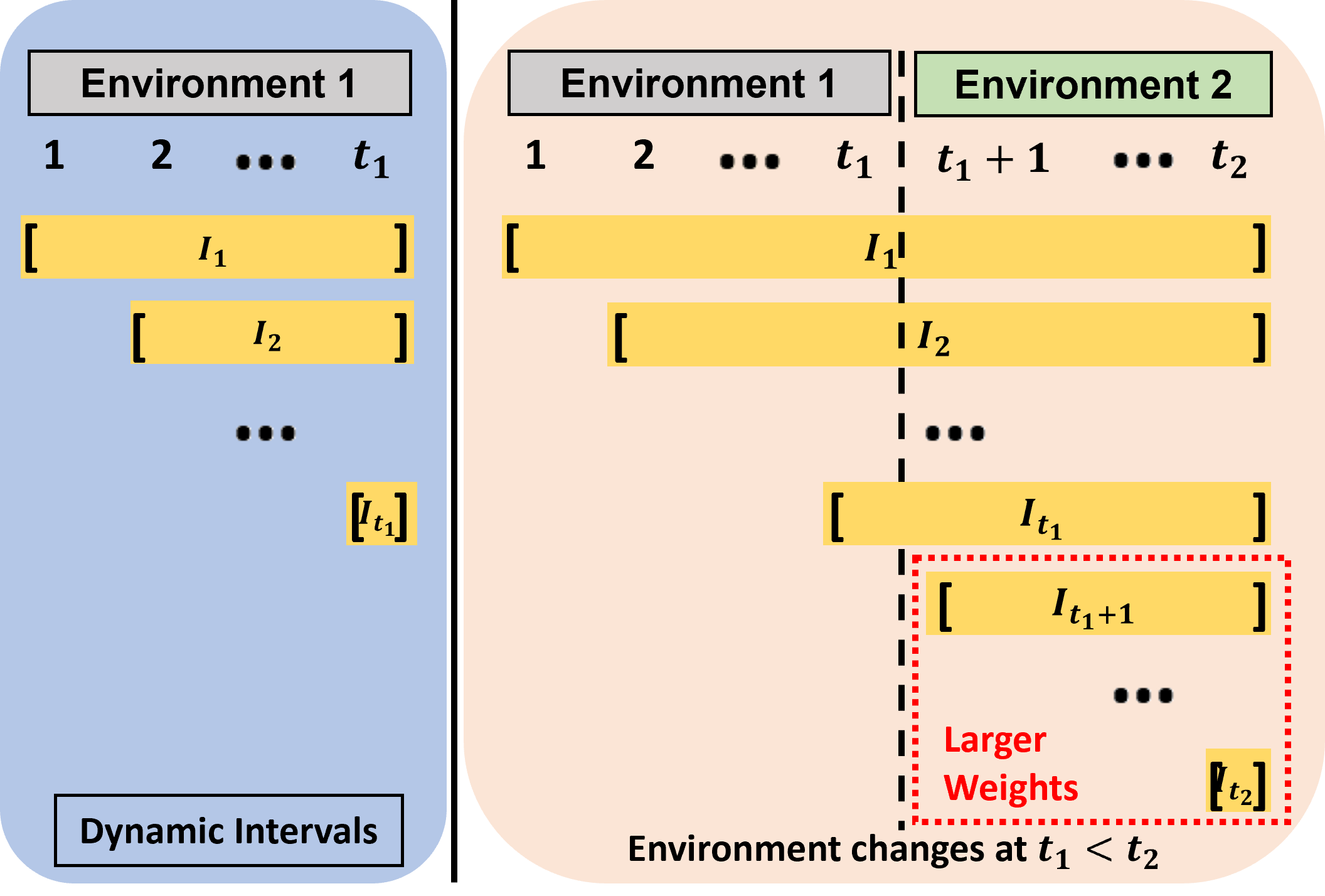}
    \caption{An illustration of adapting to changing environments using dynamic intervals (DI). \textbf{(Left)} At time $t_1\in[T]$, a number of intervals $\{I_1,\cdots,I_{t_1}\}$ are selected from the interval set $\mathcal{I}_{DI}$. 
    \textbf{(Right)} At time $t_2>t_1$, a different interval set are selected. Assume that the environment changes at $t_1$, to adapt to the change quickly, larger weights are given to the outputs through interval-level experts, where such outputs are not based on intervals prior to $t_1$.}
    \label{fig:DIandExamples}
\end{figure}

\subsection{Intervals}
\label{sec:intervals}
In Eq.(\ref{eq:ourRegret}), \sysnameregret{} evaluates the learner's performance on each time interval, and it is the maximum regret over any contiguous intervals. 
This subsection introduces three alternative interval sets to adapt to changing environments: dynamic intervals (DI) and two geometric covering-based intervals (AGC and DGC). Each interval in an interval set refers to a range of time indices associated with a collection of data batches, as data batches arrive one after another over time. 
Inspired by \textit{learning with expert advice} \cite{Jun-2017-AISTATS}, each interval is built upon a learning process, defined as an \textit{expert}, and each expert updates model parameters via $\mathcal{G}$ and outputs interval-level parameters with respect to a specific interval. Details of the interval-level learning are given in Section \ref{sec:interval-level learning}.

\subsubsection{Dynamic Intervals (DI)}
A heuristic method in designing an effective online learning algorithm for changing environments is to initiate a set of intervals $\mathcal{I}_{DI}$ dynamically, where 
\begin{equation}
\begin{aligned}
\label{eq:DI}
    \mathcal{I}_{DI} = \bigcup_{k\in[T]}\mathcal{I}_k \quad \text{where} \quad \mathcal{I}_k=\{I_k|I_k=[k,q],\forall q\in\{k,\cdots,T\}\}
\end{aligned}
\end{equation}
An interval $I_k\in\mathcal{I}_{DI}$ refers to a collection of time indices $\{k,\cdots,q\}$ associated with corresponding data batches $\{\mathcal{D}_i\}_{i=k}^q$.

Furthermore, at each time $t$, we introduce a target set $\mathcal{C}_t^{DI}\subset\mathcal{I}_{DI}$ which includes a set of intervals in $\mathcal{I}_{DI}$.
\begin{align}
\label{eq:C_DI}
    \mathcal{C}_t^{DI}= \{I_i|I_i=[i,t], I_i\in\mathcal{I}_{DI}, \forall i\in[t]\}
\end{align}
$\mathcal{C}_t^{DI}$ dynamically selects a subset of intervals from $\mathcal{I}_{DI}$. An example is illustrated in Figure \ref{fig:DIandExamples}. At time $t_1$, the target set $\mathcal{C}_{t_1}^{DI}$ selects $t_1$ intervals where $\mathcal{C}_{t_1}^{DI}=\{[1,t_1], [2,t_1],\cdots,[t_1,t_1]\}$. Similarly, when at time $t_2$ where $t_2>t_1$, $\mathcal{C}_{t_2}^{DI}=\{[1,t_2], [2,t_2],\cdots,[t_1,t_2],\cdots,[t_2,t_2]\}$.

To adapt to changing environments, at each time, a number of experts are initiated based on intervals selected in the target set. At time $t_2$, each expert corresponding to an interval $I_j\in\mathcal{C}_{t_2}^{DI}$, where $I_j=[j,t_2], \forall j\in[t_2]$, takes the parameter $\boldsymbol{\theta}_{t_2-1}$ as well as its corresponding dataset $\{\mathcal{D}_i\}_{i=j}^{t_2}$ as input. 
Each expert independently gives an interval-level solution $\boldsymbol{\theta}_{t_2,I_j}$ on $I_j$. 
A good $\boldsymbol{\theta}_{t_2}$ is therefore achieved at time $t_2$ by further combining the decisions $\{\boldsymbol{\theta}_{t_2,I_j}\}_{j=1}^{t_2}$ through weighted average. 
More details are stated in Section \ref{sec:experts}.

The key idea of constructing dynamic intervals is that at time $t_2$, some of the outputs $\{\boldsymbol{\theta}_{t_2,I_j}\}_{j=t_1+1}^{t_2}$ on intervals $\{I_{t_1+1},\cdots,I_{t_2}\}$ are not based on any data prior to time $t_1$ where $t_1<t_2$, so that if the environment changes at $t_1$, those outputs may be given a larger weight by the meta-algorithm, allowing it to adapt more quickly to the change. 

A main drawback with the construction of dynamic intervals, however, is a factor of $t$ increase in the time complexity. The number of intervals and learning processes increases linearly in time.
To avoid this, we reduce the complexity to $\mathcal{O}(\log t)$ by restarting algorithms on a designed set of geometric covering intervals, \textit{i.e.,} AGC and DGC intervals in Section \ref{sec:agc intervals and experts} and \ref{sec:dgc intervals and experts}, respectively.

\begin{table}[t!]
    \centering
    \caption{Comparison of key different interval settings.}
    \begin{tabular}{l|c|c|c}
    \toprule
         & DI & AGC & DGC\\
        \cmidrule(lr){1-4}
        Require $T$ in advance? & No & Yes & No\\
        Interval lengths in $\mathcal{I}_k$ & $[T-k+1]$ & $2^k$ & $2^k$\\
        $\#$ total experts $|\mathcal{U}_t|$ at time $t$ & $t$ & $\lfloor\log_2^T\rfloor$  &  $\lfloor\log_2^t\rfloor+1$ \\
        $\#$ active experts $|\mathcal{A}_t|$ at time $t$ & $t$ & $\leq\lfloor\log_2^T\rfloor$ & $\leq\lfloor\log_2^t\rfloor+1$\\
        $\#$ sleeping experts $|\mathcal{S}_t|$ at time $t$ & $0$ & $<\lfloor\log_2^T\rfloor$ & $<\lfloor\log_2^t\rfloor+1$\\
        Complexity & $\mathcal{O}(t)$ & $\mathcal{O}(\log t)$ & $\mathcal{O}(\log t)$\\
    \bottomrule
    \end{tabular}
    \label{tab:intervals_comp}
\end{table}

\begin{figure*}[t!]
  \begin{center}
    \includegraphics[width=0.9\textwidth]{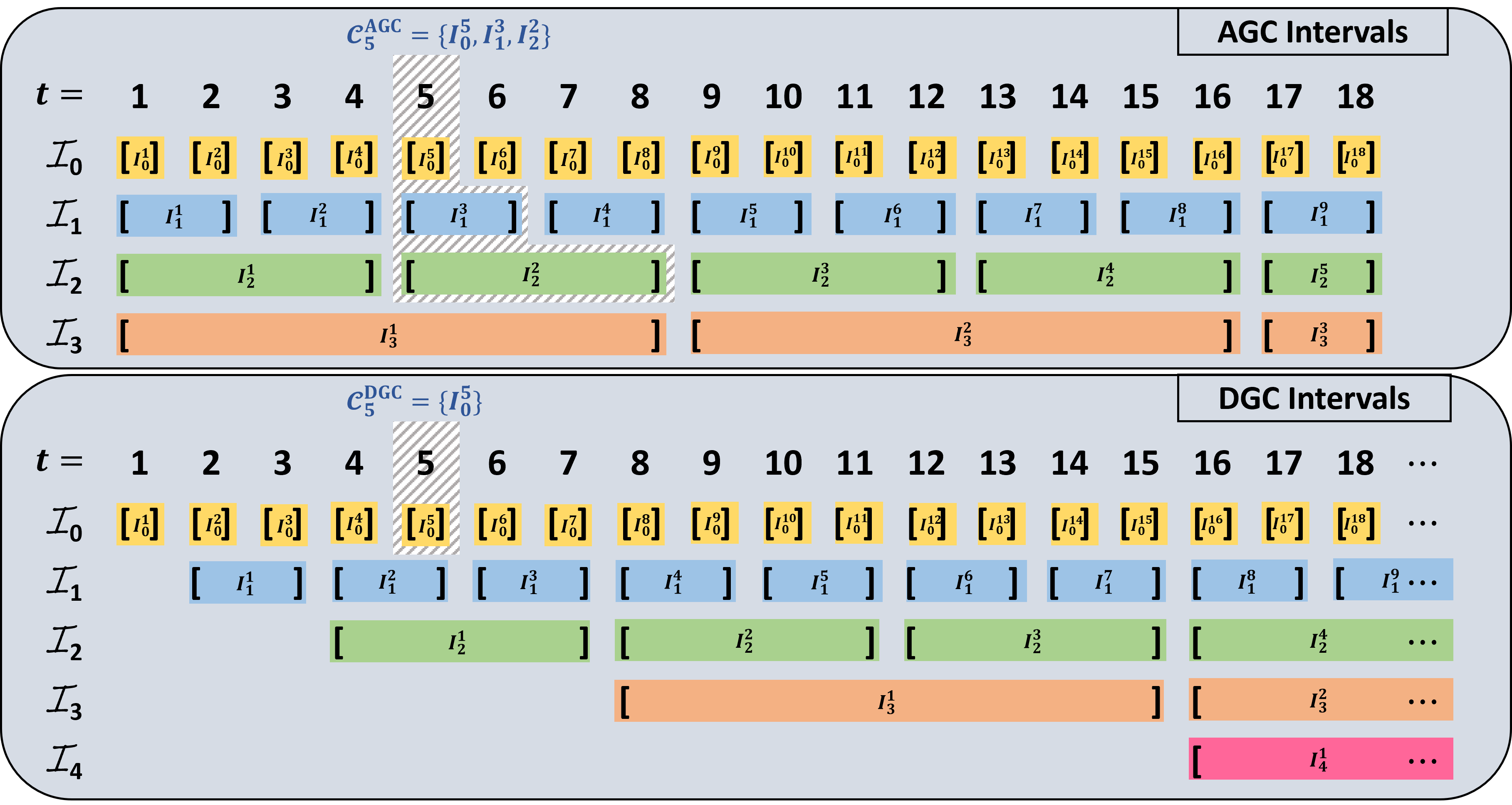}
  \end{center}
  \caption{\textbf{(Upper)} A graphical illustration of AGC intervals (base=2) with $T=18$. The interval set $\mathcal{I}_{AGC}$ consists of 4 subsets $\{\mathcal{I}_0,\mathcal{I}_1,\mathcal{I}_2,\mathcal{I}_3\}$ and each contains different numbers of intervals with fixed length. Intervals covered by shadow is an example of target subset $\mathcal{C}_5^{AGC}$ when $t=5$. \textbf{(Lower)} An illustration of DGC intervals (base=2) with $T$ is unknown in advance. The interval subsets $\{\mathcal{I}_0,\mathcal{I}_1,\cdots\}$ increase as $t$ increases. Similar to the setting of AGC intervals, when $t=5$, the target set $\mathcal{C}_5^{DGC}$ only includes one interval $I_0^5$.}
\label{fig:AGCandExamples}
\end{figure*}

\subsubsection{Adaptive Geometric Covering (AGC) Intervals}
\label{sec:agc intervals and experts}
Inspired by the seminal work of SAR \cite{daniely15-2015-ICML}, given the total number of time $T$, we improve dynamic intervals by constructing a number of interval sets where each set $\mathcal{I}_k\subset\mathcal{I}_{AGC}$ contains various intervals with fixed lengths. We name them adaptive geometric covering (AGC) intervals.
A set of contiguous AGC intervals $\mathcal{I}_{AGC}$ are defined as
\begin{equation}
\begin{aligned}
\label{eq:AGC}
    \mathcal{I}_{AGC} = \bigcup_{k\in[\lfloor\log_2^T\rfloor-1]\cup \{0\}} \mathcal{I}_k \quad \text{where} \quad \forall k, \: \mathcal{I}_k=\{I_k^i|I_k^i = [(i-1)\cdot 2^k+1, \min{\{T,i\cdot 2^k\}}], \forall i\in\mathbb{N}\}
\end{aligned}
\end{equation}
An example with $T=18$ is given in Figure \ref{fig:AGCandExamples} to illustrate the composition of AGC intervals. 
With selecting $2$ as the $\log$ base in Eq.(\ref{eq:AGC}), intervals are hence decomposed into $\lfloor\log_2^{18}\rfloor=4$ subsets (\textit{i.e.,} $\mathcal{I}_0, \mathcal{I}_1, \mathcal{I}_2$, and $\mathcal{I}_3$) with fixed lengths of $2^0=1$, $2^1=2$, $2^2=4$, and $2^3=8$.
Notice that the $\log$ base equals $2$ is not required, but a larger base number leads to fewer interval subsets. AGC intervals can be considered a special case of a more general set of intervals, and they efficiently reduce the time complexity to $\mathcal{O}(\log t)$.  

Similar to DI, a target set $\mathcal{C}_{t}^{AGC}\subset\mathcal{I}_{AGC}$ including a set of intervals starting from $t$ are selected from $\mathcal{I}_{AGC}$ at each time:
\begin{align}
\label{eq:C_AGC}
    \mathcal{C}_t^{AGC}=\{I|I\in \mathcal{I}_{AGC}, t\in I, (t-1)\notin I \}
\end{align}
As shown in Figure \ref{fig:AGCandExamples}, given $T=18$ and when $t=5$, the target set $\mathcal{C}_5^{AGC}$ contains three intervals, $[5,5]$, $[5,6]$ and $[5,8]$, where each initiates at $t=5$ with interval lengths $1$, $2$, and $4$, respectively.

\subsubsection{Dynamic Geometric Covering (DGC) Intervals}
\label{sec:dgc intervals and experts}
Although the setting of AGC intervals efficiently reduces the complexity, one limitation is that the total number of times $T$ needs to be known and fixed in advance. 
However, this assumption does not always hold. This leads to the number of interval sets (\textit{i.e.} $\lfloor\log_2^T\rfloor$) being unchanged in AGC, as $k\in\{0,\cdots,\lfloor\log_2^T\rfloor-1\}$.

To tackle this limitation, we alternatively propose another type of interval set, namely dynamic geometric covering (DGC) intervals, $\mathcal{I}_{DGC}$. 
\begin{equation}
\begin{aligned}
\label{eq:DGC}
    \mathcal{I}_{DGC} = \bigcup_{k\in\{0,1,2,\cdots\}} \mathcal{I}_k \quad \text{where} \quad \forall k, \: \mathcal{I}_k=\{I_k^i|I_k^i=[i\cdot2^k,(i+1)\cdot2^k-1],\forall i\in\mathbb{N}\}
\end{aligned}
\end{equation}
Figure \ref{fig:AGCandExamples} illustrates the difference between the settings of DGC and AGC intervals.
Since the total number of times is unknown in advance, the number of interval sets, $\lfloor\log_2 t\rfloor+1$, in DGC increases as $t$ becomes larger. 
For each interval set $\mathcal{I}_k$, its first interval $I_k^{1}\in\mathcal{I}_k$ initializes at the $2^k$-th time, and each interval $I_k^i\in\mathcal{I}_k$ holds the same length of $2^k$. Furthermore, the setting of the target set $\mathcal{C}_t^{DGC}\subset\mathcal{I}_{DGC}$ at time $t$ is the same as the one in AGC, referring to Eq.(\ref{eq:C_AGC}). As indicated in Figure \ref{fig:AGCandExamples}, in contrast to AGC, in DGC, an additional interval set $\mathcal{I}_4$ is initialized at time $t=16$ with an interval length of $2^4=16$. Similarly, when $t=5$, the target set $\mathcal{C}_5^{DGC}$ includes the interval $[5,5]$ only, as there is one interval in $\mathcal{I}_{DGC}$ that starts at time $5$.

A brief comparison between different interval settings introduced in this section is listed in Table \ref{tab:intervals_comp}.

\subsection{Learning Experts}
\label{sec:experts}
\subsubsection{The Interval-level Learning within An Expert}
\label{sec:interval-level learning}
As we mentioned at the beginning of Section \ref{sec:intervals}, each interval is built upon a learning process, defined as an expert. At time $t$, an expert $E$ is a learning algorithm $\mathcal{G}:\Theta\times\mathbb{R}^d\rightarrow\Theta$ (\textit{a.k.a.,} a base learner, such as one or multiple gradient steps \cite{Finn-ICML-2017-(MAML)}) within an interval that inputs parameters $\boldsymbol{\theta}_{t-1}$ and outputs interval-level parameters $\boldsymbol{\theta}_{t,I}$ specific to the interval $I$.
The \textit{interval-level} parameter update for an expert $E$ on interval $I$ at time $t$ is defined
\begin{equation}
\small
\begin{aligned}
\label{eq:inner-problem}
    \boldsymbol{\theta}_{t,I} := \mathcal{G}_{t}(\boldsymbol{\theta},\mathcal{D}^{S}_{t,I}) = \arg\min_{\boldsymbol{\theta}} \: f_{t}(\boldsymbol{\theta},\mathcal{D}^{S}_{t,I}) \quad
    \text{subject to} \quad g_{i}(\boldsymbol{\theta},\mathcal{D}^{S}_{t,I})\leq 0, \forall i \in [m]  
\end{aligned}
\end{equation}
where the loss function $f_{t}(\cdot)$ and the fairness function $g_{i}(\cdot)$ are defined based on the support set $\mathcal{D}_{t,I}^S\subset\mathcal{D}_{t,I}$ associated with $E$.

\begin{figure*}
    \centering
    \includegraphics[width=\linewidth]{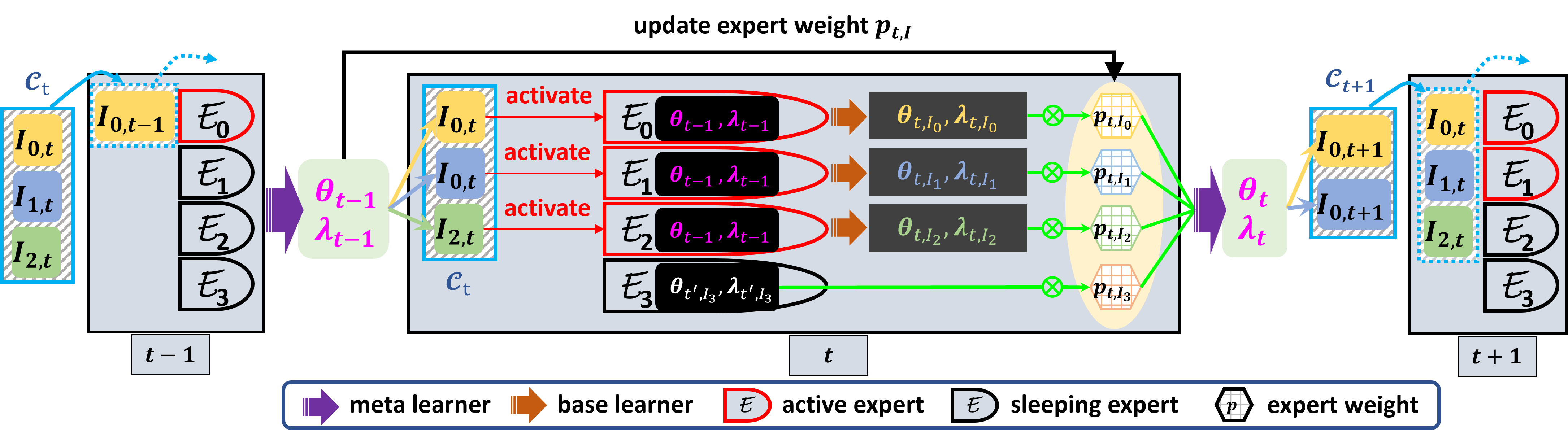}
    \caption{An overview of \sysname{} with AGC or DGC intervals to determine model parameter pair at each round. A target set (shadowed) of intervals is initially selected and is later used to activate corresponding experts. Each active expert runs through a base learner for the interval-level parameter-pair adaption, and its weight is updated. The meta-level parameter pair is finally attained through the meta-learner by combining the weighted actions of all experts.}
    \label{fig:overview}
\end{figure*}

\subsubsection{Active and Sleeping Experts}
\label{sec:target-set and active experts}
Inspired by \textit{learning with expert advice} problems \cite{Jun-2017-AISTATS}, we dynamically construct a set of experts $\mathcal{U}_t=\{E_k\}_{k=0}^{|\mathcal{U}_t|}$ at each time $t$. Recall that we introduce three types of interval sets in Section \ref{sec:intervals}. The number of total experts for each setting is various, where at time $t$
\begin{itemize}
    \item for dynamic intervals, $|\mathcal{U}^{DI}_t|=t$. The number of experts increases by $1$ at each time.
    \item for AGC intervals, $|\mathcal{U}^{AGC}_t|=\lfloor\log_2 T\rfloor$. The number of experts is unchanged at different times, resulting from $T$ being known in advance and fixed.
    \item for DGC intervals, $|\mathcal{U}^{DGC}_t|=\lfloor\log_2 t\rfloor+1$. The number of experts slowly increases as needed, without known $T$ in advance.
\end{itemize}

Furthermore, to adapt to changing environments efficiently, all experts are dynamically partitioned into active and sleeping (or inactive) experts at each time $t$, denoted $\mathcal{A}_t\subseteq\mathcal{U}_t$ and $\mathcal{S}_t=\mathcal{U}_t\backslash\mathcal{A}_t$, respectively. As indicated in Section \ref{sec:intervals}, a target set $\mathcal{C}_t$, for all types of intervals $\mathcal{I}_{DI}$, $\mathcal{I}_{AGC}$, and $\mathcal{I}_{DGC}$, is subsetted from the interval set. Active experts are experts corresponding to intervals in the target sets, wherein active experts update model parameters at interval-level using Eq.(\ref{eq:inner-problem}). 
For sleeping experts, as no corresponding intervals are selected in the target set at time $t$, their interval-level model parameters are not updated and remain at the last update. Similarly, the number of active/sleeping experts varies by applying different types of interval sets, where at time $t$
\begin{itemize}
    \item for dynamic intervals, all experts are active experts $\mathcal{A}_t^{DI}=\{E_k\}_{k=1}^t$ and the number of sleeping experts is zero, $\mathcal{S}_t^{DI}=\emptyset$.
    \item for AGC and DGC intervals, the number of active experts is the cardinality of the selected target set. 
    As the example shown in Figure \ref{fig:AGCandExamples}, when $t=5$, active experts are $\mathcal{A}_5^{AGC}=\{E_0, E_1, E_2\}$ and $\mathcal{A}_5^{DGC}=\{E_0\}$, and sleeping experts are $\mathcal{S}_5^{AGC}=\{E_3\}$ and $\mathcal{S}_5^{DGC}=\{E_1, E_2\}$ (experts $E_3$ and $E_4$ are not initialized until $t=8$ and $t=16$, respectively).
\end{itemize}

\subsection{Learning Dynamically for Bi-Level Adaptation}
\label{sec:bi-level}
Recall that in the protocol of fairness-aware online learning (Section \ref{sec:fairness-aware online learning}), the main goal for the learner is to sequentially decide on the model parameter $\boldsymbol{\theta}_t$ that performs well on the loss sequence and the long-term fair constraints. Crucially, inspired by \cite{Finn-ICML-2017-(MAML)}, we consider a setting where at each round $t$ the learner can perform a number of expert-specific updates at an interval level in the active set $\mathcal{A}_t$.

As specified in Eq.(\ref{eq:ourRegret}), model parameters at each round $t$ are determined by formulating problems with a nested bi-level adaptation process: interval-level and meta-level. Each level corresponds to a sub-learner, \textit{i.e.} base and meta learner, respectively, described in Figure \ref{fig:overview}.
The problem of learning a meta-level parameter $\boldsymbol{\theta}_t$ is embedded with the optimization problem of finding interval-level parameters $\boldsymbol{\theta}_{t,I}$ in Eq.(\ref{eq:inner-problem}). For experts in the sleeping set $\mathcal{S}_t$, the base learner is not applied.
The \textit{meta-level} problem takes the form in Eq.(\ref{eq:outer-problem}). 
\begin{equation}
\begin{aligned}
\label{eq:outer-problem}
    \min_{\boldsymbol{\theta}\in\mathcal{B}} &\sum_{E_k\in\mathcal{A}_t} p_{t,I_k} \cdot f_{t}(\mathcal{G}_{t}(\boldsymbol{\theta},\mathcal{D}_{t,I_k}^S),\mathcal{D}_{t,I_k}^Q) + \sum_{E_k\in\mathcal{S}_t} p_{t,I_k}\cdot f_{t}(\boldsymbol{\theta}_{t',I_k},\mathcal{D}_{t,I_k}^Q) \\
    \text{subject to} &\sum_{E_k\in\mathcal{A}_t} p_{t,I_k} \cdot g_i(\mathcal{G}_{t}(\boldsymbol{\theta},\mathcal{D}_{t,I_k}^S),\mathcal{D}_{t,I_k}^Q) + \sum_{E_k\in\mathcal{S}_t} p_{t,I_k}\cdot g_i(\boldsymbol{\theta}_{t',I_k},\mathcal{D}_{t,I_k}^Q)\leq 0
\end{aligned}
\end{equation}
where $p_{t,I_k}\geq 0$ is the expert weight of $E_k$ at $t$. $\mathcal{D}_{t,I_k}^Q\subset\mathcal{D}_{t,I_k}$ is the query set where $\mathcal{D}_{t,I_k}^Q\cap\mathcal{D}_{t,I_k}^S=\emptyset$.
$\boldsymbol{\theta}_{t',I_k}$ is the interval-level model parameter for an sleeping expert $E_k\in\mathcal{S}_t$ where the round index $t'<t$ represents the last time this expert was activated.

In the following section, we introduce our proposed algorithm \sysname{}. In stead of optimizing primal parameters only, it efficiently deals with the bi-level optimization problem of Eq.(\ref{eq:inner-problem})(\ref{eq:outer-problem}) by approximating a sequence of pairs of primal-dual meta parameters $\{(\boldsymbol{\theta}_t,\boldsymbol{\lambda}_t)\}_{t=1}^T$ where the pair respectively responds for adjusting accuracy and fairness level.

\begin{algorithm}[!t]
\caption{\sysname{}}
\label{alg:ouralgorithm}
\begin{algorithmic}[1]
\State Initialize meta-parameters pair $(\boldsymbol{\theta}_0, \boldsymbol{\lambda}_0)$, where $\boldsymbol{\theta}_0$ is the center of $\mathcal{B}$ and $\boldsymbol{\lambda}_0\in\mathbb{R}_+^m$ is randomly chosen
\State \multiline{%
    Create an object-oriented expert $E$ containing interval level parameter pair $(\boldsymbol{\theta}_I,\boldsymbol{\lambda}_I)$, learning rate $\eta_I$, constants $R_I, C_I$}
\For{each $t\in[T]$}
    \State \multiline{%
        Sample $\mathcal{D}^{V}_t\subset\mathcal{D}_t$ and record the performance of $\boldsymbol{\theta}_{t-1}$}
    \State Subset $\mathcal{C}_t$ from $\mathcal{I}$ using either Eq.(\ref{eq:C_DI}) or Eq.(\ref{eq:C_AGC}).
    \For{each interval $I_k\in\mathcal{C}_t$}
        \State \textsc{ActivateExperts}(DI or AGC or DGC)
    \EndFor
    \For{each expert $E_k\in\mathcal{U}_t$}
        \State Update $p_{t,I_k}$ using $R_{t,I_k}, C_{t,I_k}$ in Eq.(\ref{weight})
    \EndFor
    \For{$n=1,...,N_{meta}$ steps}
        \For{each active expert $E_k\in\mathcal{A}_t$}    
            \State Sample support set $\mathcal{D}_{t,I_k}^{S}\subset\mathcal{D}_{t,I_k}$
            \State \multiline{%
                Adapt interval-level primal and dual variables with $\mathcal{D}_{t,I_k}^{S}$ using Eq.(\ref{eq:inner-pd-update})}
        \EndFor
        \For{each expert $E_k\in\mathcal{U}_t$}
            \State Sample query set $\mathcal{D}_{t,I_k}^{Q}\subset\mathcal{D}_{t,I_k}$
        \EndFor
        \State \multiline{%
            Update meta-level primal and dual variables with $D^Q_{t,I_k}$ using Eq.(\ref{eq:outer-pd-update})}
    \EndFor
    \For{each expert $E_k\in\mathcal{U}_t$}
        \State \multiline{%
            $R_{t+1,I_k} = R_{t,I_k} + \mathcal{F}_{t,I_k}(\boldsymbol{\theta}_t,\boldsymbol{\lambda}_t) - \mathcal{F}_{t,I_k}(\boldsymbol{\theta}_{t,I_k},\boldsymbol{\lambda}_{t,I_k})$}
        \State \multiline{%
            $C_{t+1,I_k} = C_{t,I_k} + \Big | \mathcal{F}_{t,I_k}(\boldsymbol{\theta}_t,\boldsymbol{\lambda}_t) - \mathcal{F}_{t,I_k}(\boldsymbol{\theta}_{t,I_k},\boldsymbol{\lambda}_{t,I_k})\Big |$}
    \EndFor
\EndFor
\end{algorithmic}
\end{algorithm}

\subsection{An Efficient Algorithm: \sysname{}}
To find a good model parameter pair $(\boldsymbol{\theta}_t, \boldsymbol{\lambda}_t)$ at each time, an efficient working flow is proposed in Algorithm \ref{alg:ouralgorithm}.
Inspired by dynamic programming and expert-tracking \cite{luo-2015-achieving} techniques, experts at each time are recursively divided into active and sleeping ones. Model parameters in active experts are locally updated, but those in sleeping experts are directly inherited from the previous time.
Specifically, at the beginning of $t$, a target set $C_t$ containing intervals is used to activate a subset of experts in $\mathcal{U}_t$. For each active expert $E_k$ in $\mathcal{A}_t$, an interval-level algorithm takes the meta-level solution $(\boldsymbol{\theta}_{t-1}, \boldsymbol{\lambda}_{t-1})$ and outputs an expert-specific solution pair $(\boldsymbol{\theta}_{t,I_k}, \boldsymbol{\lambda}_{t,I_k})$. 
Finally, through the meta-learner, we combine the weighted solutions of all experts and move to the next time.

\begin{algorithm}[!t]
\caption{\sysname{} Subroutines}
\label{alg:subroutines}
\begin{algorithmic}[1]
\State \textbf{Require}: Choose one of the three settings of intervals.
\Procedure{\textsc{ActivateExperts}}{DI}
    \State \multiline{%
        Activate expert $E_{k}$ by letting $\eta_{t,I_k}=S/(G\sqrt{|I_k|})$, $(\boldsymbol{\theta}_{t,I_k},\boldsymbol{\lambda}_{t,I_k})\leftarrow(\boldsymbol{\theta}_{t-1},\boldsymbol{\lambda}_{t-1})$} 
    \If{Identify an $E_j\in\mathcal{U}_t$ where $|I_j|=|I_k|-1$}
        \State Update $R_{t,I_k}\leftarrow R_{t-1,I_j}$ and $C_{t,I_k}\leftarrow C_{t-1,I_j}$
        \State Replace $E_j$ with $E_{k}$ in $\mathcal{U}_t$ 
    \Else
        \State Set $R_{t,I_k}=0$ and $C_{t,I_k}=0$.
        \State Add the $E_k$ to $\mathcal{U}_t$
    \EndIf
\EndProcedure     
\Procedure{\textsc{ActivateExperts}}{AGC}
    \State \multiline{%
        Activate expert $E_{k}$ by letting $\eta_{t,I_k}=S/(G\sqrt{|I_k|})$, $(\boldsymbol{\theta}_{t,I_k},\boldsymbol{\lambda}_{t,I_k})\leftarrow(\boldsymbol{\theta}_{t-1},\boldsymbol{\lambda}_{t-1})$}
    \State Identify an $E_j\in\mathcal{U}_t$ where $|I_j|=|I_k|$
    \State Update $R_{t,I_k}\leftarrow R_{t-1,I_j}$ and $C_{t,I_k}\leftarrow C_{t-1,I_j}$
    \State Replace $E_j$ with $E_{k}$ in $\mathcal{U}_t$ 
\EndProcedure   
\Procedure{\textsc{ActivateExperts}}{DGC}
    \State \multiline{%
        Activate expert $E_{k}$ by letting $\eta_{t,I_k}=S/(G\sqrt{|I_k|})$, $(\boldsymbol{\theta}_{t,I_k},\boldsymbol{\lambda}_{t,I_k})\leftarrow(\boldsymbol{\theta}_{t-1},\boldsymbol{\lambda}_{t-1})$} 
    \If{Identify an $E_j\in\mathcal{U}_t$ where $|I_j|=|I_k|$}
        \State Update $R_{t,I_k}\leftarrow R_{t-1,I_j}$ and $C_{t,I_k}\leftarrow C_{t-1,I_j}$
        \State Replace $E_j$ with $E_{k}$ in $\mathcal{U}_t$ 
    \Else
        \State Set $R_{t,I_k}=0$ and $C_{t,I_k}=0$.
        \State Add the $E_k$ to $\mathcal{U}_t$
    \EndIf
\EndProcedure
\end{algorithmic}
\end{algorithm}

We explain the main steps in Algorithm \ref{alg:ouralgorithm} below. In Step 4, when a new task arrives at time $t$, a batch of data $\mathcal{D}_t^{V}$ is randomly sampled from $\mathcal{D}_t$ for validation purposes, and the performance on $\boldsymbol{\theta}_{t-1}$ achieved is recorded. 
A target set of intervals $\mathcal{C}_t$ is selected from $\mathcal{I}$ in Step 5. 
For each interval $I_k \in\mathcal{C}_t$ (Step 6-8), the corresponding expert $E_{t,I_k}$ is activated, according to a specific \textsc{ActivateExperts} procedure on the choice of interval sets indicated in the subroutines of \sysname{} in Algorithm \ref{alg:subroutines}.

We present three distinct expert activation procedures in Algorithm \ref{alg:subroutines}. For each active expert, we set adaptive stepsizes $\eta_{t,I}=S/(G\sqrt{|I_k|})$, where $S$ is the radius of the Euclidean ball $\mathcal{B}$, and there exists a constant $G>0$ that bounds the (sub)gradients of $f_t$ and $g_i$. Following the setting used in \cite{AdpOLC-2016-ICML}, empirically we set $S=\sqrt{1+2\epsilon}-1$ and $G=\max\{\sqrt{d}+S,\max_t\{||\boldsymbol{e}_{I_k}||_2, \boldsymbol{e}_{I_k}\in\mathcal{P}_t\} \}$, where $\boldsymbol{e}_{I_k}$ is the non-protected features lied in the interval $I_k$ and $d$ is its feature dimension. $\mathcal{P}_t$ is a set which includes all past intervals until time $t$. 
Specifically in DI and DGC, at some time $t$, new experts are initiated. We set the constants $R_{t,I_k}$ and $C_{t,I_k}$ to zeros that are further used to change the corresponding expert weight to adapt to changing environments.

In Steps 9-11 of Algorithm \ref{alg:ouralgorithm}, for all experts in $\mathcal{U}_t$, a following weight $p_{t,I_k}$ is estimated:
\begin{equation}
\begin{aligned}
\label{weight}
    p_{t,I_k} = \frac{w(R_{t,I_k}, C_{t,I_k})}{\sum_{E_k\in\mathcal{U}_t}w(R_{t,I_k}, C_{t,I_k})}
\end{aligned}
\end{equation}
Here, a weight function \cite{luo-2015-achieving} is defined 
as $w(R,C) = \frac{1}{2}\big(\Phi(R+1,C+1)-\Phi(R-1,C-1)\big)$, where $\Phi(R,C) = \exp([R]^2_+/3C)$
and $[r]_+=\max(0,r)$ and $\Phi(0,0)=1$.
In Steps 12-21, our \sysname{} responds to the bi-level adaptation stated in Eq.(\ref{eq:inner-problem}) and (\ref{eq:outer-problem}). Specifically, to solve the interval-level problem in Eq.(\ref{eq:inner-problem}), for each active expert $E_k$ in $\mathcal{A}_t$, we consider following Lagrangian function
\begin{equation}
\begin{aligned}
    \mathcal{F}_{t,I_k}(\boldsymbol{\theta}_{t-1},\boldsymbol{\lambda}_{t-1})
    =f_{t}(\boldsymbol{\theta}_{t-1},\mathcal{D}_{t,I_k}^S) + \sum_{i=1}^m \lambda_{t-1,i}\cdot g_{i}(\boldsymbol{\theta}_{t-1},\mathcal{D}_{t,I_k}^S)
\end{aligned}
\end{equation}
where the interval-level parameter pair for an active expert $E_k$ are initialized with the meta-level parameter $(\boldsymbol{\theta}_{t-1}, \boldsymbol{\lambda}_{t-1})$ . 
For optimization with simplicity, cumulative constraints in Eq.(\ref{eq:inner-problem}) are approximated with the summarized regularization. Interval-level parameters are updated through a base learner $\mathcal{G}_t(\cdot)$. One example for the learner is updating with one gradient step \cite{Finn-ICML-2017-(MAML)} using the pre-determined adaptive stepsize $\eta_{t,I_k}$. Notice that for multiple gradient steps, $\boldsymbol{\theta}_{t,I_k}$ and $\boldsymbol{\lambda}_{t,I_k}$ interplay each other for updating.
\begin{equation}
\begin{aligned}
\label{eq:inner-pd-update}
    \boldsymbol{\theta}_{t,I_k} = \boldsymbol{\theta}_{t-1} - \eta_{t,I_k}\nabla_{\boldsymbol{\theta}}\mathcal{F}_{t,I_k}(\boldsymbol{\theta}_{t-1},\boldsymbol{\lambda}_{t-1}); \quad
    \boldsymbol{\lambda}_{t,I_k} = \boldsymbol{\lambda}_{t-1} + \eta_{t,I_k}\nabla_{\boldsymbol{\lambda}}\mathcal{F}_{t,I_k}(\boldsymbol{\theta}_{t,I},\boldsymbol{\lambda}_{t-1})
\end{aligned}
\end{equation}
Next, to solve the meta-level problem in Eq.(\ref{eq:outer-problem}), we combine the actions of active experts together with sleeping experts. 
We consider the following augmented Lagrangian function and abuse the symbol $t'$ with $t$ in Eq.(\ref{eq:outer-problem}):
\begin{equation}
\begin{aligned}
\label{eq:L_t}
    \mathcal{L}_t(\boldsymbol{\theta}_{t,I_k},\boldsymbol{\lambda}_{t,I_k}) = \sum_{E_k\in\mathcal{U}_t} p_{t,I_k} \Bigg( f_{t}(\boldsymbol{\theta}_{t,I_k},\mathcal{D}_{t,I_k}^Q) 
    + \sum_{i=1}^m \Big(\lambda_{i,t,I_k}\cdot g_i(\boldsymbol{\theta}_{t,I_k},\mathcal{D}_{t,I_k}^Q) 
    - \frac{\delta(\eta_1+\eta_2)}{2}\lambda_{i,t,I_k}^2 \Big) \Bigg)
\end{aligned}
\end{equation}
where $\delta > 0$ is a constant determined by analysis. 
Note that the last augmented term on the dual variable is devised to prevent $\boldsymbol{\lambda}$ from being too large. The update rule for meta-level parameters follows:
\begin{equation}
\begin{aligned}
\label{eq:outer-pd-update}
    \boldsymbol{\theta}_{t} 
    = \prod_\mathcal{B}\Big(\boldsymbol{\theta}_{t-1} - \eta_1\nabla_{\boldsymbol{\theta}}\mathcal{L}_t(\boldsymbol{\theta}_{t,I_k},\boldsymbol{\lambda}_{t,I_k})\Big); \quad
    \boldsymbol{\lambda}_{t} = \Big[ \boldsymbol{\lambda}_{t-1} + \eta_2\nabla_{\boldsymbol{\lambda}}\mathcal{L}_t(\boldsymbol{\theta}_{t,I_k},\boldsymbol{\lambda}_{t,I_k}) \Big]_+
\end{aligned}
\end{equation}
where $\prod_\mathcal{B}$ is the projection operation to the relaxed domain $\mathcal{B}$ that is introduced in Section \ref{sec:settings and problem formulation}. This approximates the true desired projection with a simpler closed form.
Finally, in Steps 22-25, we update each expert's $R$ and $C$ values, determining the expert weight for the next time. The intuition of weight update is to re-adjust the difference between the meta-solution and the interval-level solution given by the expert.

\section{Analysis}
\label{sec:analysis}
    To analyze, we first make the following assumptions as in \cite{zhang-2020-AISTATS,OGDLC-2012-JMLR}. 
Examples where these assumptions hold include logistic regression and $L_2$ regression over a bounded domain. As for constraints, a family of fairness notions, such as DDP stated in Definition \ref{dbc definition}, are applicable as discussed in \cite{Lohaus-2020-ICML}. For simplicity, in this section we omit $\mathcal{D}$ used in $f_t(\cdot),\forall t$ and $g_i(\cdot),\forall i$.

\begin{assumption}[Convex domain]
\label{assmp1}
    The convex set $\Theta$ is non-empty, closed, bounded, and it is described by $m$ convex functions as $\Theta = \{\boldsymbol{\theta}:g_i(\boldsymbol{\theta})\leq 0, \forall i\in[m]\}$. The relaxed domain $\mathcal{B}$ (where $\Theta\subseteq\mathcal{B}$) contains the origin $\boldsymbol{0}$ and its diameter is bounded by $S$.
\end{assumption}

\begin{assumption}
Both the loss functions $f_t(\cdot), \forall t$ and constraint functions $g_i(\cdot), \forall i\in[m]$ satisfy the following assumptions 
\label{assmp2}
\begin{enumerate}[leftmargin=*]
    \item
    \begin{sloppypar}
    (Lipschitz Continuous) $\forall \boldsymbol{\theta}_1,\boldsymbol{\theta}_2\in\mathcal{B}$,  $||f_t(\boldsymbol{\theta}_1)-f_t(\boldsymbol{\theta}_2)||\leq L_f||\boldsymbol{\theta}_1-\boldsymbol{\theta}_2||, ||g_i(\boldsymbol{\theta}_1)-g_i(\boldsymbol{\theta}_2)||\leq L_g||\boldsymbol{\theta}_1-\boldsymbol{\theta}_2||$. 
    Let $G = \max\{L_f,L_g\}$, $F = \max_{t\in[T]}\max_{\boldsymbol{\theta}_1,\boldsymbol{\theta}_2\in\mathcal{B}} f_t(\boldsymbol{\theta}_1)-f_t(\boldsymbol{\theta}_2)\leq 2L_f S$, and $D = \max_{i\in[m]}\max_{\boldsymbol{\theta}\in\mathcal{B}}g_i(\boldsymbol{\theta})\leq L_g S$.
    
    \item (Lipschitz Gradient) $f_t(\boldsymbol{\theta}), \forall t$ are $\beta_f$-smooth and $g_i(\boldsymbol{\theta}),\forall i$ are $\beta_g$-smooth, that is, $\forall \boldsymbol{\theta}_1,\boldsymbol{\theta}_2\in\mathcal{B}$, $||\nabla f_t(\boldsymbol{\theta}_1)-\nabla f_t(\boldsymbol{\theta}_2)||\leq \beta_f||\boldsymbol{\theta}_1-\boldsymbol{\theta}_2||, ||\nabla g_i(\boldsymbol{\theta}_1)-\nabla g_i(\boldsymbol{\theta}_2)||\leq \beta_g||\boldsymbol{\theta}_1-\boldsymbol{\theta}_2||$.
    
    \item (Lipschitz Hessian) Twice-differentiable functions $f_t(\boldsymbol{\theta}), \forall t$ and $g_i(\boldsymbol{\theta}),\forall i$ have $\rho_f$ and $\rho_g$- Lipschitz Hessian, respectively. That is, $\forall \boldsymbol{\theta}_1-\boldsymbol{\theta}_2\in\mathcal{B}$, $||\nabla^2 f_t(\boldsymbol{\theta}_1)-\nabla^2 f_t(\boldsymbol{\theta}_2)||\leq \rho_f||\boldsymbol{\theta-\phi}||, ||\nabla^2 g_i(\boldsymbol{\theta}_1)-\nabla^2 g_i(\boldsymbol{\theta}_2)||\leq \rho_g||\boldsymbol{\theta}_1-\boldsymbol{\theta}_2||$.
    \end{sloppypar}
\end{enumerate}
\end{assumption}

\begin{assumption}[Strongly convexity]
\label{assmp3}
    \begin{sloppypar}
    Suppose $f_t(\boldsymbol{\theta}), \forall t$ and $g_i(\boldsymbol{\theta}),\forall i$ have strong convexity, that is, $\forall \boldsymbol{\theta}_1, \boldsymbol{\theta}_2\in\mathcal{B}$, $||\nabla f_t(\boldsymbol{\theta}_1)-\nabla f_t(\boldsymbol{\theta}_2)||\geq \mu_f||\boldsymbol{\theta}_1-\boldsymbol{\theta}_2||, ||\nabla g_i(\boldsymbol{\theta}_1)-\nabla g_i(\boldsymbol{\theta}_2)||\geq \mu_g||\boldsymbol{\theta}_1-\boldsymbol{\theta}_2||$.
    \end{sloppypar}
\end{assumption}

Under the above assumptions, we first state the key Theorem \ref{main theorem} that the proposed \sysname{} enjoys a sub-linear guarantee for both regret and long-term fairness constraints in the long run for Algorithm \ref{alg:ouralgorithm}. 

\begin{theorem}
\label{main theorem}
Set $\boldsymbol{\theta}^*=\arg\min_{\boldsymbol{\theta}\in\Theta} \sum_{t=s}^{s+\tau-1} f_t(\mathcal{G}_t(\boldsymbol{\theta}))$ where $[s,s+\tau-1]\subseteq[T]$.
Under Assumptions \ref{assmp1}, \ref{assmp2} and \ref{assmp3}, the regret FairSAR proposed in Eq.(\ref{eq:ourRegret}) of \sysname{} in Algorithm \ref{alg:ouralgorithm} satisfies the bounds in Eq.(\ref{eq:regret-bounds}) for all three interval settings that stated in Section \ref{sec:intervals}.
\begin{equation}
\label{eq:regret-bounds}
\begin{aligned}
    &\max_{[s,s+\tau-1]\subseteq[T]} \bigg( \sum_{t=s}^{s+\tau-1} f_t\Big(\mathcal{G}_t(\boldsymbol{\theta}_t)\Big) - f_t\Big(\mathcal{G}_t(\boldsymbol{\theta}^*)\Big) \bigg) \leq \mathcal{O}\Big((\tau\log T)^{1/2}\Big)\\
    &\max_{[s,s+\tau-1]\subseteq[T]} \bigg (\sum_{t=s}^{s+\tau-1} g_i\Big(\mathcal{G}_t(\boldsymbol{\theta}_t)\Big) \bigg)\leq \mathcal{O}\Big((\tau T\log T)^{1/4}\Big), \quad \forall i \in[m]
\end{aligned}
\end{equation}
\end{theorem}

Under Assumptions \ref{assmp1}, \ref{assmp2} and \ref{assmp3}, we target Eq.(\ref{eq:L_t}) and have
\begin{lemma}[Theorem 1 in \cite{zhao-KDD-2021}]
\label{lemma:convex-concave}
\begin{sloppypar}
    Suppose $f$ and $g:\Theta\times\mathbb{R}_+^m \rightarrow \mathbb{R}$ satisfy Assumptions \ref{assmp1}, \ref{assmp2} and \ref{assmp3}. The interval-level update and the augmented Lagrangian function $\mathcal{L}_t(\boldsymbol{\theta, \lambda})$ are defined in Eq.(\ref{eq:inner-pd-update})(12) and Eq.(\ref{eq:L_t}). Then, the function $\mathcal{L}_t(\boldsymbol{\theta, \lambda})$ is convex-concave with respect to the arguments $\boldsymbol{\theta}$ and $\boldsymbol{\lambda}$, respectively. Furthermore, as for $\mathcal{L}_t(\boldsymbol{\cdot,\lambda})$, if stepsize $\eta_{t,I}$ for each active expert $E_I$ is selected as $\eta_{t,I}\leq\min\{\frac{\mu_f+\Bar{\lambda}m\mu_g}{8(L_f+\Bar{\lambda}mL_g)(\rho_f+\Bar{\lambda}m\rho_g)}, \frac{1}{2(\beta_f+\Bar{\lambda}m\beta_g)}\}$, then $\mathcal{L}_t(\boldsymbol{\cdot,\lambda})$ enjoys $\frac{9}{8}(\beta_f+\Bar{\lambda}m\beta_g)$-smooth and $\frac{1}{8}(\mu_f+\Bar{\lambda} m\mu_g)$-strongly convex, where $\Bar{\lambda}\geq 0$ is the mean value of $\boldsymbol{\lambda}$.
\end{sloppypar}
\end{lemma}
According to Theorems 1 and 3 in \cite{luo-2015-achieving} and the Lemma 1 in \cite{zhang-2020-AISTATS}, we have the following lemma with respect to Eq.(\ref{eq:L_t}) that
\begin{lemma}
\label{lemma:meta-regret}
    Under Assumption \ref{assmp2}, for any interval $I=[i,j]\in\mathcal{I}$, \sysname{} satisfies
    \begin{equation*}
    \begin{aligned}
        \sum_{u=i}^t \mathcal{L}_u(\boldsymbol{\theta}_u,\boldsymbol{\lambda}_u)-\sum_{u=i}^t \mathcal{L}_u(\boldsymbol{\theta}_{u,I},\boldsymbol{\lambda}_{u,I}) \leq S\sqrt{6L_f L_g(t-i-1)c(t)}
    \end{aligned}
    \end{equation*}
    where $c(t)\leq 1+\ln t + \ln(1+\log_2^T)+\ln\frac{5+3\ln (1+t)}{2}$.
\end{lemma}
By applying Lemma \ref{lemma:meta-regret} with the Theorem 2 in \cite{zhang-2020-AISTATS}, we have
\begin{lemma}
\label{lemma:regret-single-interval}
    Under Assumption \ref{assmp1} and \ref{assmp2}, for any interval $I=[i,j]\in\mathcal{I}$, for any $(\boldsymbol{\theta,\lambda})\in\Theta\times\mathbb{R}^m_+$ \sysname{} satisfies
    \begin{equation*}
    \begin{aligned}
        \sum_{t\in I} \mathcal{L}_t(\mathcal{G}_t(\boldsymbol{\theta}_t),\boldsymbol{\lambda})-
        \sum_{t\in I} \mathcal{L}_t(\boldsymbol{\theta},\boldsymbol{\lambda}_{t,I})
        \leq S\sqrt{|I|}(\sqrt{6L_f L_g c(t)}+G)
    \end{aligned}
    \end{equation*}
\end{lemma}
To extend our Lemma \ref{lemma:regret-single-interval} to any interval $I=[r,s]\subseteq [T]$, we refer the following lemma
\begin{lemma}[Lemma 3 in \cite{zhang-2020-AISTATS}]
\label{lemma:interval patitions}
    For any interval $[r,s]\subseteq[T]$, it can be partitioned into two sequences of disjoint and consecutive intervals, denoted by $I_{-p},...,I_0\in\mathcal{I}$ and $I_1,...,I_q\in\mathcal{I}$, such that
    \begin{align*}
        |I_{-i}|/|I_{-i+1}|\leq 1/2, \forall i\geq 1 \quad \textit{and} \quad |I_i|/|I_{i-1}|\leq 1/2, \forall i\geq 2
    \end{align*}
\end{lemma}
Finally, we prove the proposed Theorem \ref{main theorem}.

\begin{proof}
\begin{sloppypar}
By applying Lemma \ref{lemma:regret-single-interval} onto Lemma \ref{lemma:interval patitions} and set $\boldsymbol{\theta}^*$ being the optimal solution for $\min_{\boldsymbol{\theta}\in\Theta} \sum_{t=r}^s f_t(\mathcal{G}_t(\boldsymbol{\theta}))$ where $[r,s]\subseteq[T]$, we have
\end{sloppypar}
\begin{equation}
\begin{aligned}
\label{eq:L_t regret}
    \sum_{t=r}^s \mathcal{L}_t(\mathcal{G}_t(\boldsymbol{\theta}_t),\boldsymbol{\lambda})- \sum_{t=r}^s \mathcal{L}_t(\mathcal{G}_t(\boldsymbol{\theta}^*),\boldsymbol{\lambda}_{t,I}) 
    =&\sum_{i=-p}^q \Big( \sum_{t\in I_i} \mathcal{L}_t(\mathcal{G}_t(\boldsymbol{\theta}_t),\boldsymbol{\lambda})- \sum_{t\in I_i} \mathcal{L}_t(\mathcal{G}_t(\boldsymbol{\theta}^*),\boldsymbol{\lambda}_{t,I})\Big) \\
    \leq&\sum_{i=-p}^q S\sqrt{|I_i|}(\sqrt{6L_f L_g c(s)}+G)\\
    \leq& 2S(\sqrt{6L_f L_g c(s)}+G)\sum_{i=0}^{\infty} \sqrt{2^{-i}|I|}\\
    \leq& 8S(\sqrt{6L_f L_g c(s)}+G) \sqrt{|I|}
\end{aligned}
\end{equation}
By expanding Eq.(\ref{eq:L_t regret}) using Eq.(\ref{eq:L_t}) and following the Theorem 3.1 in \cite{Cambridge-book-2006}, we have
\begin{equation*}
\begin{aligned}
    &\sum_{t=r}^s\Big\{f_t(\mathcal{G}_t(\boldsymbol{\theta}_t))-f_t(\mathcal{G}_t(\boldsymbol{\theta}^*))\Big\} 
    +\sum_{i=1}^m\Big\{\lambda_i\sum_{t=r}^s g_i(\mathcal{G}_t(\boldsymbol{\theta}_t))-\sum_{t=r}^s\lambda_{t,i}g_i(\mathcal{G}_t(\boldsymbol{\theta}^*))\Big\}\\
    &-\frac{\delta(\eta_1+\eta_2) (s-r+1)}{2}||\boldsymbol{\lambda}||^2+\frac{\delta(\eta_1+\eta_2)}{2}\sum_{t=r}^s||\boldsymbol{\lambda}||^2 
    \leq 8S \Big(\sqrt{6L_f L_g c(s)}+G\Big) \sqrt{|I|}
\end{aligned}
\end{equation*}
Here, we approximately average $p_{t,I_k}$ for all experts $E_k\in\mathcal{U}_t$ at time $t$, and hence the subscription $k$ is omitted. 
Inspired by the proof of Theorem 4 in \cite{OGDLC-2012-JMLR}, we take maximization for $\boldsymbol{\lambda}$ over $(0,+\infty)$ and get
\begin{equation*}
\begin{aligned}
    \sum_{t=r}^s\Big\{f_t(\mathcal{G}_t(\boldsymbol{\theta}_t))-f_t(\mathcal{G}_t(\boldsymbol{\theta}^*))\Big\}
    &+\sum_{i=1}^m\Big\{\frac{\big[\sum_{t=r}^s g_i(\mathcal{G}_t(\boldsymbol{\theta}_t))\big]^2_+}{2(\delta(\eta_1+\eta_2)(s-r+1)+\frac{m}{\eta_1+\eta_2})}-\sum_{t=r}^s\lambda_{t,i}g_i(\mathcal{G}_t(\boldsymbol{\theta}^*))\Big\}\\
    &\leq 8S\Big(\sqrt{6L_f L_g c(s)}+G\Big) \sqrt{|I|}
\end{aligned}
\end{equation*}
Since $g_i(\mathcal{G}_t(\boldsymbol{\theta}^*))\leq 0$ and $\lambda_{t,i}\geq 0, \forall i\in[m]$, the resulting inequality becomes
\begin{equation*}
\begin{aligned}
    \sum_{t=r}^s\Big\{f_t(\mathcal{G}_t(\boldsymbol{\theta}_t))-f_t(\mathcal{G}_t(\boldsymbol{\theta}^*))\Big\}
    +\sum_{i=1}^m\frac{\big[\sum_{t=r}^s g_i(\mathcal{G}_t(\boldsymbol{\theta}_t))\big]^2_+}{2(\delta(\eta_1+\eta_2)(s-r+1)+\frac{m}{\eta_1+\eta_2})} \leq 8S\Big(\sqrt{6L_f L_g c(s)}+G\Big) \sqrt{|I|}
\end{aligned}
\end{equation*}
Due to non-negative of $\frac{\big[\sum_{t=r}^s g_i(\mathcal{G}_t(\boldsymbol{\theta}_t))\big]^2_+}{2(\delta(\eta_1+\eta_2)(s-r+1)+\frac{m}{\eta_1+\eta_2})}$, we have
\begin{equation*}
\begin{aligned}
    \sum_{t=r}^s\Big\{f_t(\mathcal{G}_t(\boldsymbol{\theta}_t))-f_t(\mathcal{G}_t(\boldsymbol{\theta}^*))\Big\} \leq 8S\Big(\sqrt{6L_f L_g c(s)}+G\Big) \sqrt{|I|} =\mathcal{O}\Big((|I|\log s)^{1/2}\Big)
\end{aligned}
\end{equation*}
Furthermore, we have $\sum_{t=r}^s\Big\{f_t(\mathcal{G}_t(\boldsymbol{\theta}_t))-f_t(\mathcal{G}_t(\boldsymbol{\theta}^*))\Big\}\geq-F(s-r+1)$ according to the assumption and set $\eta_1=\eta_2=\mathcal{O}(1/\sqrt{s})$.
We have
\begin{equation*}
\begin{aligned}
    \sum_{t=r}^s g_i(\mathcal{G}_t(\boldsymbol{\theta}_t))\leq \mathcal{O}\Big((|I|s\log s)^{1/4}\Big)
\end{aligned}
\end{equation*}
Therefore, as for FairSAR proposed in Eq.(\ref{eq:ourRegret}), we complete the proof.
\end{proof}

\begin{table*}[t]
\caption{Comparison of upper bounds in loss regret and constraint violations across various methods.}
\scriptsize
\setlength\tabcolsep{1.2pt}
\begin{tabular}{cccccccc}
\cline{2-8}
\multicolumn{1}{l}{} & \multicolumn{4}{c}{Static Environment} & \multicolumn{3}{|c}{Changing Environment} \\ 
\hline
\multicolumn{1}{c|}{Algorithms}  & \multicolumn{1}{c}{FTML\cite{Finn-ICML-2019}} & \multicolumn{1}{c}{FairFML\cite{zhao-KDD-2021}} & \multicolumn{1}{c}{FairAOGD\cite{AdpOLC-2016-ICML}} & FairGLC\cite{GenOLC-2018-NeurIPS} & \multicolumn{1}{|c}{AOD\cite{zhang-2020-AISTATS}} & \multicolumn{1}{c}{CBCE\cite{Jun-2017-AISTATS}} & FairSAOML(Ours) \\ 
\hline
\multicolumn{1}{c|}{Loss Regret} & \multicolumn{1}{c}{$\mathcal{O}(\log T)$} & \multicolumn{1}{c}{$\mathcal{O}(\log T)$} & \multicolumn{1}{c}{$\mathcal{O}(T^{2/3})$} & $\mathcal{O}(\log T)$ & \multicolumn{1}{|c}{$\mathcal{O}\big((\tau\log T)^{1/2}\big)$} & \multicolumn{1}{c}{$\mathcal{O}\big((\tau\log T)^{1/2}\big)$} & $\mathcal{O}\big((\tau\log T)^{1/2}\big)$ \\ 
\hline
\multicolumn{1}{c|}{\begin{tabular}[c]{@{}c@{}} Constraint Violations\end{tabular}} & \multicolumn{1}{c}{-} & \multicolumn{1}{c}{$\mathcal{O}\big((T\log T)^{1/2}\big)$} & \multicolumn{1}{c}{$\mathcal{O}(T^{2/3})$} & $\mathcal{O}\big((T\log T)^{1/2}\big)$ & \multicolumn{1}{|c}{-} & \multicolumn{1}{c}{-} & $\mathcal{O}\big((\tau T\log T)^{1/4}\big)$ \\ 
\hline
\end{tabular}
\label{tab:analysis comparison}
\end{table*}
\textbf{Discussion for Upper Bounds.} Under aforementioned assumptions and provable convexity of Eq.(\ref{eq:L_t regret}) in $\boldsymbol{\theta}$ (see Lemma \ref{lemma:convex-concave}), the proposed \sysname{} in Algorithm \ref{alg:ouralgorithm} achieves sub-linear bounds in FairSAR for both loss regret and violation of fairness constraints. 
Although such bounds are comparable with the strongly adapted loss regret in \cite{Jun-2017-AISTATS,zhang-2020-AISTATS} (see Table \ref{tab:analysis comparison})
in terms of online learning in changing environment paradigms, we bound loss regret and cumulative fairness constraints simultaneously. On the other hand, in terms of fairness-aware online learning, our proposed method outperforms \cite{zhao-KDD-2021,AdpOLC-2016-ICML,GenOLC-2018-NeurIPS} by giving a tighter bound of fair constraint violations.

\textbf{Complexity.} The computational complexity of \sysname{} in Algorithm \ref{alg:ouralgorithm} at each time $t\in[T]$ is $\mathcal{O}(N_{meta}\cdot |\mathcal{U}_t|)$ where $N_{meta}$ is the number of meta-level iterations and $|\mathcal{U}_t|$ is the total number of experts that needs to be maintained at $t$, and the complexity of each expert is $\mathcal{O}(1)$.

\section{Experimental Settings}
\label{sec:experiments}


\subsection{Datasets} 
We use the following publicly available datasets. 
(1) \textit{New York Stop-and-Frisk (NYSF)} \cite{Koh-icml-2021} is a prominent dataset of a real-world application on policing in New York City from 2009 to 2010. It documents whether a pedestrian who was stopped on suspicion of weapon possession would in fact possess a weapon. 
As this data had a pronounced racial bias on African Americans, for each frisked record, we consider race as the binary protected attribute, that is black and non-black. Besides, this dataset consists of records collected in five different sub-districts, Manhattan (M), Brooklyn (B), Queens (Q), Bronx (R), and Staten (S). Since there are large performance disparities across districts and race groups, each district is viewed as an independent domain. To adapt the online learning setting, data in each domain is further split into 32 tasks and each task corresponds to ten days of a month with 111 non-protected features. According to \textit{DDP} values in Definition \ref{dbc definition}, the fairness levels from low to high are Bronx (0.74), Queens (0.68), Staten (0.65), Manhattan (0.53) and Brooklyn (0.44). The larger \textit{DDP} values indicate a lower fairness level.
We hence consider two settings for domain adaptation where each setting contains 96 tasks in total: (i) fairness level from high to low: Brooklyn to Manhattan to Staten (B$\rightarrow$M$\rightarrow$S); and (ii) fairness level from low to high: Bronx to Queens then Staten (R$\rightarrow$Q$\rightarrow$S).
(2) \textit{MovieLens}\footnote{https://grouplens.org/datasets/movielens/100k/} contains 100k ratings by 943 users on 1682 movies, and each rating is given a binary label (``recommending" if rating greater than 3, ``not recommending" otherwise). We consider gender as the protected attribute. To generate dynamic environments, following \cite{Wan_2021_AAAI}, we construct a larger dataset by combining three copies of the original data and flipping the original values of non-protected attributes by multiplying -1 for the middle copy. Therefore, each copy is considered as a data domain. Furthermore, each data copy is split into 30 tasks by timestamps, and there are 90 tasks in total.

\subsection{Evaluation Metrics} 
Two popular evaluation metrics are introduced that each allows quantifying the extent of bias taking into account the protected attribute.
\textit{Demographic Parity} (DP) \cite{Dwork-2011-CoRR} and \textit{Equalized Odds} (EO) \cite{Hardt-NIPS-2016} can be formalized as
\begin{align*}
    &\text{DP}=k, \:\text{if DP}\leq 1; \text{DP}=1/k, \:\text{otherwise}, \quad \text{where}\: k=\frac{\mathbb{P}(\hat{Y}=1|S=-1)}{\mathbb{P}(\hat{Y}=1|S=1)} \\
    &\text{EO}=k, \:\text{if EO}\leq 1; \text{EO}=1/k, \:\text{otherwise}, \quad \text{where}\: k=\frac{\mathbb{P}(\hat{Y}=1|S=-1, Y=y)}{\mathbb{P}(\hat{Y}=1|S=1, Y=y)}
\end{align*}
where $y\in\{-1,1\}$. 
The EO metric requires that $\hat{Y}$ have equal true and false positive rates between sub-groups. For both metrics, a value closer to 1 indicates fairness.

\subsection{Competing Methods}
We compare the performance of our algorithm \sysname{} on various interval settings (hyphenated by DI, AGC, and DGC) with six baseline methods. These baselines are chosen from three perspectives: online meta-learning (MaskFTML, FairFML), online fairness learning (FairFML, FairAOGD, FairGLC), and online learning in changing environments (AOD, CBCE).
\begin{itemize}
    \item \textbf{MaskFTML} \cite{Finn-ICML-2019}: the original FTML finds a sequence of meta parameters by simply applying MAML \cite{Finn-ICML-2017-(MAML)} at each round. To focus on fairness learning, this approach is applied to modified datasets by simply removing protected attributes.
    \item \textbf{FairFML} \cite{zhao-KDD-2021} controls bias in an online working paradigm and aims to attain zero-shot generalization with task-specific adaptation. Different from our \sysname{}, FairFML focuses on a static environment and assumes tasks sampled from an unchangeable distribution.
    \item \textbf{FairAOGD} \cite{AdpOLC-2016-ICML} is proposed for online learning with long-term constraints. In order to fit bias-prevention and compare them to \sysname{}, we specify such constraints as DDP stated in Definition \ref{dbc definition}.
    \item \textbf{FairGLC} \cite{GenOLC-2018-NeurIPS} rectifies FairAOGD by square-clipping the constraints in place of $g_i(\cdot), \forall i$.
    \item \textbf{AOD} \cite{zhang-2020-AISTATS} minimizes the strongly adaptive regret by running multiple online gradient descent algorithms over a set of dense geometric covering intervals. 
    \item \textbf{CBCE} \cite{Jun-2017-AISTATS} adapts to changing environments in an online learning paradigm by combining the sleeping bandits idea with the coin betting algorithm.
\end{itemize}

\subsection{Settings}
As discussed in Section \ref{sec:analysis}, the performance of our proposed method has been well justified theoretically for machine learning models
whose objectives are strongly convex and smooth.
However, in machine learning and fairness studies, due to the nonlinearity of neural networks, many problems have a non-convex landscape where theoretical analysis is challenging. Nevertheless, algorithms originally developed for convex optimization problems like gradient descent have shown promising results in practical non-convex settings \cite{Finn-ICML-2019}. Taking inspiration from these successes, we describe practical instantiations for the proposed online algorithm and empirically evaluate the performance in Section \ref{sec:results}. 

For each task, we set the number of fairness constraints to one, \textit{i.e.} $m=1$. For the rest, we follow the same settings as used in online meta-learning \cite{Finn-ICML-2019,zhao-KDD-2021}. In particular, we meta-train with a support size of 400 for each class and 800 for a query set, whereas $90\%$ (hundreds of datapoints) of task samples for evaluation. 
Besides, for the \textit{NYSF} dataset, we choose the base of $2$, and the total number of experts is $96$ for DI, $6$ for AGC, and $7$ for DGC. Similarly, we choose the base of $3$ for the \textit{MovieLens} dataset; hence, the number of experts is $90$ for DI, $4$ for AGC, and $5$ for DGC.
All the baseline models used to compare with our proposed approach share the same neural network architecture and parameter settings.
All the experiments are repeated ten times with the same settings, and the mean and standard deviation results are reported.


\subsection{Implementation Details and Hyperparameter Tuning}
Our neural network trained follows the same architecture used in \cite{Finn-ICML-2017-(MAML)}, which contains two hidden layers of size 40 with ReLU activation functions. In the training process of the \textit{MovieLens (NYSF)} data, each gradient is computed using a batch size of 200 (800) examples where each binary class contains 100 (400) examples. For each dataset, we tune the folowing hyperparameters: (1) the initial dual meta parameter $\boldsymbol{\lambda}_0$ is chosen from $\{$0.00001, 0.0001, 0.001, 0.01, 0.1, 1, 10, 100, 1000, 10000$\}$; (2) the interval-level gradient steps are chosen from 1 to 10; (3) the number of iterations $N_{meta}$ are chosen from $\{$20, 25, 30, 35, 40, 45, 50, 55, 60, 65, 70, 75, 80, 85, 90, 95, 100$\}$; (4) learning rates $\eta_1$ and $\eta_2$ for updating meta-level parameters in Eq.(\ref{eq:outer-pd-update}) and (15) are chosen from $\{$0.0001, 0.0005, 0.001, 0.005, 0.01, 0.05, 0.1, 0.5, 1, 5, 10, 50, 100, 500, 1000$\}$; (5) the positive constant $\delta$ used in the augmented term are chosen from $\{$10, 25, 50, 75, 100$\}$. 

\section{Results}
\label{sec:results}
    \begin{figure*}[!t]
\captionsetup[subfigure]{aboveskip=-1pt,belowskip=-1pt}
\centering
    \begin{subfigure}[b]{0.245\textwidth}
        \includegraphics[width=\textwidth]{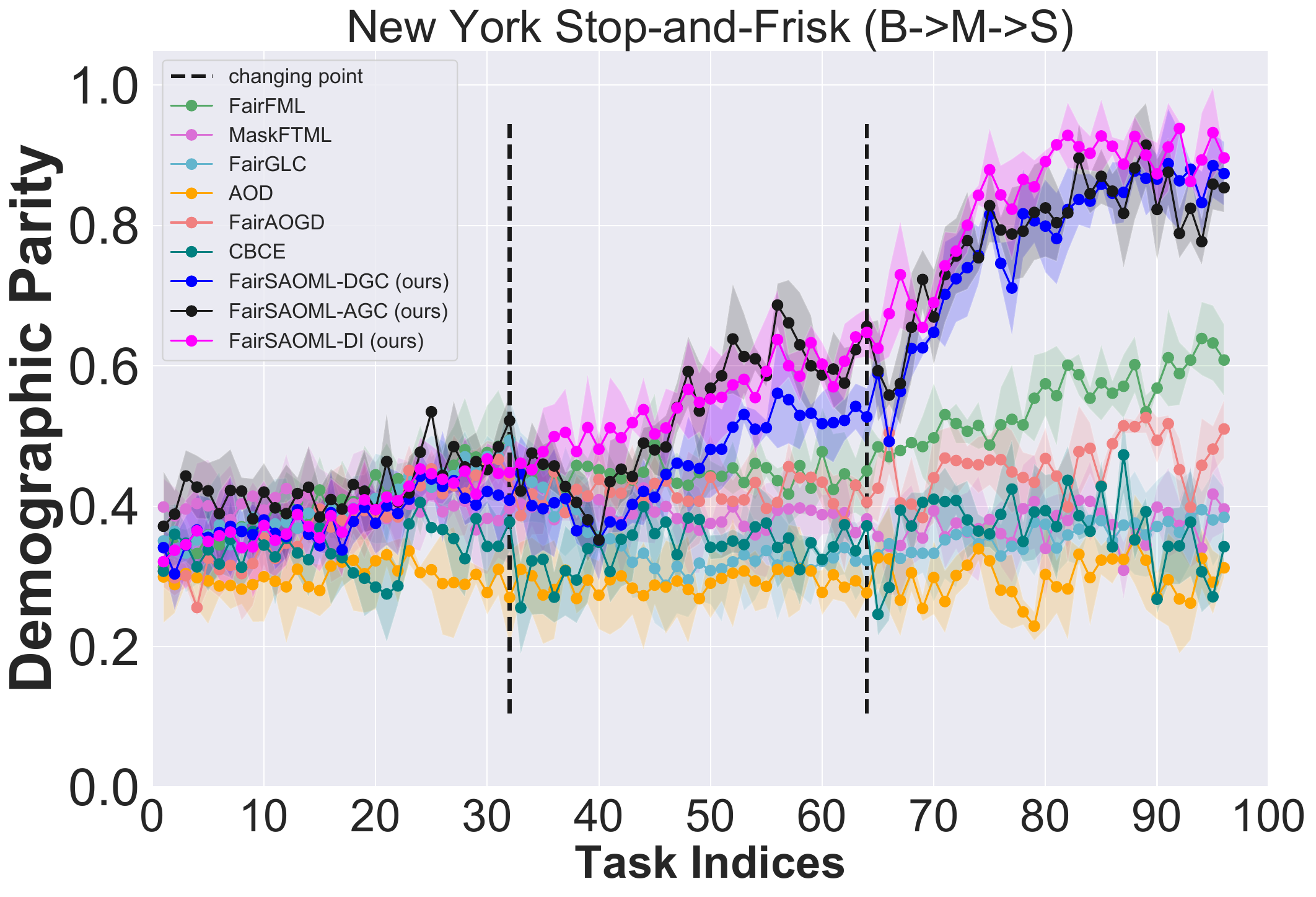}
        \caption{}
    \end{subfigure}
    \begin{subfigure}[b]{0.245\textwidth}
        \includegraphics[width=\textwidth]{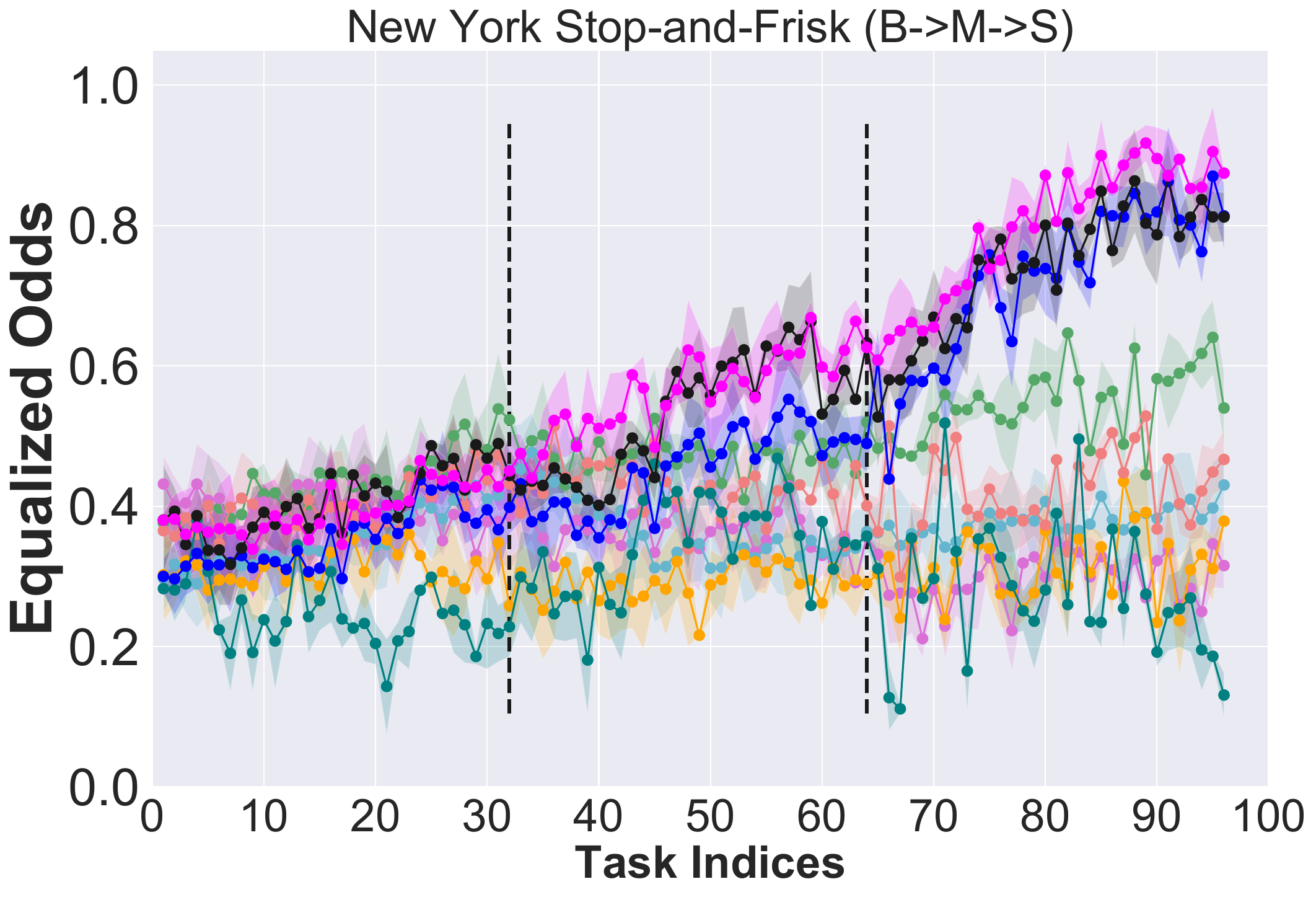}
        \caption{}
    \end{subfigure}
    \begin{subfigure}[b]{0.245\textwidth}
        \includegraphics[width=\textwidth]{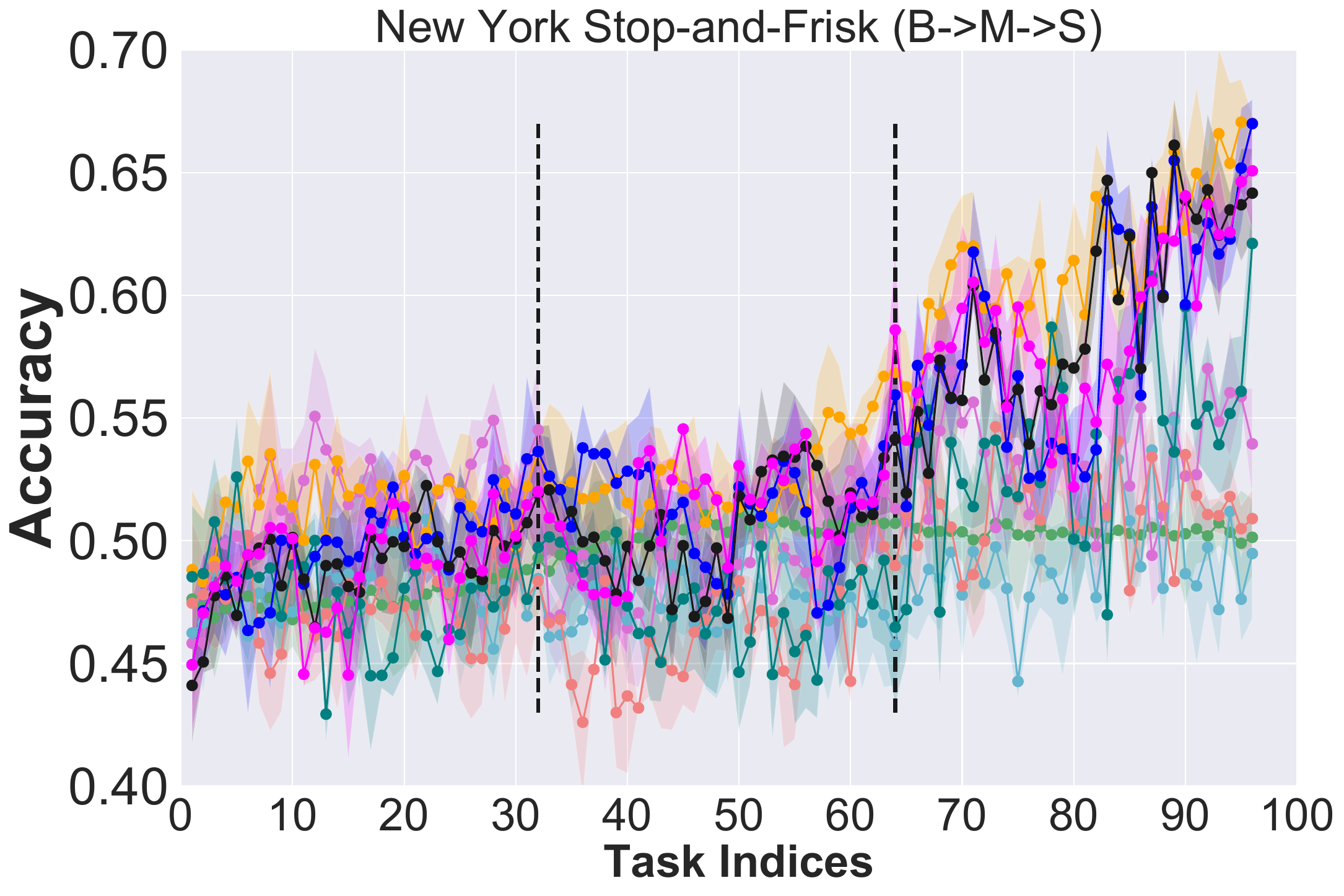}
        \caption{}
    \end{subfigure}
    \begin{subfigure}[b]{0.245\textwidth}
        \includegraphics[width=\textwidth]{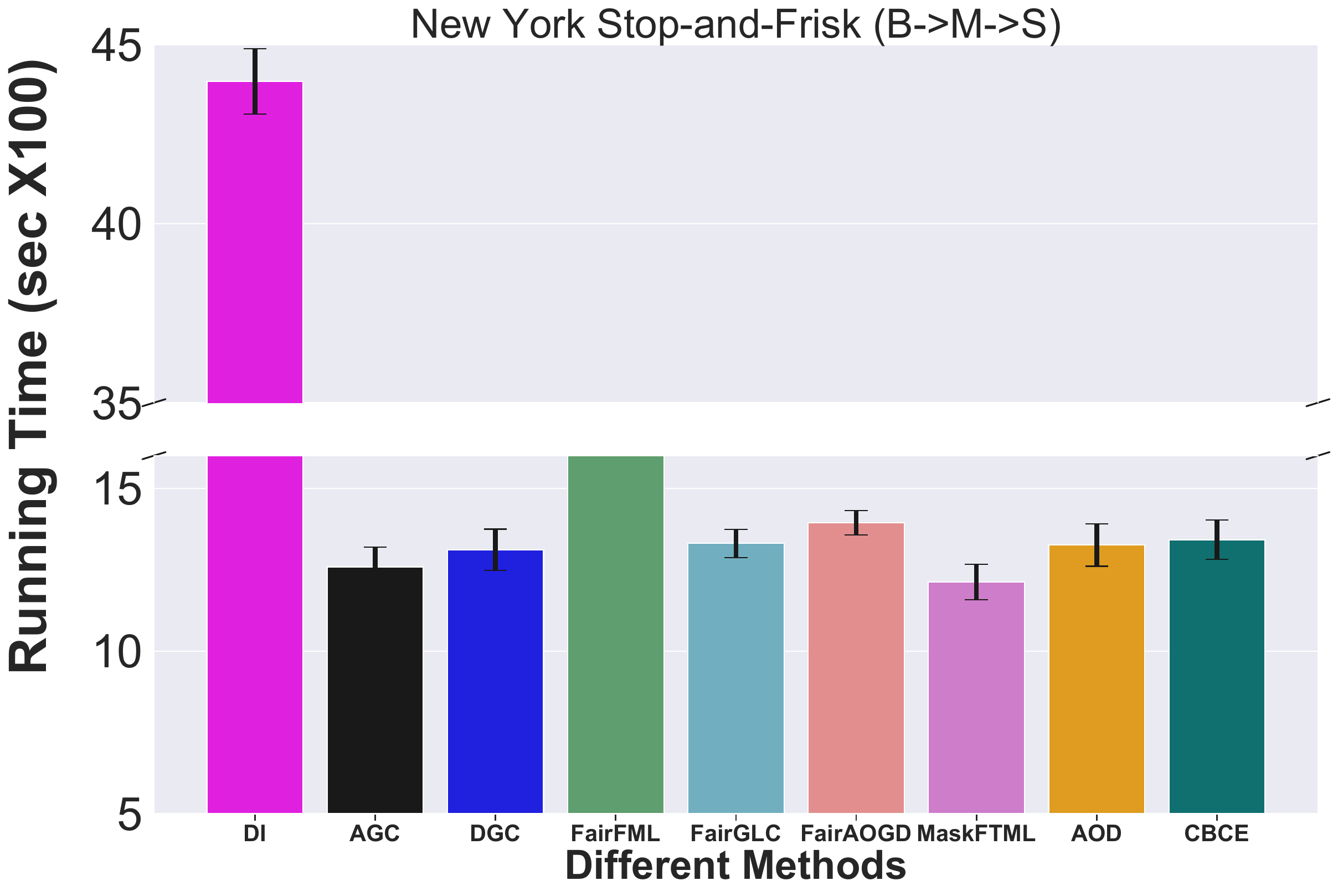}
        \caption{}
    \end{subfigure}

    \begin{subfigure}[b]{0.245\textwidth}
        \includegraphics[width=\textwidth]{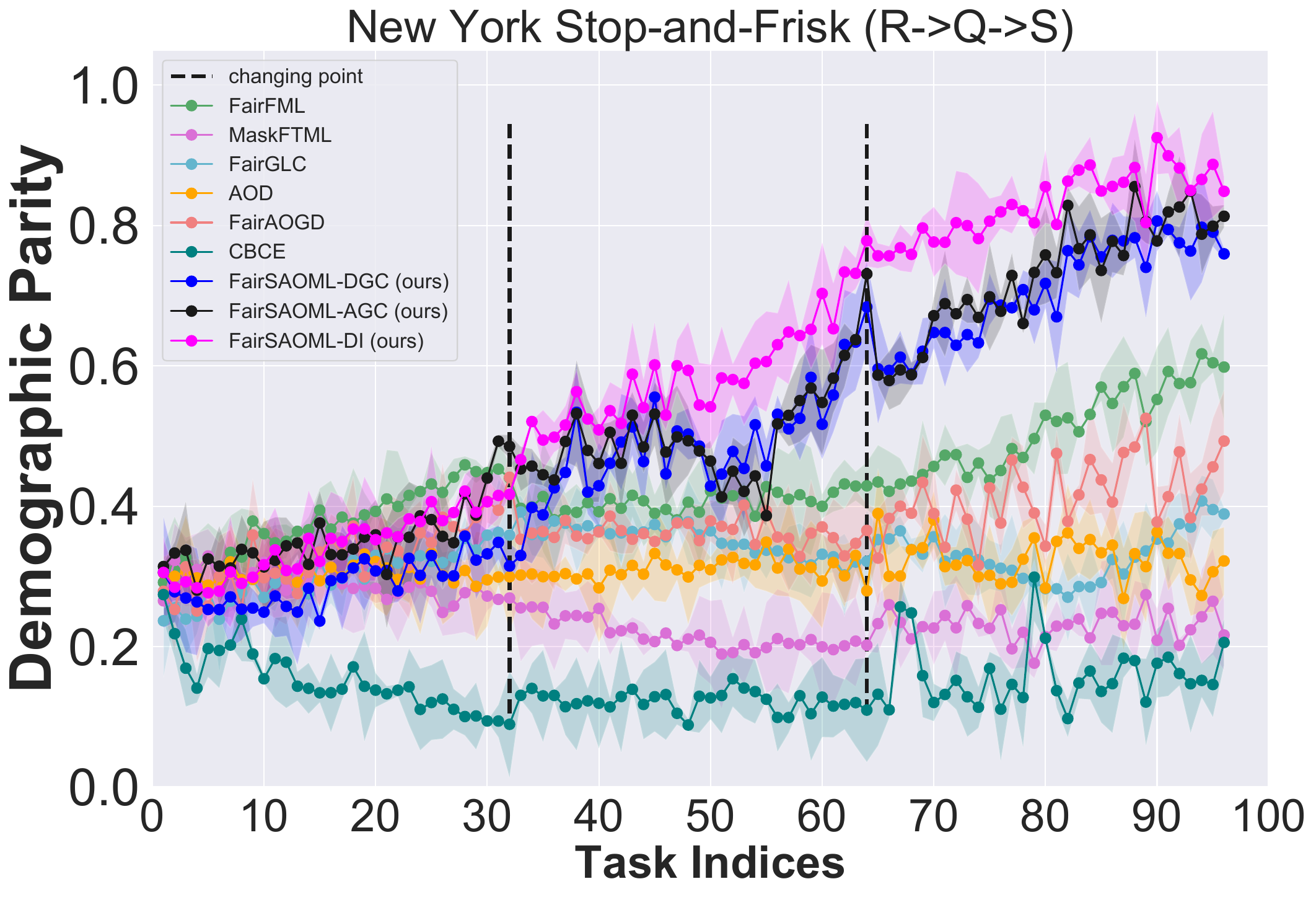}
        \caption{}
    \end{subfigure}
    \begin{subfigure}[b]{0.245\textwidth}
        \includegraphics[width=\textwidth]{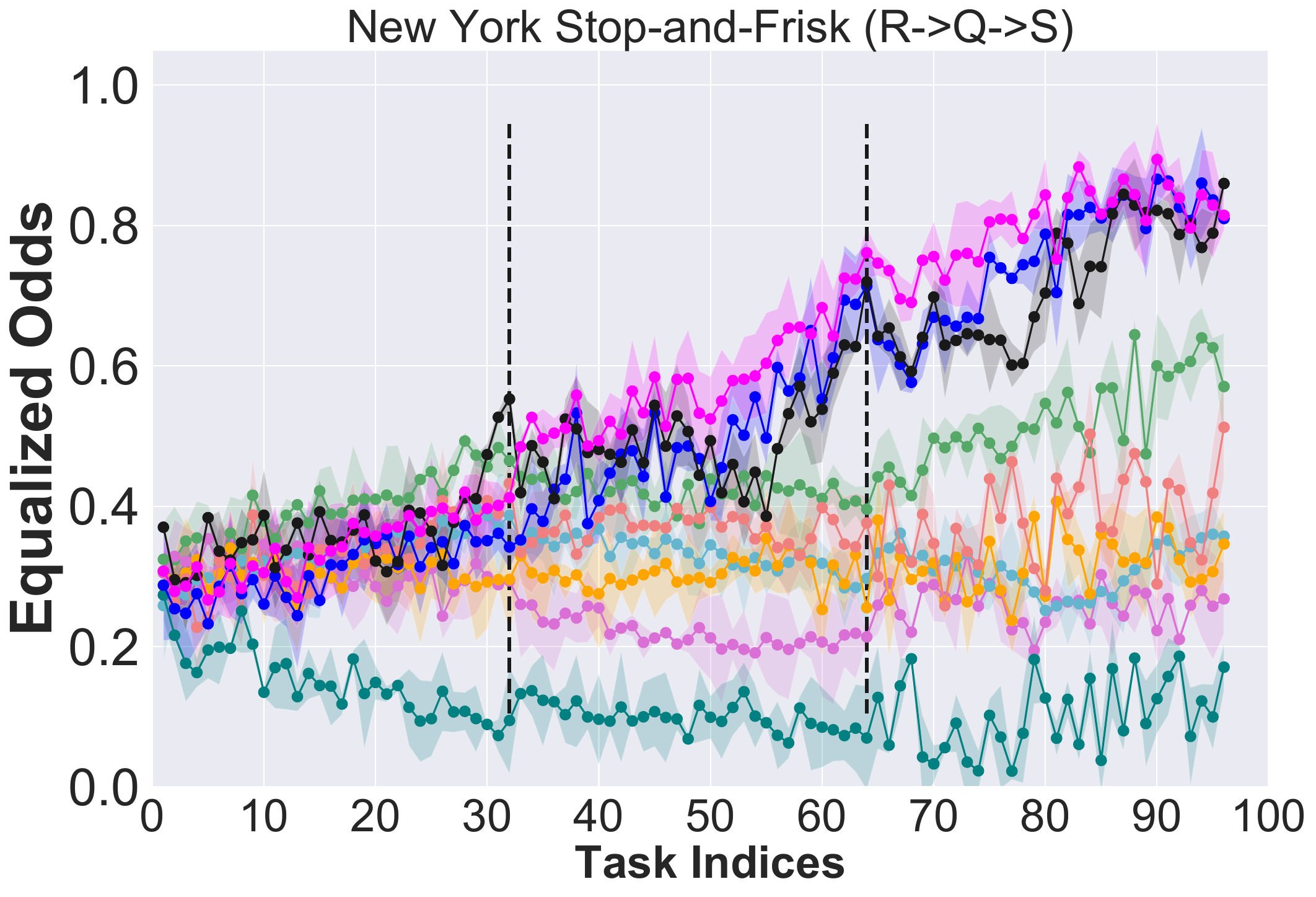}
        \caption{}
    \end{subfigure}
    \begin{subfigure}[b]{0.245\textwidth}
        \includegraphics[width=\textwidth]{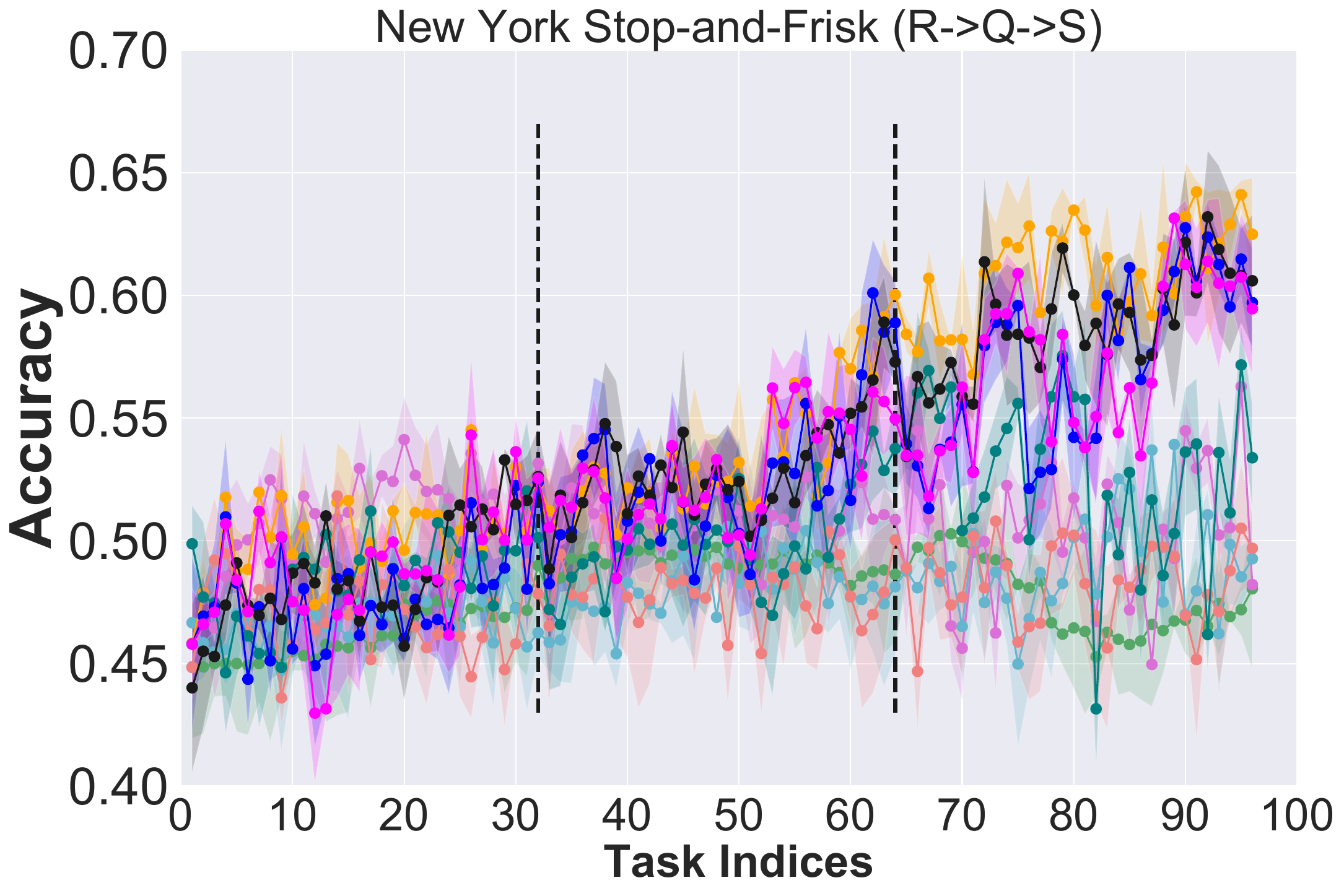}
        \caption{}
    \end{subfigure}
    \begin{subfigure}[b]{0.245\textwidth}
        \includegraphics[width=\textwidth]{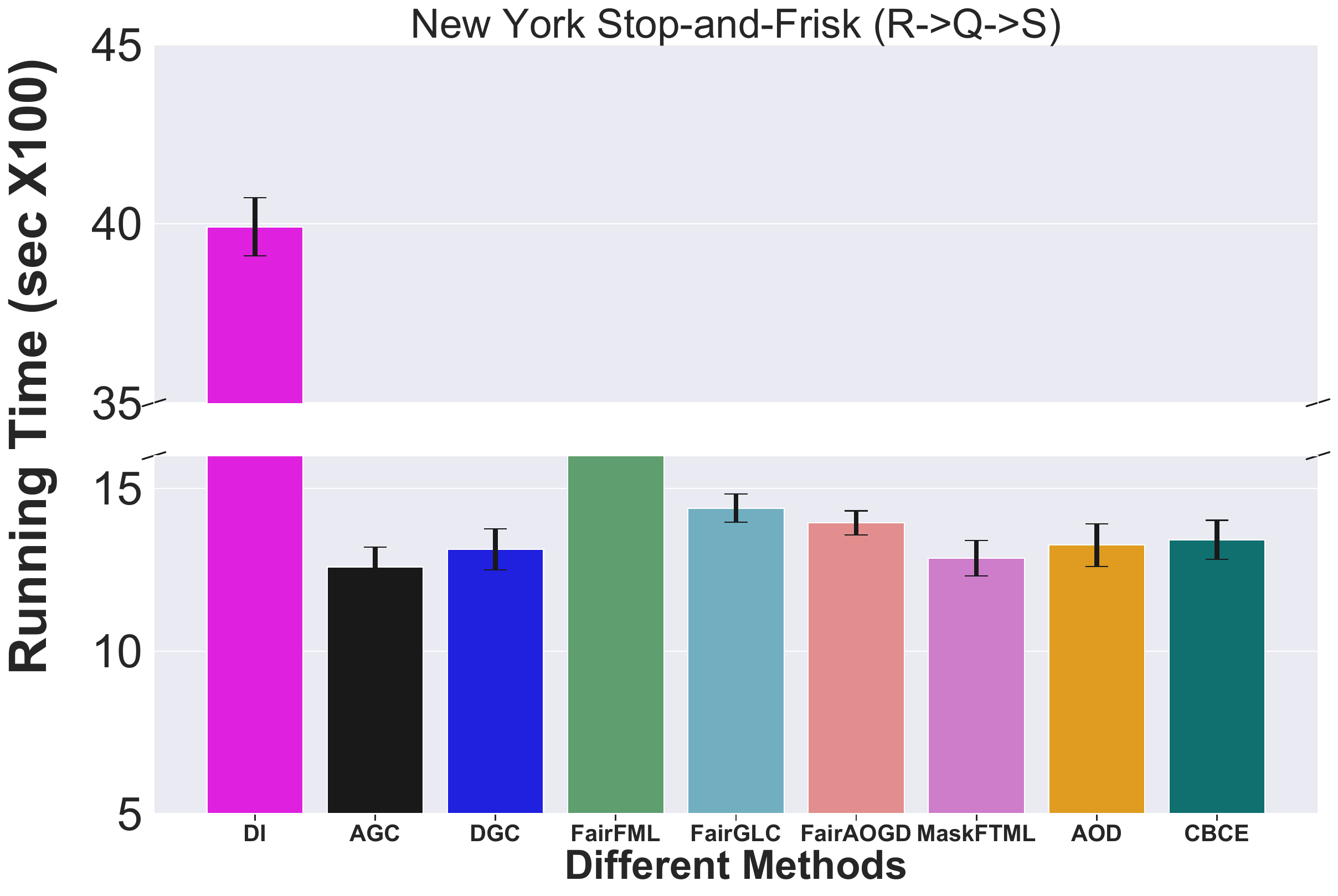}
        \caption{}
    \end{subfigure}

    \begin{subfigure}[b]{0.245\textwidth}
        \includegraphics[width=\textwidth]{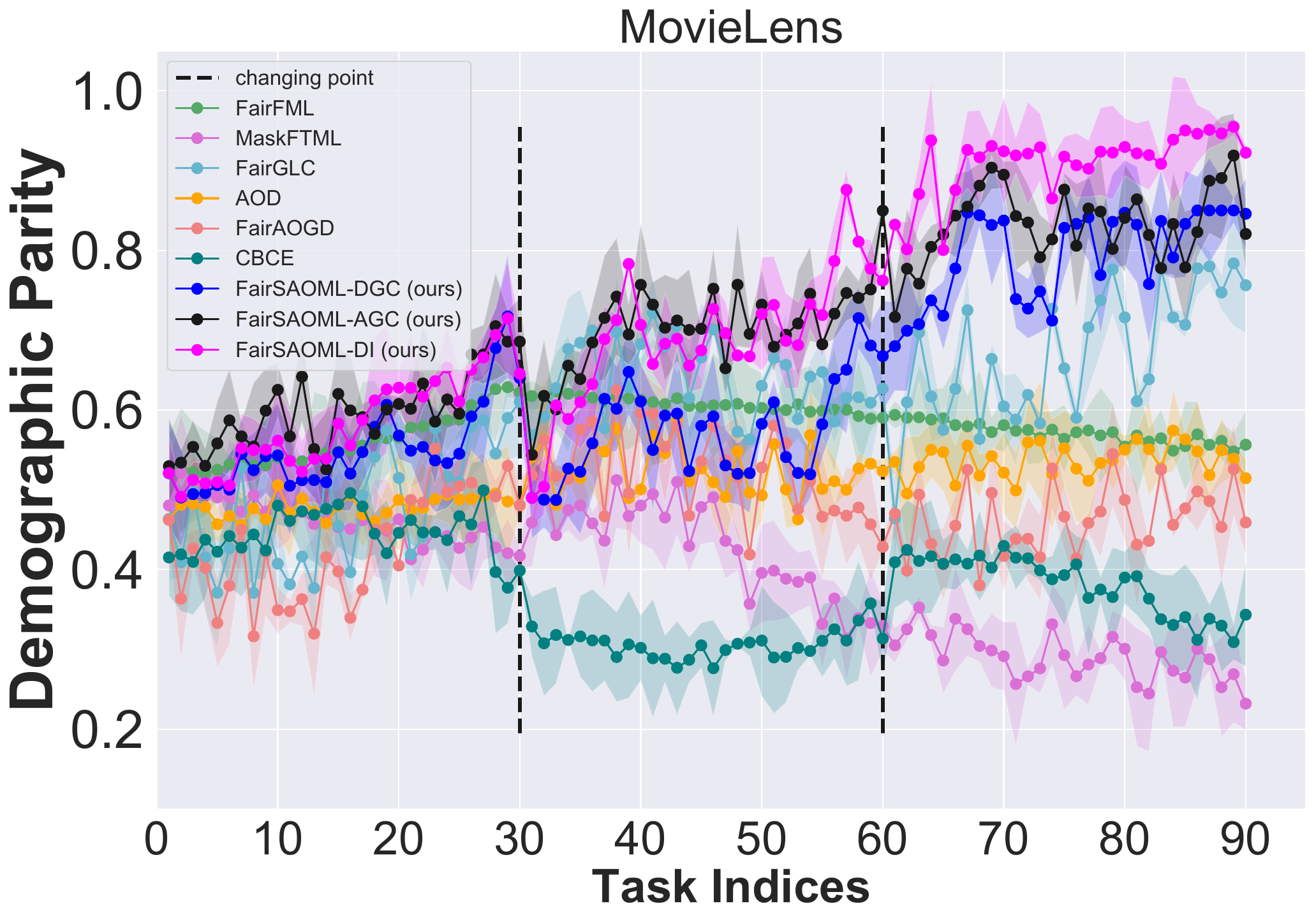}
        \caption{}
    \end{subfigure}
    \begin{subfigure}[b]{0.245\textwidth}
        \includegraphics[width=\textwidth]{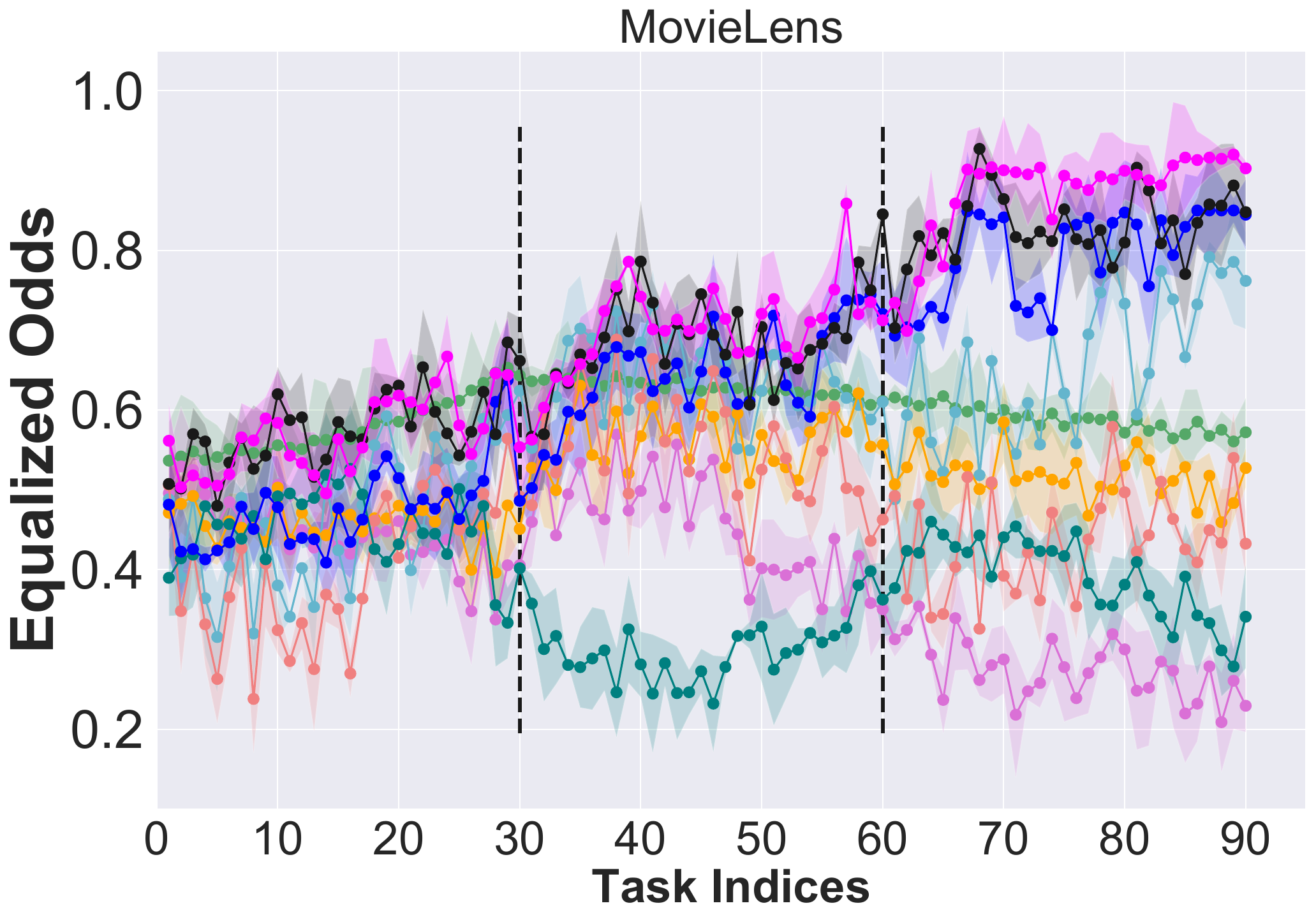}
        \caption{}
    \end{subfigure}
    \begin{subfigure}[b]{0.245\textwidth}
        \includegraphics[width=\textwidth]{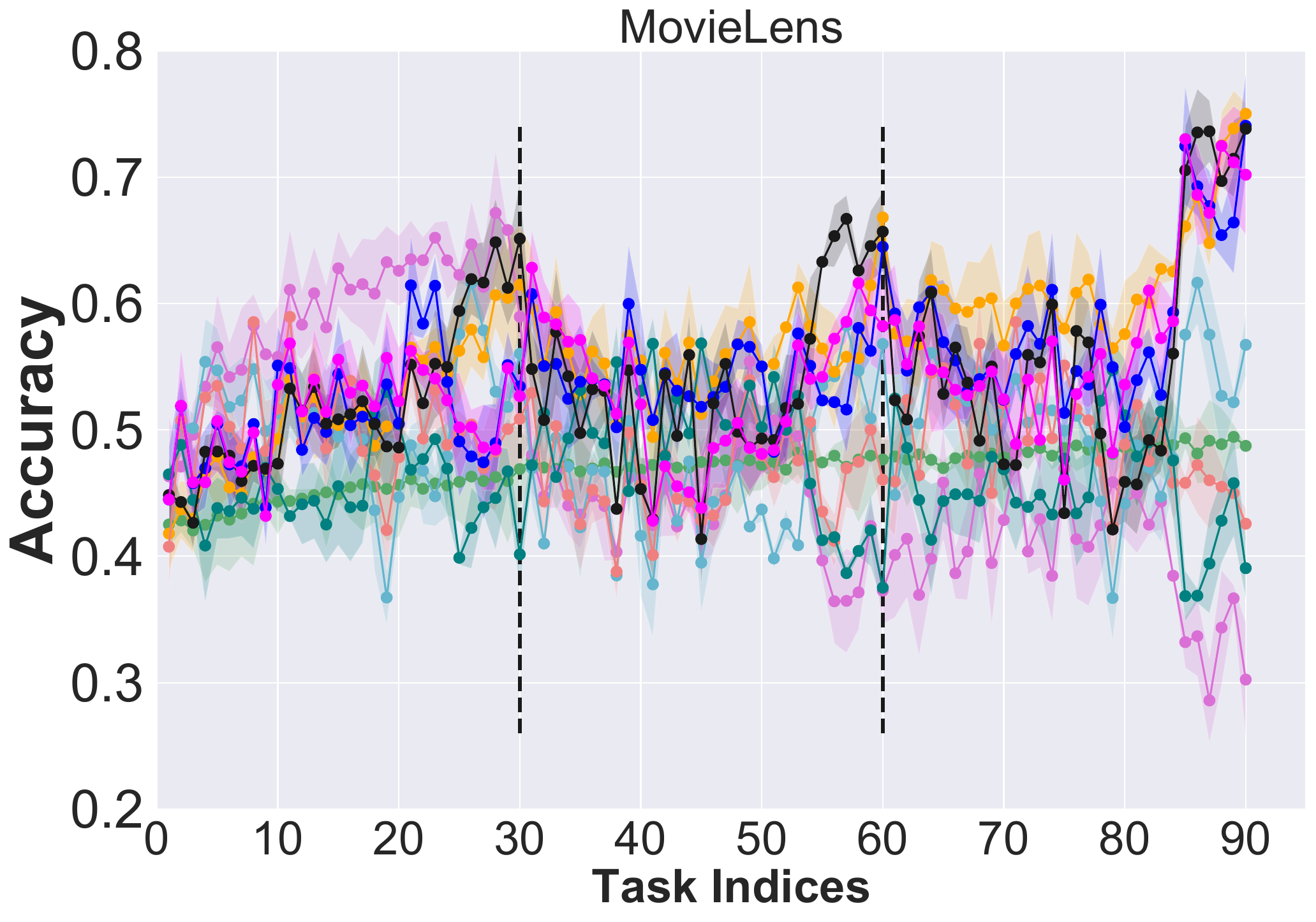}
        \caption{}
    \end{subfigure}
    \begin{subfigure}[b]{0.245\textwidth}
        \includegraphics[width=\textwidth]{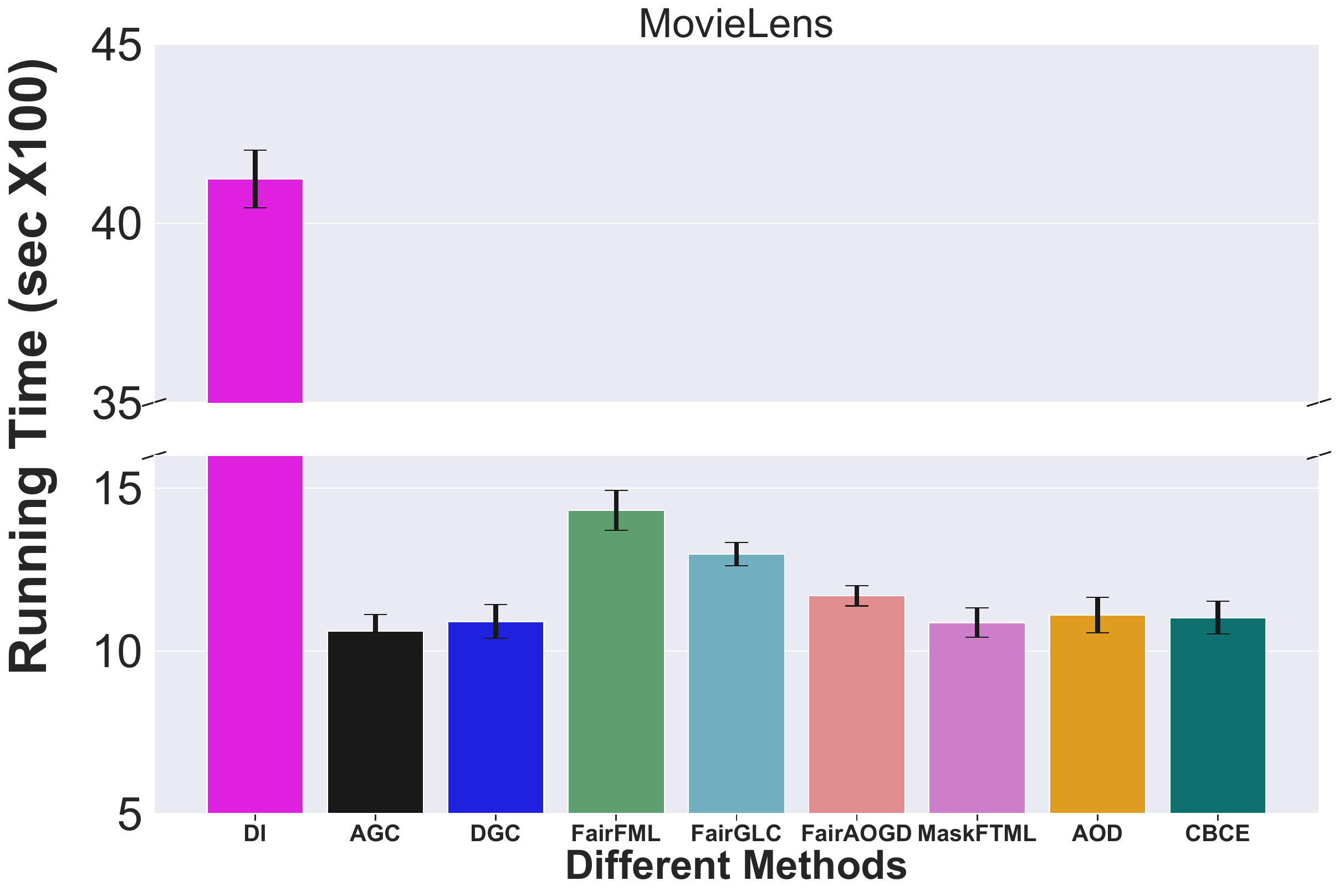}
        \caption{}
    \end{subfigure}
    \caption{Model performance over real-world datasets through each time. \textit{NYSF}
    \textbf{(a-c)} B$\rightarrow$M$\rightarrow$S, 
    \textbf{(d-f)} R$\rightarrow$Q$\rightarrow$S; 
    \textbf{(g-i)} \textit{MovieLens}.}
    \label{fig:dp-eop-acc}
\end{figure*}

\begin{figure*}[t]
\captionsetup[subfigure]{aboveskip=-1pt,belowskip=-1pt}
    \centering
    \begin{subfigure}[b]{0.325\textwidth}
        \includegraphics[width=\textwidth]{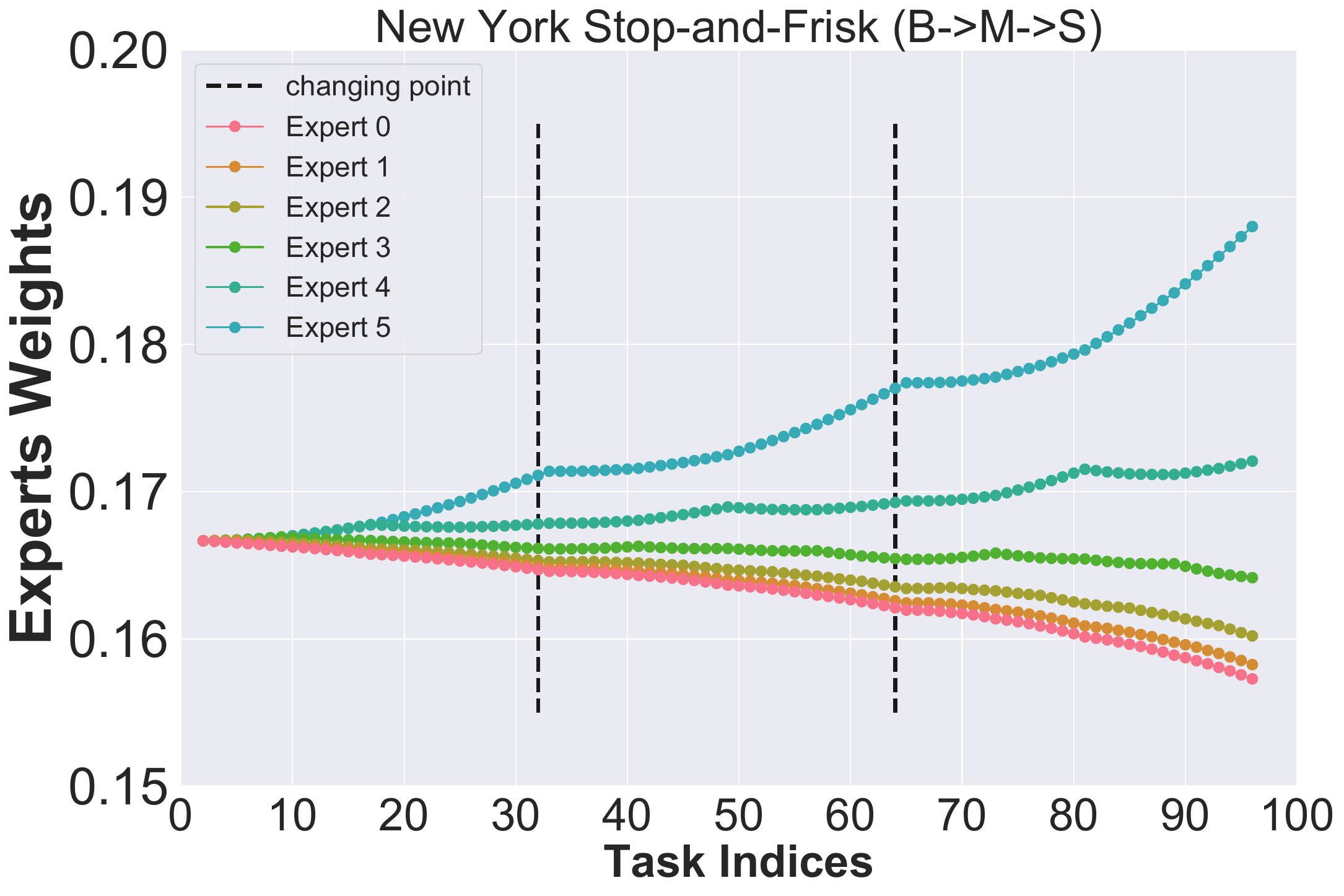}
        \caption{}
    \end{subfigure}
    \begin{subfigure}[b]{0.325\textwidth}
        \includegraphics[width=\textwidth]{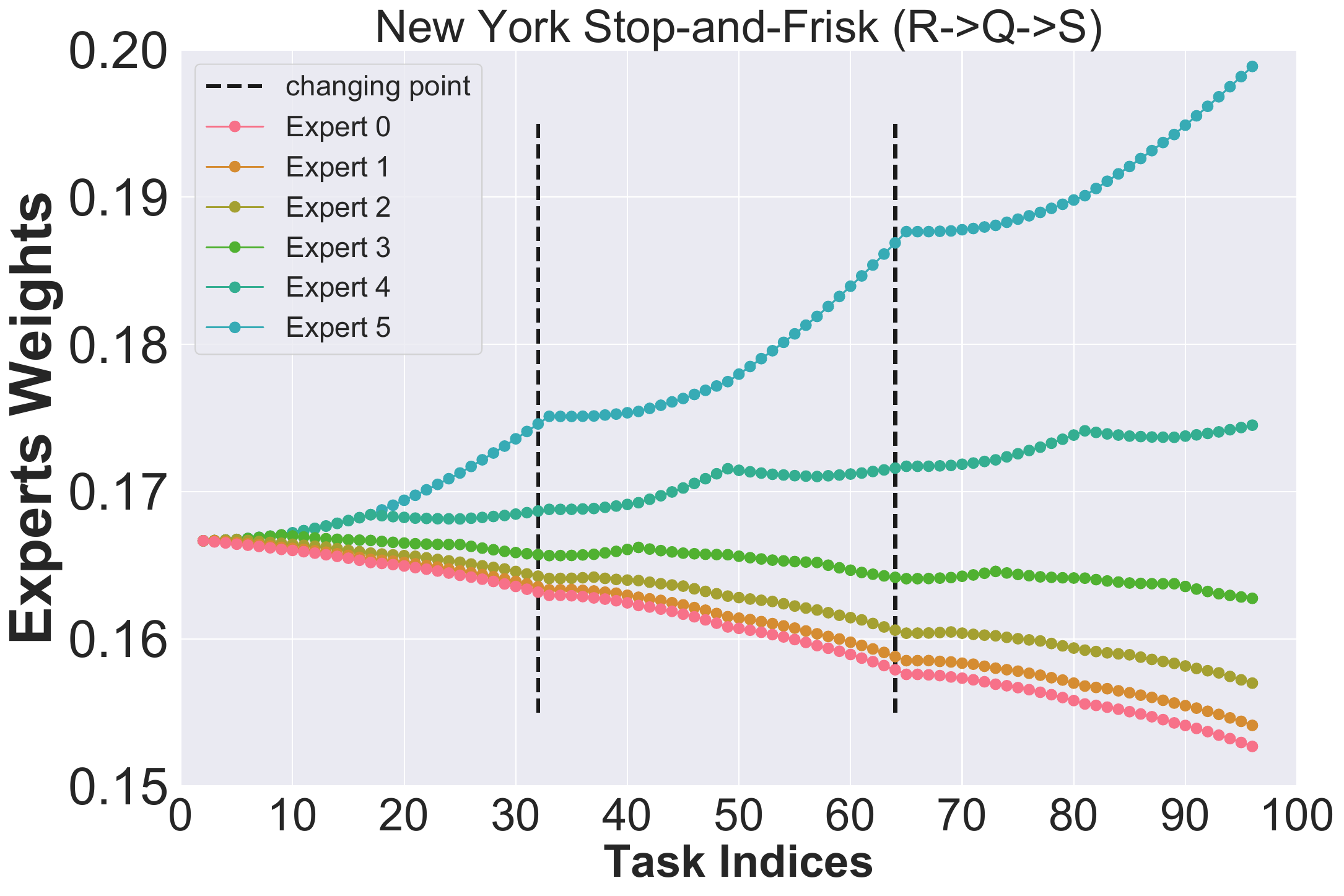}
        \caption{}
    \end{subfigure}
    \begin{subfigure}[b]{0.325\textwidth}
        \includegraphics[width=\textwidth]{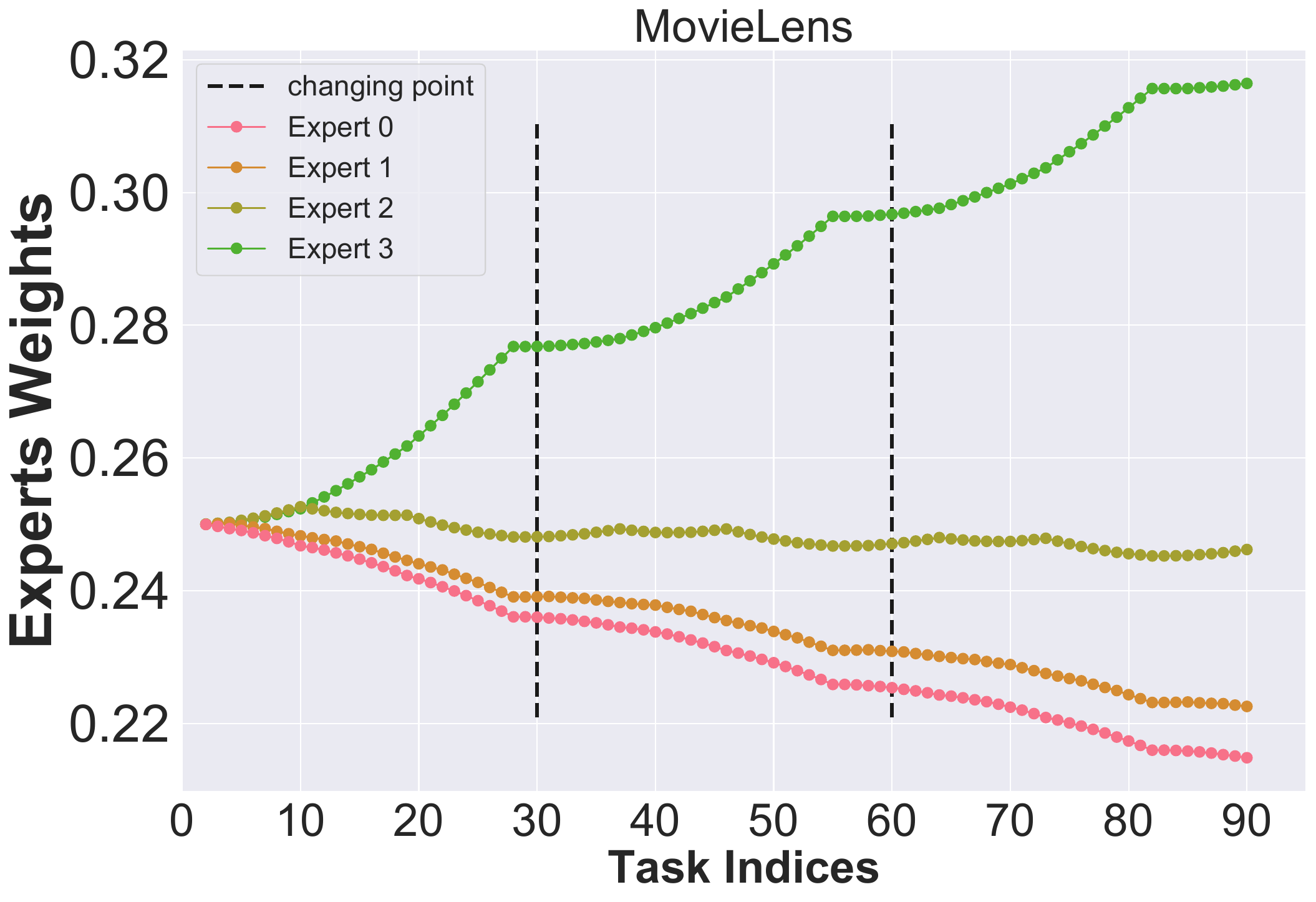}
        \caption{}
    \end{subfigure}

    \begin{subfigure}[b]{0.325\textwidth}
        \includegraphics[width=\textwidth]{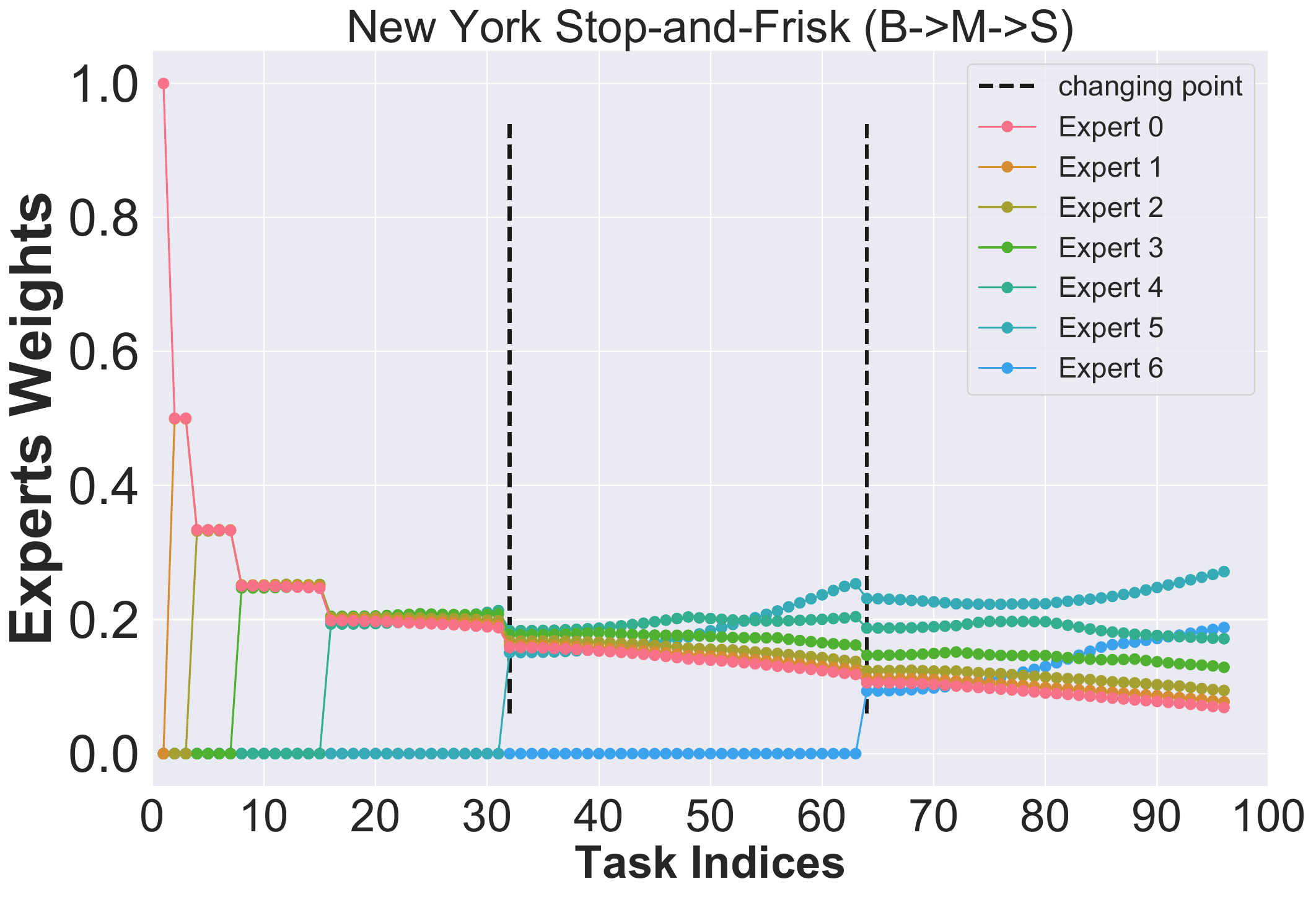}
        \caption{}
    \end{subfigure}
    \begin{subfigure}[b]{0.325\textwidth}
        \includegraphics[width=\textwidth]{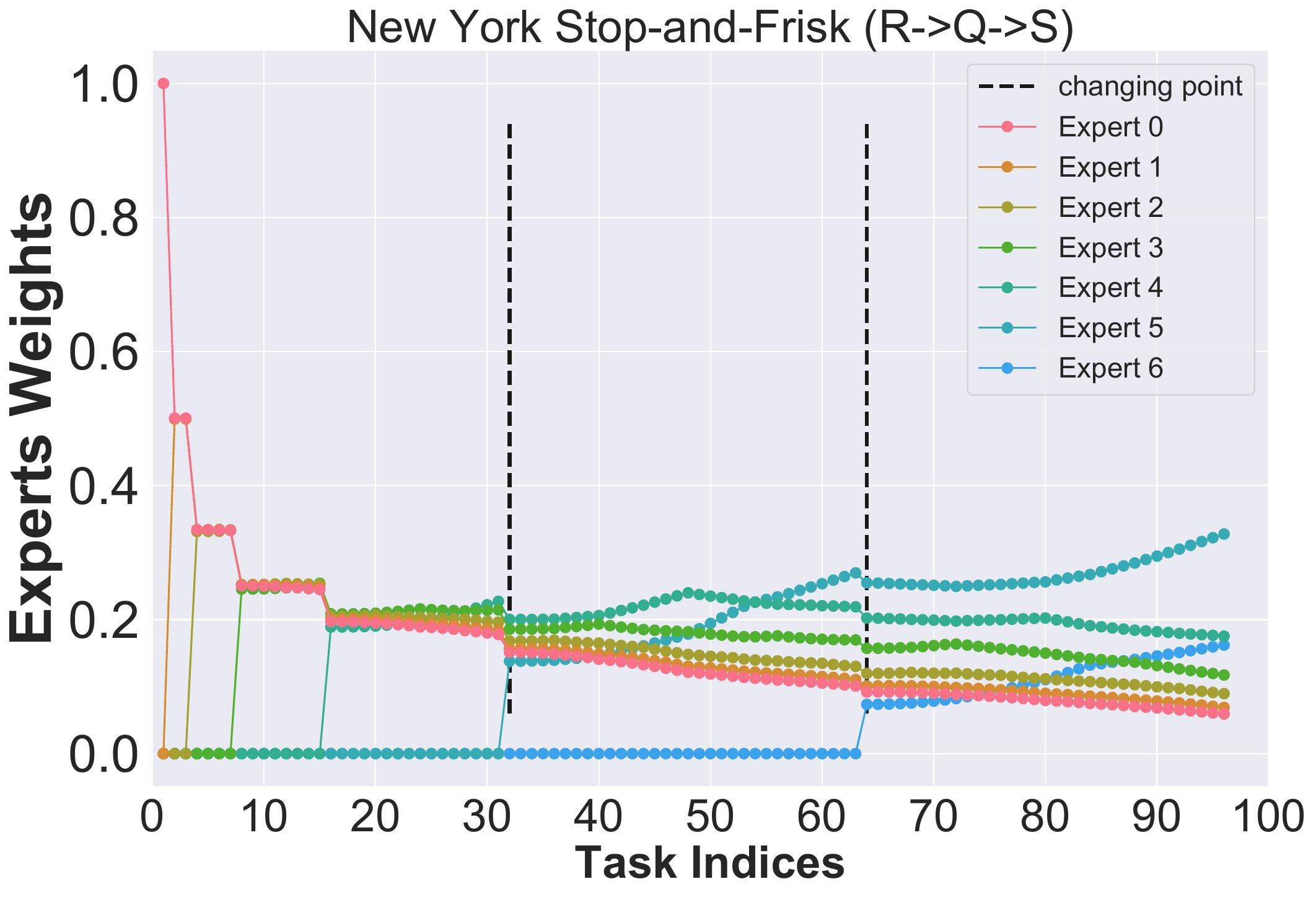}
        \caption{}
    \end{subfigure}
    \begin{subfigure}[b]{0.325\textwidth}
        \includegraphics[width=\textwidth]{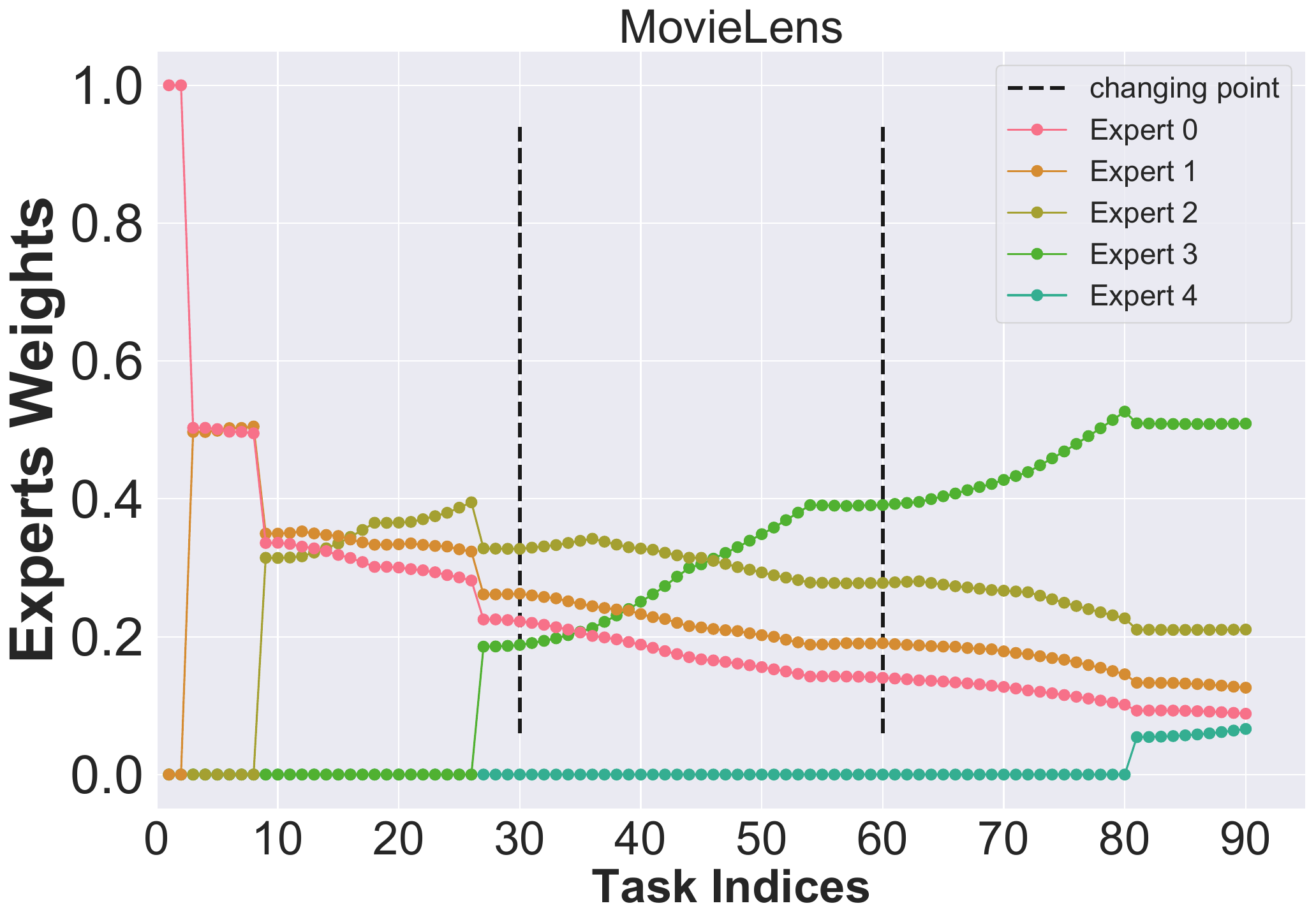}
        \caption{}
    \end{subfigure}
    
    \caption{Expert weight changes over time. \textbf{(a-c)} \sysname{}-AGC, \textbf{(d-f)} \sysname{}-DGC.}
    \label{fig:ps}
\end{figure*}

\begin{figure*}[t]
\captionsetup[subfigure]{aboveskip=-1pt,belowskip=-1pt}
    \centering
    \begin{subfigure}[b]{0.325\textwidth}
        \includegraphics[width=\textwidth]{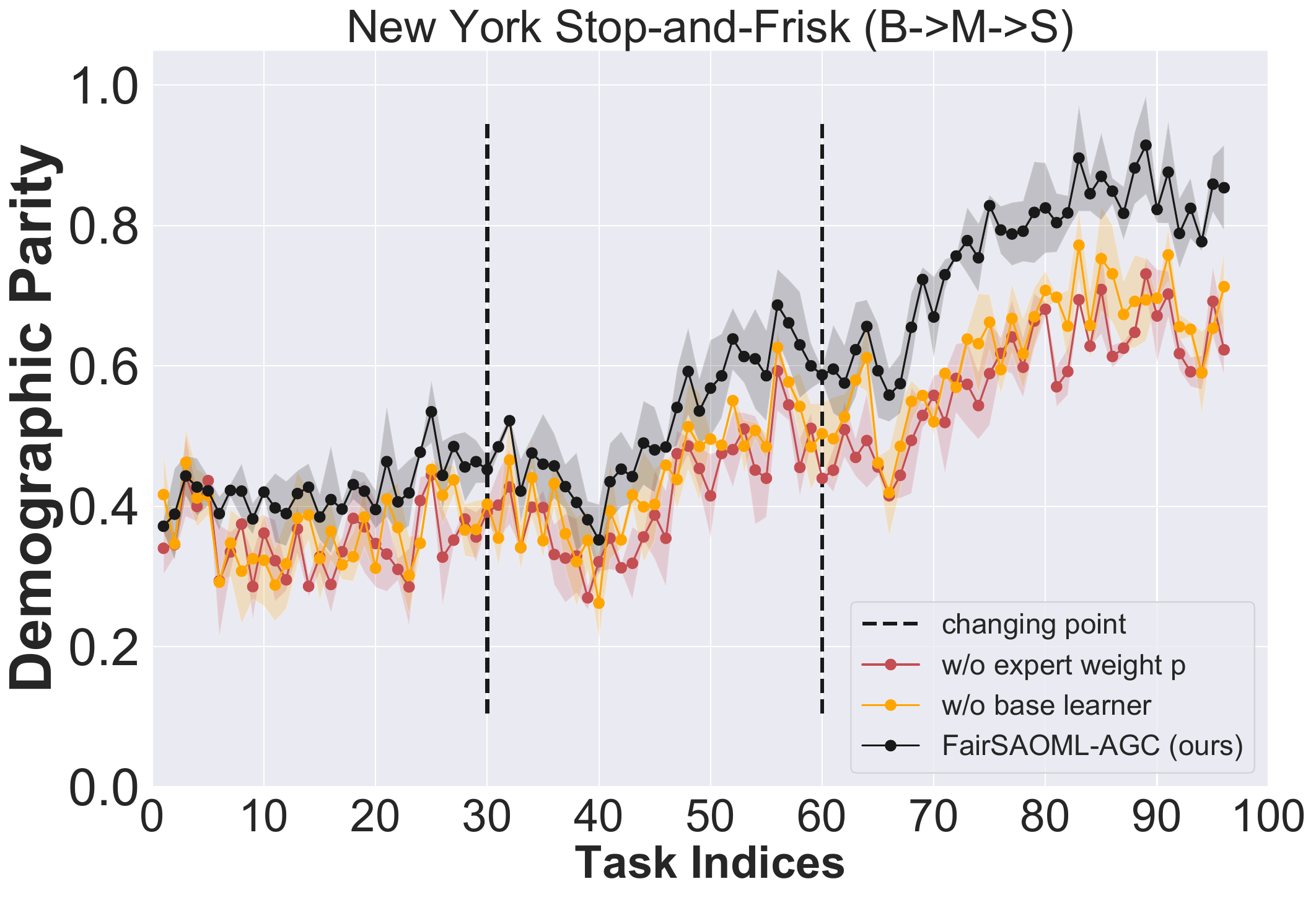}
        \caption{}
    \end{subfigure}
    \begin{subfigure}[b]{0.325\textwidth}
        \includegraphics[width=\textwidth]{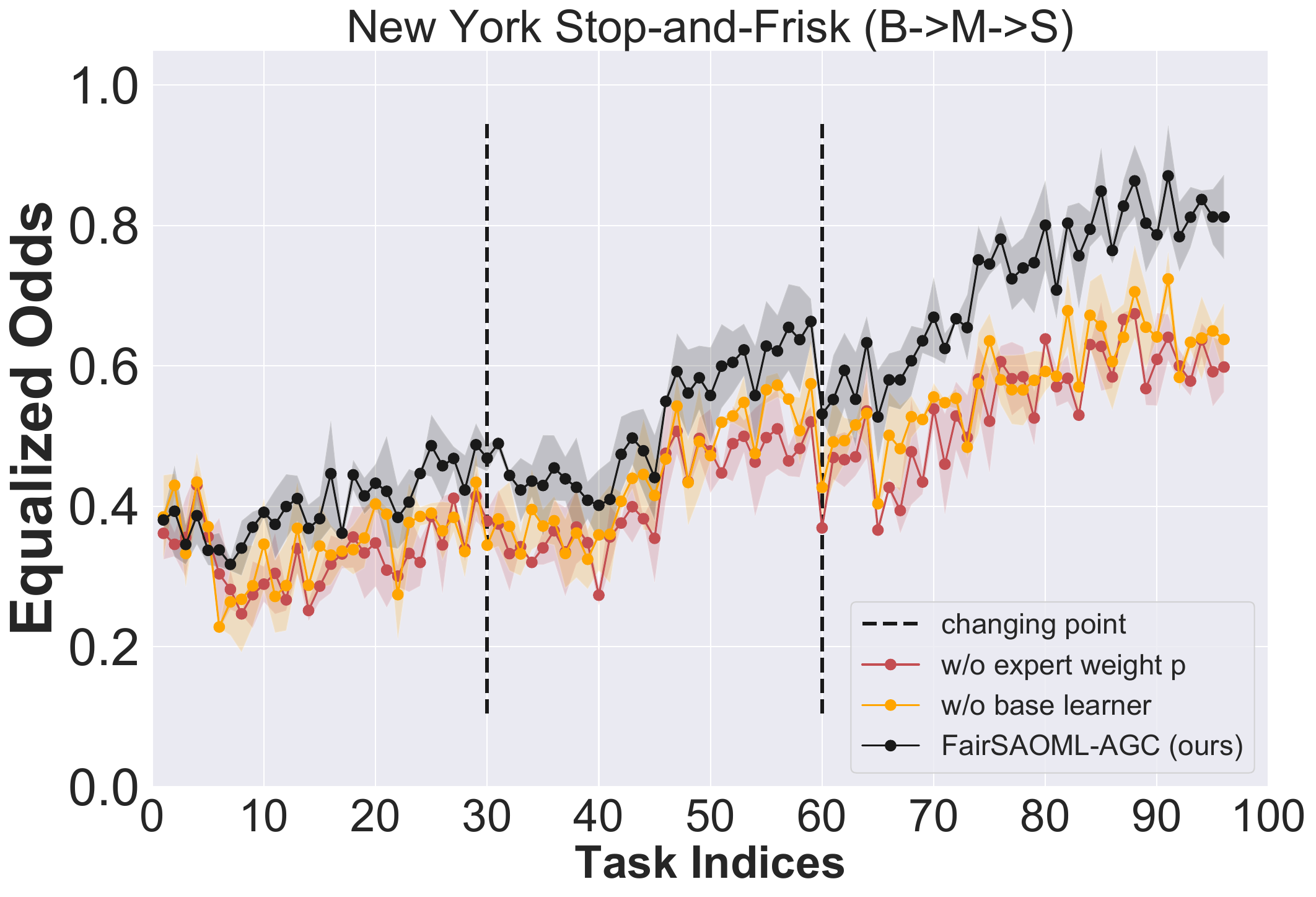}
        \caption{}
    \end{subfigure}
    \begin{subfigure}[b]{0.325\textwidth}
        \includegraphics[width=\textwidth]{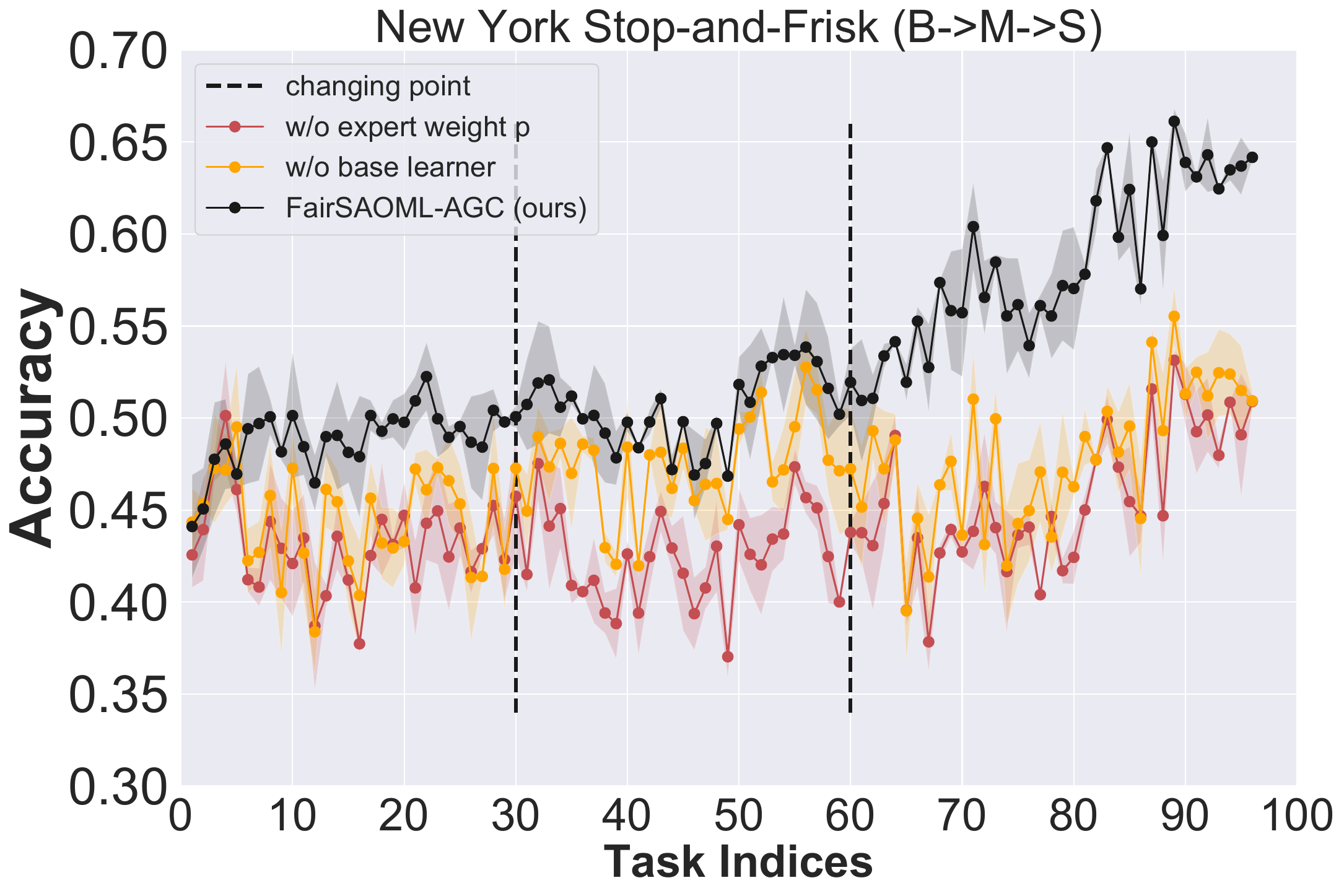}
        \caption{}
    \end{subfigure}

    \begin{subfigure}[b]{0.325\textwidth}
        \includegraphics[width=\textwidth]{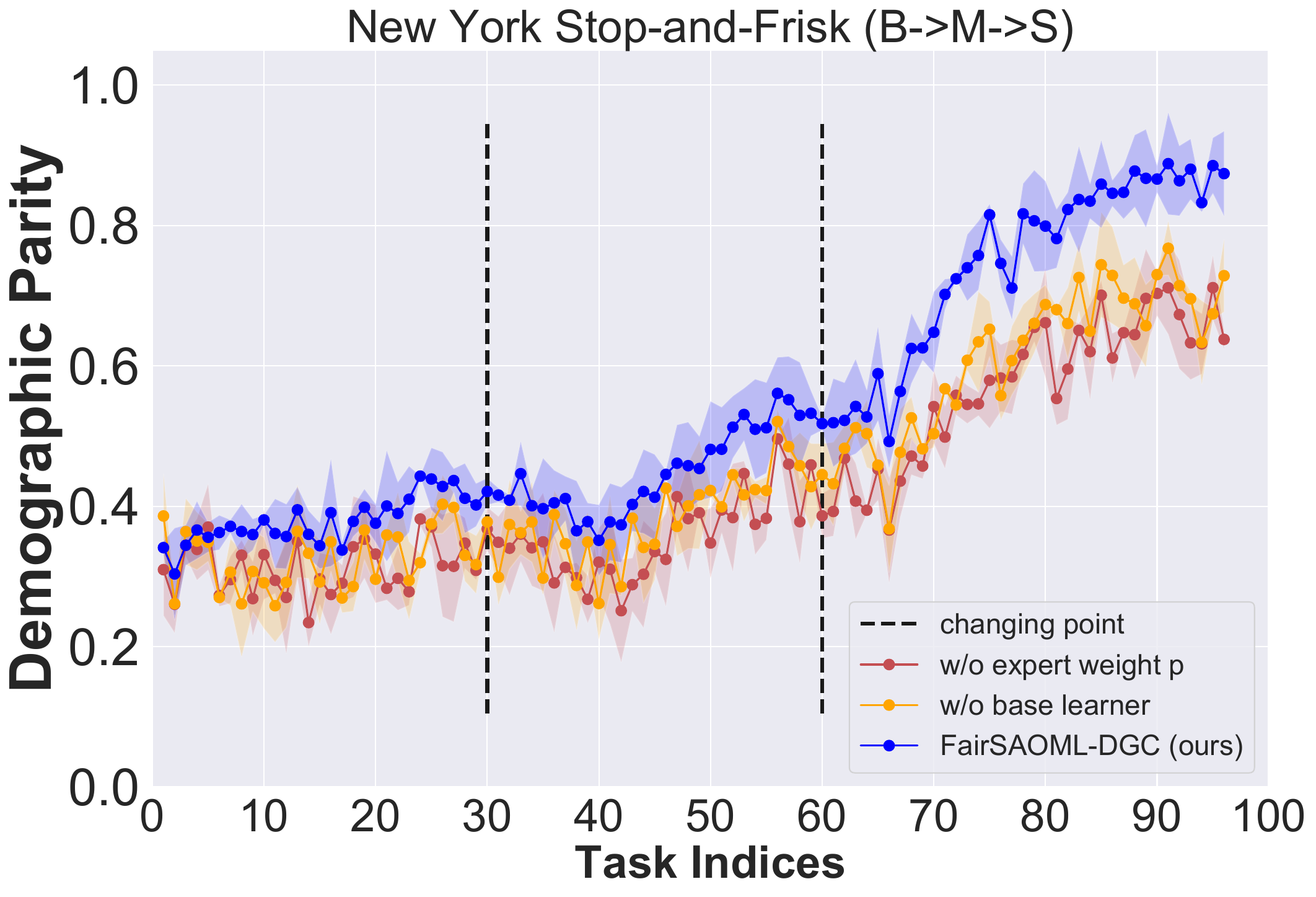}
        \caption{}
    \end{subfigure}
    \begin{subfigure}[b]{0.325\textwidth}
        \includegraphics[width=\textwidth]{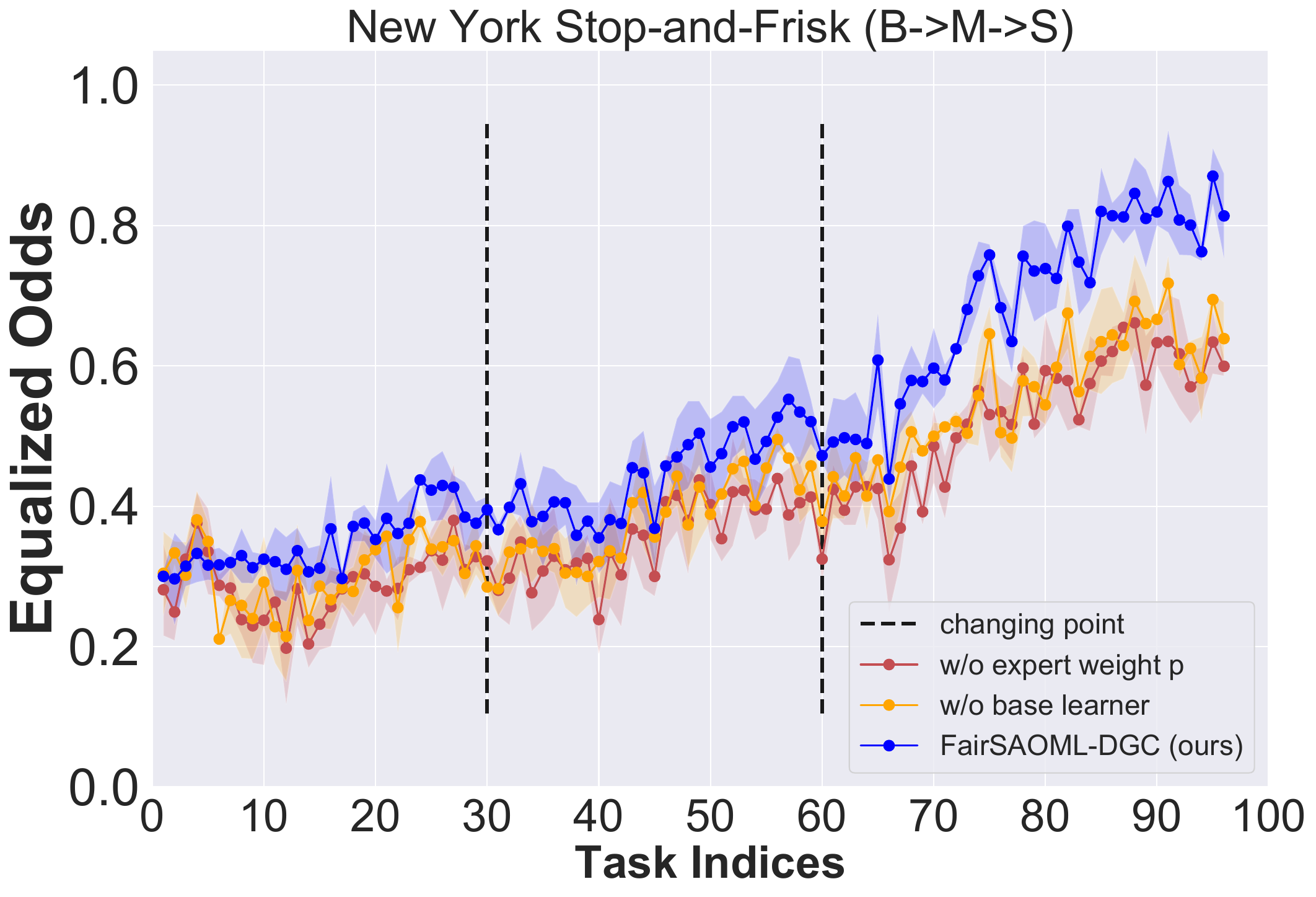}
        \caption{}
    \end{subfigure}
    \begin{subfigure}[b]{0.325\textwidth}
        \includegraphics[width=\textwidth]{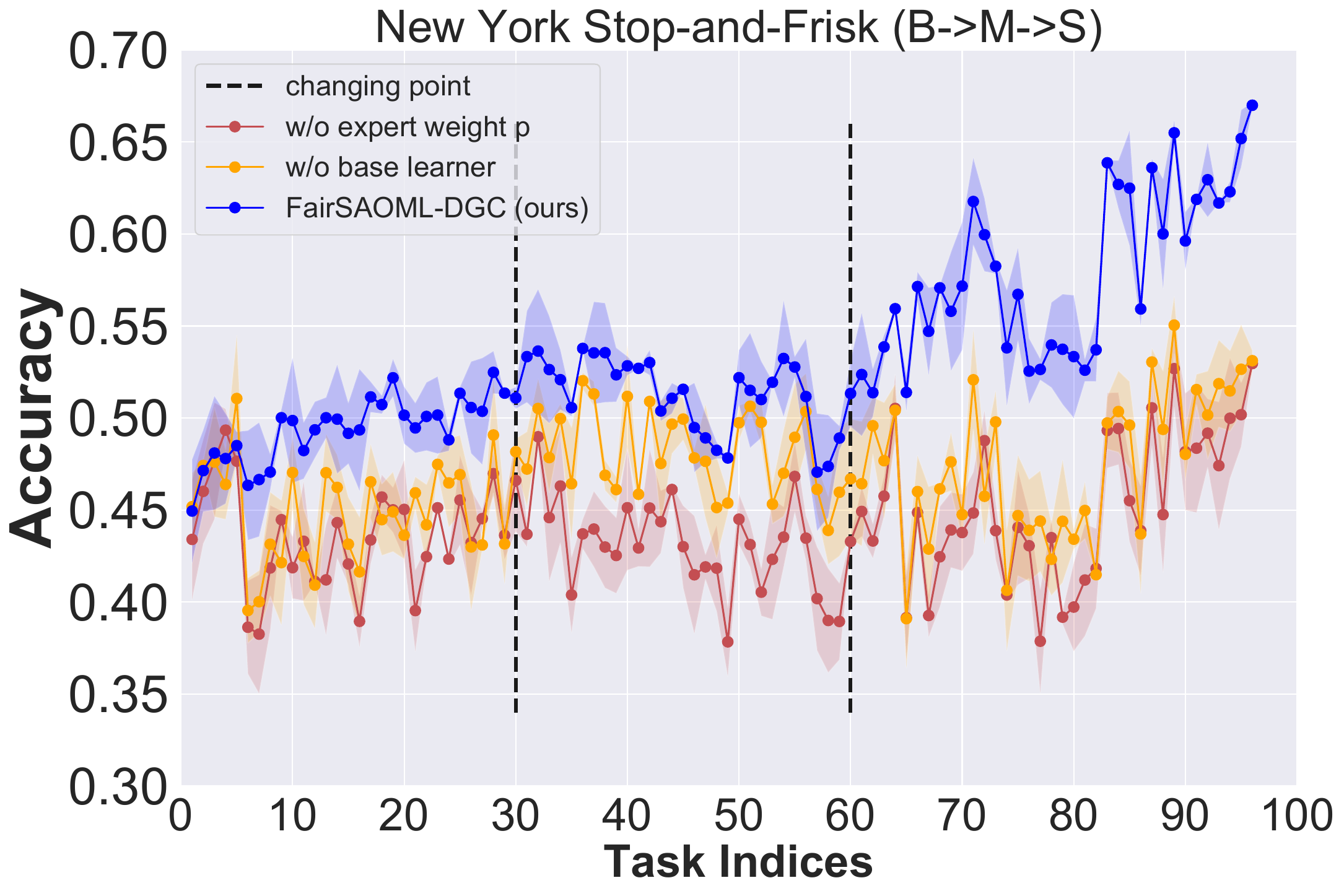}
        \caption{}
    \end{subfigure}
    
    \caption{Performance of ablation studies on the New York Stop-and-Frisk B$\rightarrow$M$\rightarrow$S dataset. \textbf{(a-c)} \sysname{}-AGC, \textbf{(d-f)} \sysname{}-DGC.}
    \label{fig:ablation}
\end{figure*}

\subsection{Overall Performance}
The consolidated results, depicted in Figure \ref{fig:dp-eop-acc}, provide a comprehensive evaluation of the effectiveness and efficiency of the proposed method, utilizing three evaluation metrics: fairness (DP and EO) and model precision (accuracy).

In all the curves presented for various methods, higher values indicate better performance across all plots. The shaded regions in the figures represent standard errors. The results demonstrate that our proposed \sysname{} with all interval settings effectively mitigates bias as the learner encounters more tasks, eventually satisfying the "80\%-rule" fairness condition \cite{Biddle-Gower-2005}, where DP and EO exceed $0.8$ in the latter stages. Furthermore, \sysname{} consistently outperforms most alternative approaches in terms of achieving the best model precision, as indicated by the high accuracy scores.

Regarding learning efficiency, our \sysname{} with the DI setting takes the most running time. In contrast, \sysname{} with AGC and DGC settings exhibit the shortest running times when compared to the baseline methods shown in the bar charts of Figure \ref{fig:dp-eop-acc}. This observation can be attributed to several factors: (1) the number of experts at each time in AGC and DGC significantly decreases compared to the one in DI; (2) only active experts, but not sleeping ones, make contributions for parameter updates; (3) instead of the entire data task, a data subset (support) is used for parameter updates within active experts.


\subsection{Adaptability to Changing Environments}
The primary objective of our experimental design is to assess the adaptability of \sysname{} concerning fairness and model accuracy as the environment transitions from one to another. To facilitate a clearer visualization of these changing environments, we have manually inserted vertical dotted lines in Figure \ref{fig:dp-eop-acc}, distinguishing the different environments at specific task indices. Our experimental findings reveal that, while \sysname{} may not initially outperform other baseline methods in the first environment, it excels in adapting to changing conditions. As a result, its performance consistently improves in terms of both model fairness and predictive accuracy as the environment evolves.

In Section \ref{sec:agc intervals and experts}, we introduced experts as crucial components in \sysname{}, where the model parameter pair $(\boldsymbol{\theta}_t,\boldsymbol{\lambda}_t)$ at time $t$ is determined by aggregating weighted expert advice. Figure \ref{fig:ps} illustrates the evolution of expert weights in \sysname{}-AGC and \sysname{}-DGC. We did not track the weight changes of experts in \sysname{}-DI due to its larger number of experts (96 in \textit{NYSF} and 90 in \textit{MovieLens}). Our observations are as follows: (1) Experts associated with longer intervals receive larger weights, and these weights continue to increase as the learner encounters more tasks; (2) Conversely, experts linked to shorter intervals receive smaller weights and become less influential over time. These findings align with expectations, as assigning heavier weights to experts with longer intervals empowers our \sysname{} to effectively adapt to the volatility in model performance induced by changing environments.

Among the baseline methods, MaskFTML demonstrates superior accuracy performance in the first environment, as evidenced in Figure \ref{fig:dp-eop-acc} (c, g, k). However, it falls short when it comes to achieving model fairness, suggesting that merely attempting to obscure the protected attribute from decision-makers is insufficient to improve prediction fairness. On the other hand, FairFML, FairAOGD, and FairGLC exhibit an ability to mitigate bias in the first environment. Still, they struggle to adapt both fairness and predictive accuracy when the environment undergoes changes. In contrast, AOD and CBCE, originally designed for online learning in dynamic environments, prioritize learning accuracy but do not effectively address model fairness when environmental shifts occur. Furthermore, the pursuit of higher accuracy in AOD often results in a trade-off in terms of fairness performance. These observations highlight the challenges and trade-offs involved in achieving a balance between accuracy and fairness across various methods in changing environments.

\subsection{Ablation Studies}
We conducted ablation studies on the \textit{NYSF} (B$\rightarrow$M$\rightarrow$S) dataset to assess the contributions of two pivotal components within \sysname{}: expert weights $p_{t,I}$ and the base learner, as described in Section \ref{sec:interval-level learning}.

To elaborate, meta-level parameters are computed at each time by aggregating expert decisions based on their respective weights. By removing expert weights, all experts contribute equally to the decision-making process. Furthermore, within active experts, base learners, as defined in Eq.(\ref{eq:inner-problem}), are employed to update model parameters at an interval level. Without base learners, all active experts share the same model parameters inherited from the previous time and are consequently assigned equal weight. The key insights from the results presented in Figure \ref{fig:ablation} are as follows: (1) Expert weights play a significant role in \sysname{}, indicating their importance in achieving effective bias control and predictive accuracy; (2) The inclusion of base learners serves to enhance model performance concerning bias control and predictive accuracy. These findings emphasize the critical contributions of expert weights and base learners to the overall effectiveness of the \sysname{} algorithm.


\subsection{Sensitive Analysis on Different Bases in AGC and DGC}
\begin{figure*}[!t]
\captionsetup[subfigure]{aboveskip=-1pt,belowskip=-1pt}
\centering
    \begin{subfigure}[b]{0.245\textwidth}
        \includegraphics[width=\textwidth]{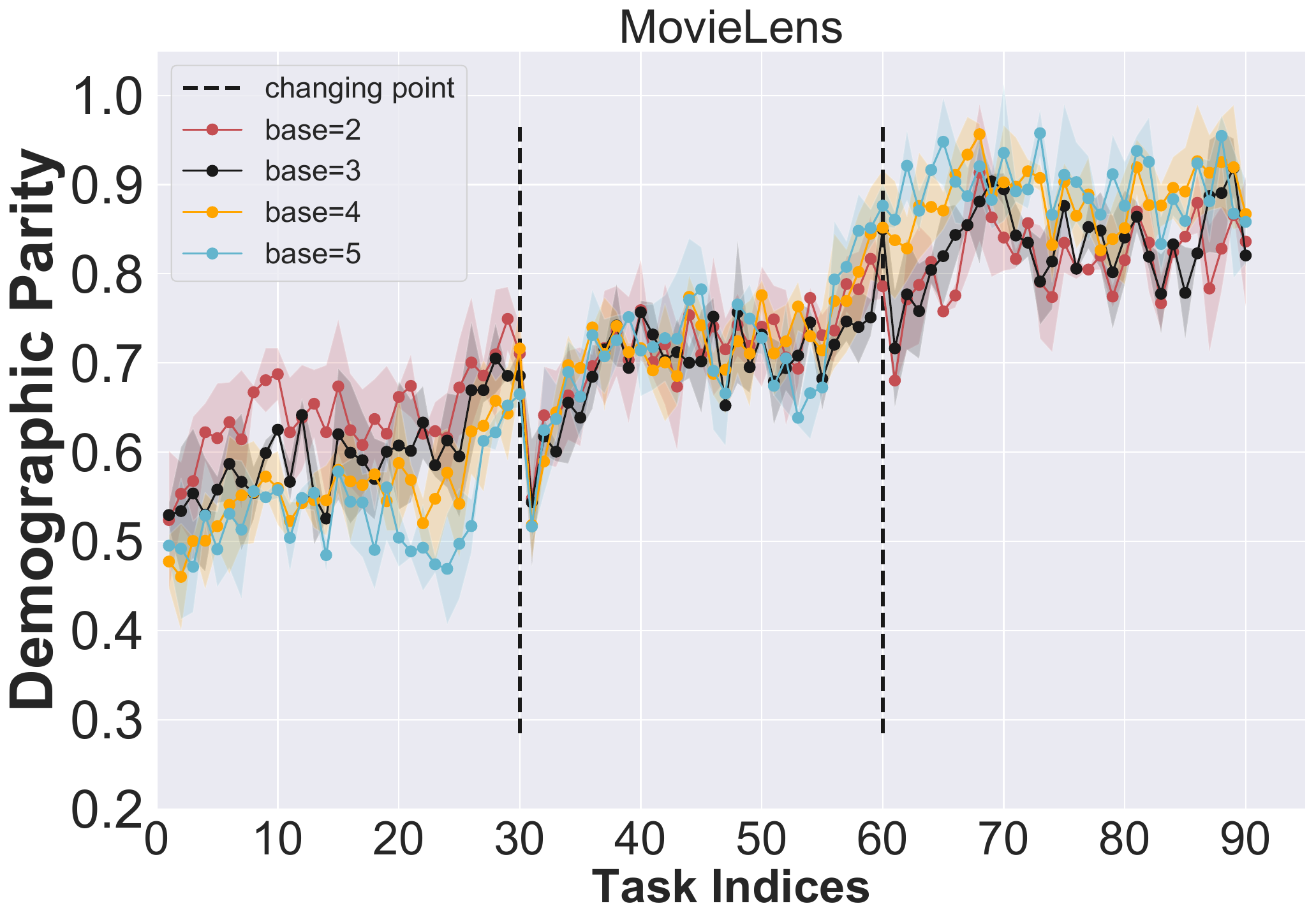}
        \caption{}
    \end{subfigure}
    \begin{subfigure}[b]{0.245\textwidth}
        \includegraphics[width=\textwidth]{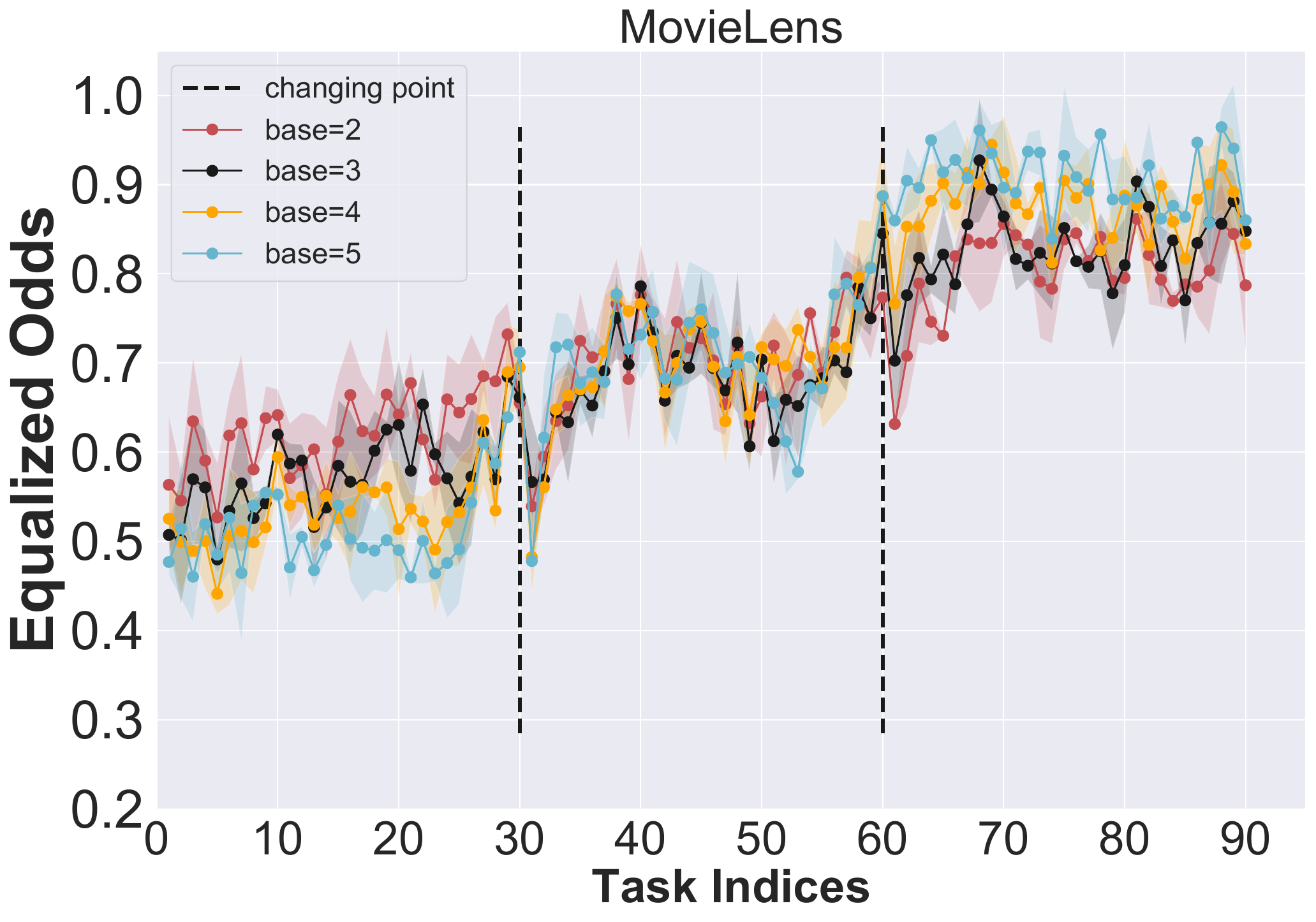}
        \caption{}
    \end{subfigure}
    \begin{subfigure}[b]{0.245\textwidth}
       \includegraphics[width=\textwidth]{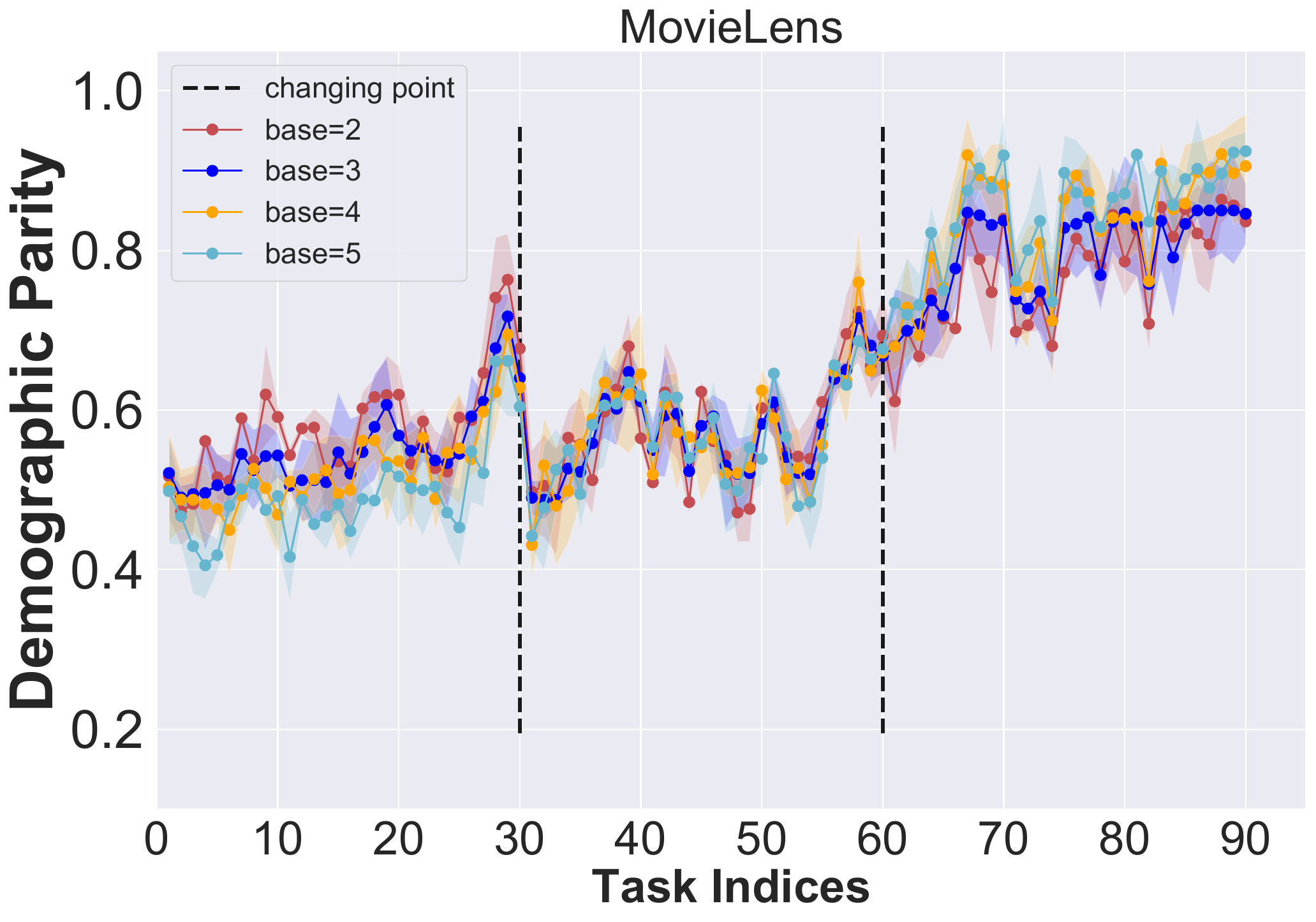}
        \caption{}
    \end{subfigure}
    \begin{subfigure}[b]{0.245\textwidth}
        \includegraphics[width=\textwidth]{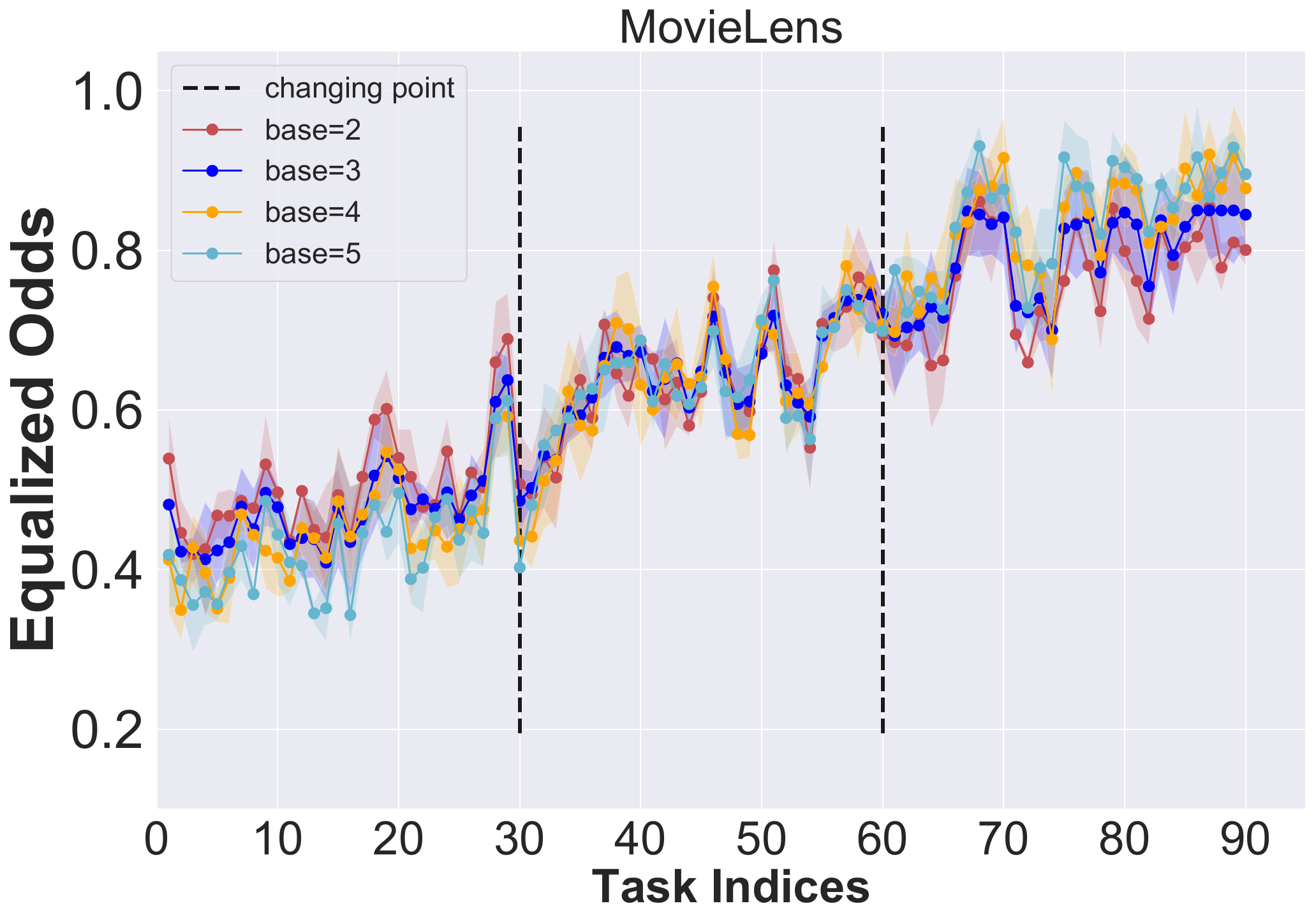}
        \caption{}
    \end{subfigure}
    \caption{Different choices of bases for \textbf{(a,b)} \sysname{}-AGC and \textbf{(c,d)} \sysname{}-DGC on the MovieLens dataset.}
    \label{fig:base-sensitive}
\end{figure*}

Sensitive analyses conducted on the \textit{MovieLens} dataset, as depicted in Figure \ref{fig:base-sensitive}, involve the subsetting of intervals using different bases selected from the set $\{2,3,4,5\}$. According to Eq.(\ref{eq:AGC}) and Eq.(\ref{eq:DGC}), the configuration with the smallest base value (\textit{i.e.,} 2) results in the highest number of experts (6 for AGC and 7 for DGC). Consequently, the largest expert in this setting carries the longest intervals (32 for AGC and 64 for DGC).

Our observations regarding model fairness reveal that settings with smaller bases exhibit slightly better performance than those with larger bases in the first environment. However, the opposite trend is observed in the last environment. This occurs for two main reasons: (1) In the first environment, the largest experts carry more information in the smaller base setting than in the larger base setting; (2) In the last environment, the largest experts in smaller base settings become less pure and incorporate data from different environments, leading to a deterioration in fairness. These findings underscore the sensitivity of the \sysname{} algorithm to the choice of base value and its impact on model fairness, particularly in different environmental contexts.


\section{Conclusion}
\label{sec:conclusion}
    To address the challenges of fairness-aware online learning in changing environments, where data tasks are sampled from diverse distributions one after another, we introduce a novel regret measure called \sysnameregret{}. \sysnameregret{} extends strongly adaptive regret by incorporating long-term fairness constraints. In technical terms, we start by proposing three alternative sets of intervals. At each time step, we dynamically select a target set consisting of multiple intervals from these sets. Next, we introduce a novel learning algorithm, named \sysname{}, to sequentially determine model parameters. In this algorithm, we dynamically activate a subset of experts based on the intervals in the target set and update their parameters at an interval level. The meta-level model parameters are then obtained by combining the weighted contributions of all experts. Detailed theoretical analysis and accompanying proofs provide justification for the efficiency and effectiveness of our proposed algorithm. We demonstrate upper bounds for loss regret and the violation of fairness constraints. Empirical studies conducted on real-world datasets demonstrate that our method outperforms state-of-the-art online learning techniques in terms of both model accuracy and fairness.


\begin{acks}
This work is supported by the Baylor University Startup funds, the National Science Foundation under grant numbers 2147375 and 1750911, and the National Center for Transportation Cybersecurity and Resiliency (TraCR) headquartered in Clemson, South Carolina, USA. 
Any opinions, findings, conclusions, and recommendations expressed in this material are those of the author(s) and do not necessarily reflect the views of TraCR, and the U.S. Government assumes no liability for the contents or use thereof.
\end{acks}

\bibliographystyle{ACM-Reference-Format}
\bibliography{references}


\end{document}